\documentclass{article}

\usepackage{iclr2026_conference,times}

\usepackage[utf8]{inputenc} 
\usepackage[T1]{fontenc}    
\usepackage{hyperref}       
\usepackage{url}            
\usepackage{booktabs}       
\usepackage{amsfonts}       
\usepackage{nicefrac}       
\usepackage{microtype}      
\usepackage{xcolor}         
\usepackage{makecell}
\usepackage{graphicx}
\usepackage{subcaption}
\usepackage{graphicx}
\usepackage{multirow}
\usepackage{array}
\usepackage{tabularx}
\usepackage{makecell}
\usepackage{booktabs}
\usepackage{enumitem}
\usepackage{xcolor}
\usepackage[table]{xcolor}
\usepackage{graphicx}
\usepackage{subcaption} 
\usepackage{caption}
\usepackage{float}       
\usepackage{tcolorbox}

\usepackage{xspace}

\definecolor{app1color}{HTML}{384995}
\definecolor{app2color}{HTML}{588169}
\definecolor{app3color}{HTML}{E86254}

\newcommand{\name}{\textsc{Generate Any Scene}\xspace}

\newcommand{\vision}{\textit{Text-to-Vision generation}\xspace}
\newcommand{\image}{\textit{Text-to-Image generation}\xspace}
\newcommand{\video}{\textit{Text-to-Video generation}\xspace}
\newcommand{\threed}{\textit{Text-to-3D generation}\xspace}

\newcommand{\sdonefive}{\textit{SDv1.5}\xspace}

\newcommand{\sdtwoone}{\textit{SDv2.1}\xspace}
\newcommand{\sdxl}{\textit{SDXL}\xspace}
\newcommand{\sdthree}{\textit{SD3 Medium}\xspace}
\newcommand{\playground}{\textit{Playground v2.5}\xspace}
\newcommand{\deepfloyd}{\textit{DeepFloyd IF}\xspace}
\newcommand{\fluxschnell}{\textit{FLUX.1-schnell}\xspace}
\newcommand{\fluxdev}{\textit{FLUX.1-dev}\xspace}
\newcommand{\pixartalpha}{\textit{PixArt-$\alpha$}\xspace}
\newcommand{\pixartsigma}{\textit{PixArt-$\Sigma$}\xspace}
\newcommand{\wuerstchen}{\textit{Wuerstchen v2}\xspace}
\newcommand{\dalle}{\textit{DaLL-E 3}\xspace}
\newcommand{\flux}{\textit{FLUX1.1 PRO}\xspace}

\newcommand{\modelscope}{\textit{ModelScope}\xspace}
\newcommand{\texttovideozero}{\textit{Text2Video-Zero}\xspace}
\newcommand{\zeroscope}{\textit{ZeroScope}\xspace}
\newcommand{\cogvideox}{\textit{CogVideoX-2B}\xspace}
\newcommand{\VideoCrafter}{\textit{VideoCrafter2}\xspace}
\newcommand{\AnimateLCM}{\textit{AnimateLCM}\xspace}
\newcommand{\AnimateDiff}{\textit{AnimateDiff}\xspace}
\newcommand{\FreeInit}{\textit{FreeInit}\xspace}
\newcommand{\OpenSora}{\textit{Open-Sora 1.2}\xspace}

\newcommand{\SJC}{\textit{SJC}\xspace}
\newcommand{\DreamFusion}{\textit{DreamFusion}\xspace}
\newcommand{\Magic}{\textit{Magic3D}\xspace}
\newcommand{\Latentnerf}{\textit{Latent-NeRF}\xspace}
\newcommand{\ProlificDreamer}{\textit{ProlificDreamer}\xspace}
\newcommand{\clipscore}{\textit{Clip Score}\xspace}
\newcommand{\vqascore}{\textit{VQA Score}\xspace}
\newcommand{\imagereward}{\textit{ImageReward Score}\xspace}
\newcommand{\tifa}{\textit{TIFA Score}\xspace}
\newcommand{\pickscore}{\textit{Pick Score}\xspace}

\title{Generate Any Scene: Scene graph driven data synthesis for  Visual Generation Training}

\author{
Ziqi Gao$^{1}$\footnotemark[1], Weikai Huang$^{1}$\footnotemark[1], Jieyu Zhang$^{1}$, Aniruddha Kembhavi$^{2}$, Ranjay Krishna$^{1,2}$\\
$^{1}$University of Washington, 
$^{2}$Allen Institute for Artificial Intelligence\\
\begin{tabular}{l}
\hspace*{0.4mm}\includegraphics[height=3mm]{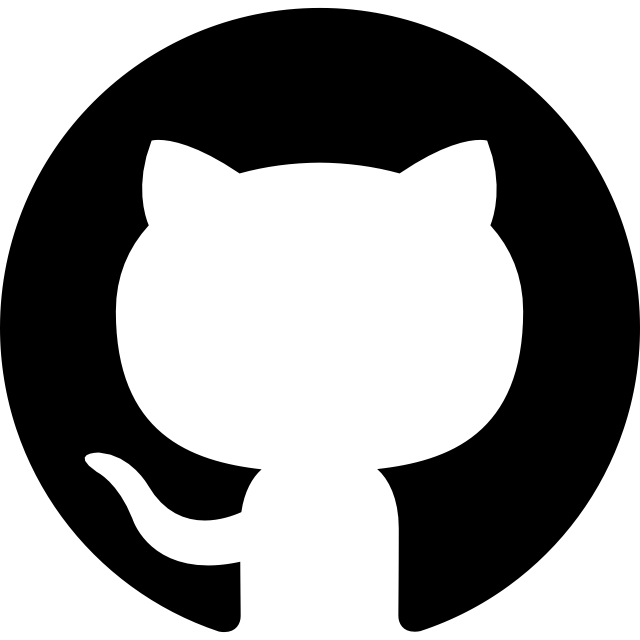} Code: \url{https://github.com/RAIVNLab/GenerateAnyScene}\\
\raisebox{-0.7mm}{\includegraphics[height=4mm]{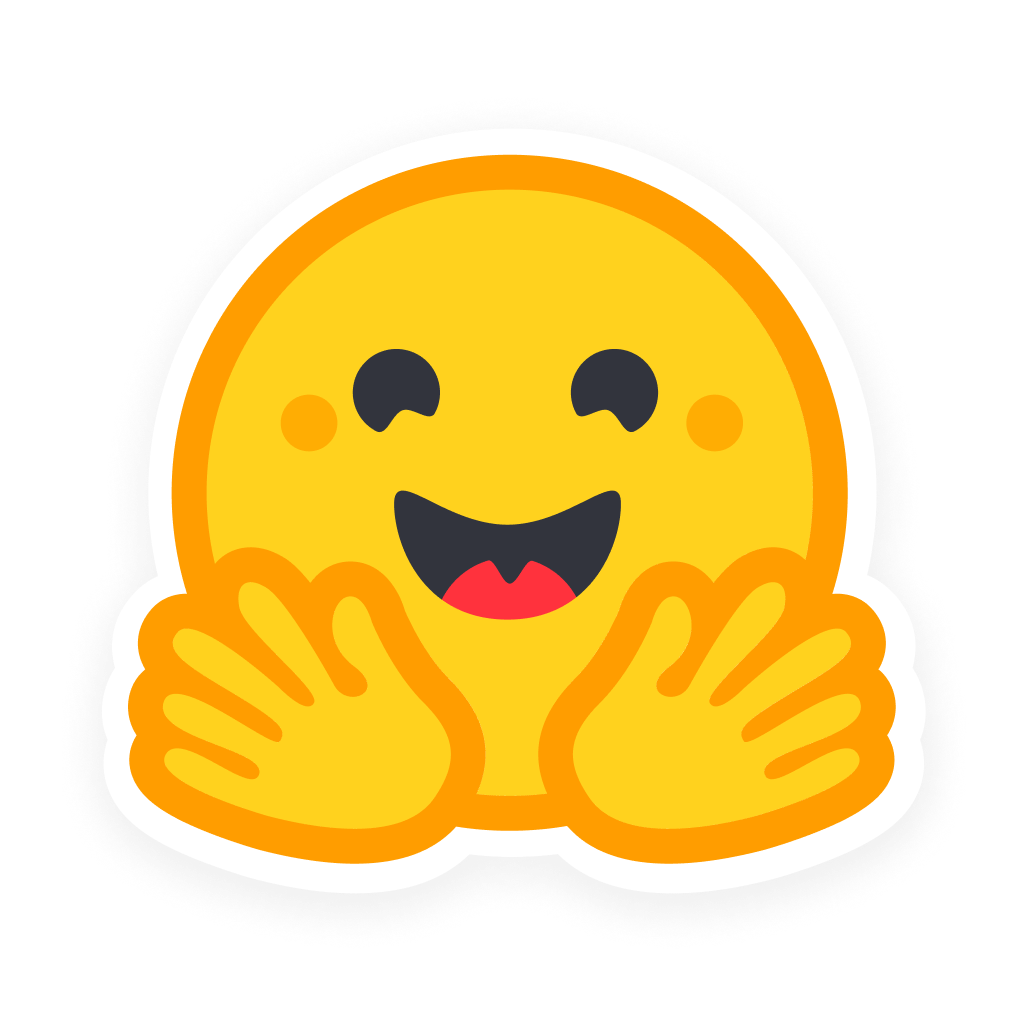}}Dataset: \href{https://huggingface.co/collections/UWGZQ/generate-any-scene-6827150ca377214d1c97dd25}{GenerateAnyScene Dataset}
\end{tabular}
}

\iclrfinalcopy
\begin{document}

\maketitle

\begin{abstract}
Recent advances in text-to-vision generation excel in visual fidelity but struggle with compositional generalization and semantic alignment. 
Existing datasets are noisy and weakly compositional, limiting models' understanding of complex scenes, while scalable solutions for dense, high-quality annotations remain a challenge.
We introduce \name, a data engine that systematically enumerates scene graphs representing the combinatorial array of possible visual scenes. 
\name dynamically constructs scene graphs of varying complexity from a structured taxonomy of objects, attributes, and relations. 
Given a sampled scene graph, \name translates it into a caption for text-to-image or text-to-video generation; it also translates it into a set of visual question answers that allow automatic evaluation and reward modeling of semantic alignment. 
Using \name, we first design a self-improving framework where models iteratively enhance their performance using generated data. \sdonefive achieves an average \textbf{4\%} improvement over baselines and surpasses fine-tuning on CC3M.
Second, we also design a distillation algorithm to transfer specific strengths from proprietary models to their open-source counterparts. Using fewer than 800 synthetic captions, we fine-tune \sdonefive and achieve a \textbf{10\%} increase in TIFA score for compositional and hard-concept generation.
Third, we create a reward model to align model generation with semantic accuracy at a low cost. Using GRPO algorithm, we fine-tune SimpleAR-0.5B-SFT and surpass CLIP-based methods by \textbf{+5\%} on DPG-Bench.
Finally, we apply these ideas to the downstream task of content moderation where we train models to identify challenging cases by learning from synthetic data. 
\end{abstract}

\section{Introduction}
Despite the high-fidelity of modern generative models (text-to-image and text-to-video), we are yet to witness wide-spread adoption~\cite{sora,dalle2,dalle3,prolificdreamer,Chen2023PixArtFT}.
Controllability remains out of reach~\cite{peng2024controlnext}.
Generated content appears realistic but often falls short of semantic alignment~\cite{Huang2023T2ICompBenchAC,Sun2024T2VCompBenchAC,ghosh2023geneval,hu2024ella}.
Users prompt models with a specific concept in mind. For example, when prompted to generate a scene of ``a black dog chasing after a rabbit that is eating grass, in Van Gogh's style, with starlight lightning'', some models are likely to generate an image of a dog but might miss the rabbit or get the style incorrect.

We hypothesize that these limitations stem not only from architectural bottlenecks but more fundamentally from the lack of structured, compositionally rich training data~\cite{dalle3}, especially those with uncommon compositions.
Popular datasets such as LAION~\cite{schuhmann2022laion} and CC3M~\cite{sharma2018conceptual} predominantly consist of web-crawled image-caption pairs, which are inherently noisy, weakly compositional, and biased toward single-object, coarse-grained descriptions. Such datasets lack explicit grounding of object-attribute relations and multi-object interactions, restricting models' ability to generalize to complex visual scenes. Efforts to enhance caption quality~\cite{dalle3,li2024laion} have demonstrated that enhancing the compositional density and semantic richness of captions can significantly improve generative performance. Nevertheless, manual curation of such dense compositional annotations is labor-intensive, while automatic annotation methods (e.g., via MLMs) suffer from hallucination and semantic noise.

Constructing a compositional dataset requires that we first define \emph{the space of the visual content}.
Scene graphs are one such representation of the visual space~\cite{krishna2017visual,johnson2015image,ost2021neural,johnson2018image,ji2020action}, grounded in cognitive science~\cite{biederman1987recognition}.
A scene graph represents objects in a scene as individual nodes in a graph. Each object is modified by attributes, which describe its properties. For example, attributes can describe the material, color, size, and location of the object in the scene. Finally, relationships are edges that connect the nodes. They define the spatial, functional, social, and interactive relationships between objects~\cite{lu2016visual}.
For example, in a living room scene, a “table” node might have attributes like “wooden” or “rectangular” and be connected to a “lamp” node through a relation: “on top of”. This systematic scene graph structure provides simple yet effective ways to define and model the scene. As such, scene graphs are an ideal foundation for systematically defining the compositional space of visual content in text-to-vision generation.

We introduce \name, a system capable of efficiently enumerating the space of scene graphs representing a wide range of visual scenes.
\name composes scene graphs of any structure using a rich taxonomy of visual elements, translating each scene graph into an input caption and visual question answers to evaluate the output image or video.
In particular, we first construct a rich taxonomy of visual concepts consisting of $28,787$ objects, $1,494$ attributes, $10,492$ relations, $2,193$ scene attributes from various sources. Based on these assets, \name can synthesize an almost infinite number of scene graphs of varying complexity~\cite{zhang2024task}. 
Besides, \name allows configurable scene graph generation. For example, evaluators can specify the complexity level of the scene graph to be generated or provide a seed scene graph to be expanded.
By automating these steps, our system ensures both scalability and adaptability, providing researchers and developers with diverse, richly detailed scene graphs and corresponding captions tailored to their specific needs. We also conduct comprehensive text-to-vision evaluations using our generated captions, as detailed in Appendix~\ref{appendix:eval}.





We show that \name can allow generation models to self-improve. Our diverse captions can facilitate a framework to iteratively improve \vision models using their own generations.
Given a model, we generate multiple images, identify the highest-scoring one, and use it as new fine-tuning data to improve the model itself.
We fine-tune \sdonefive~\cite{stablediffusion2} and achieve an average of \textbf{4\%} performance boost compared with original models, and this method is even better than fine-tuning with the same amount of real images and captions from the Conceptual Captions CC3M over different benchmarks.

We also use \name to design targeted distillation algorithms. Using our evaluations, we identify limitations in open-sourced models that their proprietary counterparts excel at.
Next, we distill these specific capabilities from proprietary models.
For example, \dalle~\cite{dalle3} excels particularly in generating composite images with multiple parts. We distill this capability into \sdonefive, effectively bridging the gap between \dalle and \sdonefive. After targeted fine-tuning, \sdonefive achieves a \textbf{10\%} increase in TIFA score~\cite{hu2023tifaaccurateinterpretabletexttoimage} for compositional tasks and hard concept generation.


Then we propose a low-cost scene graph-based reward model for RLHF~\cite{Ouyang2022TrainingLM} in text-to-image generation.
By leveraging synthetic scene graphs generated by \name, we generate exhaustive question-answer pairs that cover all objects, attributes, and relationships in the caption. Our method enables fine-grained, compositional reward modeling without manual annotation or heavy LLM inference. With GRPO~\cite{shao2024deepseekmathpushinglimitsmathematical}, we fine-tune SimpleAR-0.5B-SFT~\cite{wang2025simplear} using a scene graph reward model, achieving better compositional alignment than CLIP-based methods~\cite{radford2021learningtransferablevisualmodels} (\textbf{+5\%} on DPG-Bench~\cite{hu2024ellaequipdiffusionmodels}).

Finally, we apply \name to the downstream application of content moderation. 
Content moderation is a vital application, especially as \vision models improve. 
A key challenge lies in the limited diversity of existing training data. To address this, we leverage \name to generate diverse and compositional captions, creating synthetic training data that complements existing datasets.
By retraining a ViT-T~\cite{dosovitskiy2021imageworth16x16words} detector with our enriched dataset, we enhance its detection performance, particularly in cross-model and cross-dataset scenarios.

\section{Generate Any Scene}

\begin{figure*}[!t]
    \centering
    \includegraphics[width=1\linewidth,page=1]{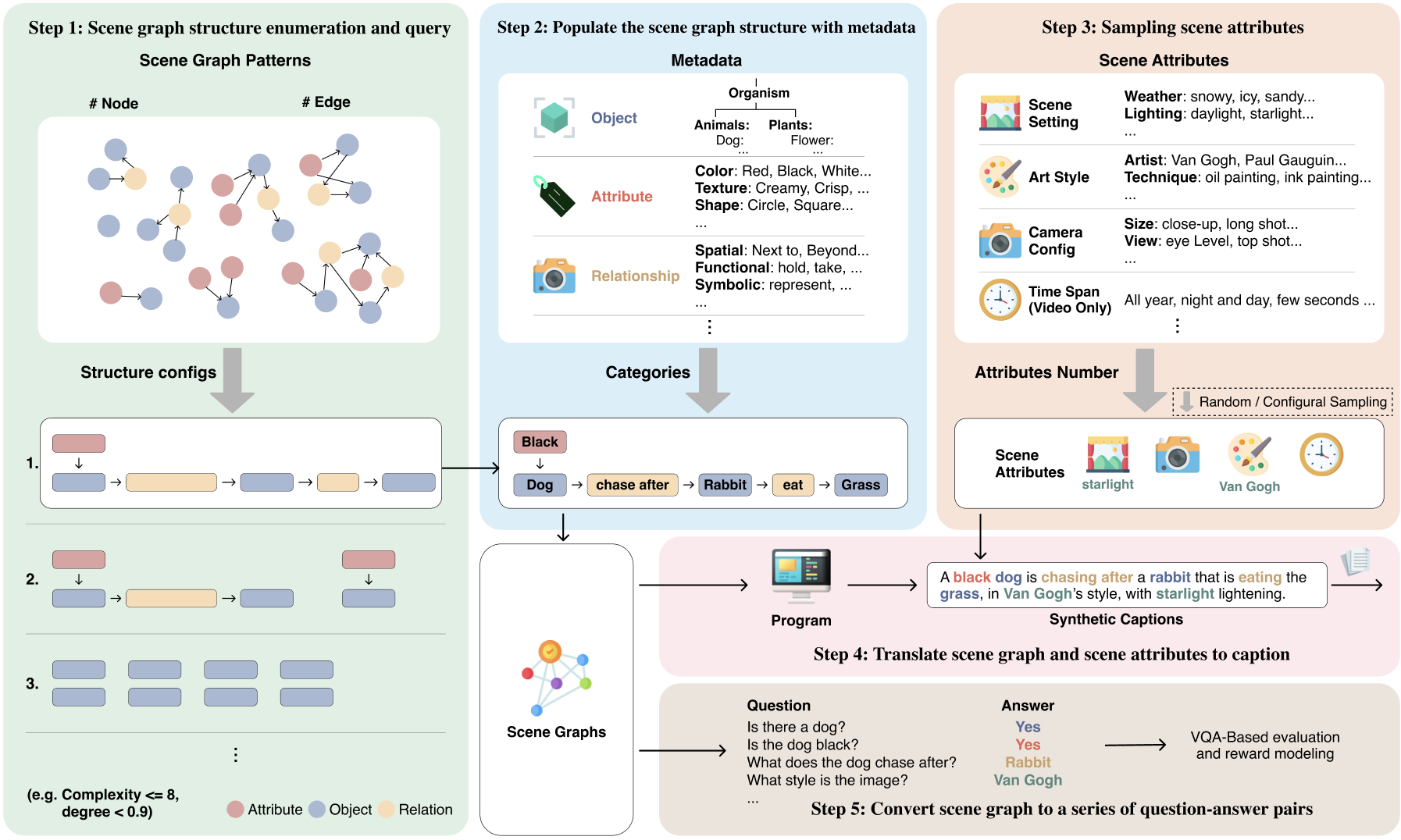}
    \caption{The generation pipeline of \name. \textbf{Step 1:} Enumerate diverse scene graph structures under user-defined constraints. \textbf{Step 2:} Populate structures with sampled objects, attributes, and relations. \textbf{Step 3:} Sample scene attributes such as style, perspective, or time span. \textbf{Step 4:} Translate scene graph and attributes into coherent captions. \textbf{Step 5:} Automatically generate QA pairs covering all elements for evaluation and reward modeling.}
    \label{FIG:scenegraphsys}
\end{figure*}

In this section, we present \name (Figure~\ref{FIG:scenegraphsys}), a data engine that systematically synthesizes diverse scene graphs in terms of both structure and content and translates them into corresponding captions.

\noindent\textbf{Scene graph.}
A scene graph is a structured representation of a visual scene, where objects are represented as nodes, their attributes (such as color and shape) are properties of those nodes, and the relationships between objects (such as spatial or semantic connections) are represented as edges. In recent years, scene graphs have played a crucial role in visual understanding tasks, such as those found in Visual Genome~\cite{krishna2017visual} and GQA~\cite{hudson2019gqa} for visual question answering (VQA). Their utility has expanded to various \vision tasks. For example, the DSG~\cite{Cho2023DavidsonianSG} and DPG~\cite{hu2024ella} benchmarks leverage scene graphs to evaluate how well generated images align with captions.


\noindent\textbf{Taxonomy of visual elements.}
To construct a scene graph, we use three main metadata types: \textbf{objects}, \textbf{attributes}, and \textbf{relations}. We further introduce \textbf{scene attributes} that capture global visual contexts, such as art style, to facilitate comprehensive caption synthesis. The statistics and source of our metadata are shown in Table~\ref{tab:metadata}. Additionally, we build a hierarchical taxonomy that categorizes metadata into distinct levels and types, enabling fine-grained analysis. This structure supports precise content synthesis, from broad concepts like “flower” to fine-grained instances such as “daisy.” 

\begin{table}[!h]
    \centering
    \small
    \caption{Summary of the quantities and sources of visual elements. Details are in Appendix~\ref{app:taxonomy}.}
    \renewcommand{\arraystretch}{1.0}
    \resizebox{0.9\textwidth}{!}{%
    \begin{tabular*}{\textwidth}{@{\extracolsep{\fill}} l l l}
        \toprule
        \textbf{Metadata Type} & \textbf{Number} & \textbf{Source} \\
        \midrule
        Objects & 28,787 & WordNet~\cite{wordnet} \\
        Attributes & 1,494 & Wikipedia~\cite{wikipedia_colors}, etc. \\
        Relations & 10,492 & Synthetic Visual Genome~\cite{robin} \\
        Scene Attributes & 2,193 & Places365~\cite{location-meta}, etc. \\
        \bottomrule
    \end{tabular*}
    }
    \label{tab:metadata}
\end{table}

\subsection{Generating data with scene graphs}
\label{PromptPipeline}





\noindent\textbf{Step 1: Scene graph structure enumeration.}
Our engine pre-computes a library of directed scene-graph topologies subject to user-specified \emph{structural constraints}: complexity (total number of objects, relations, and attributes)~\cite{grunde2021agqa}, average node degree, and number of connected components. We first sample the number of object nodes and then systematically enumerate feasible edge sets and attribute attachments that satisfy these constraints. We provide 3 optional controls: (i) \emph{degree-profile} bounds per-node in/out-degree, (ii) \emph{seed-graph preservation} embeds a user-provided seed graph as a subgraph of each enumerated structure, and (3) \emph{commonsense plausibility filtering} prunes implausible contents while retaining compositional diversity {(See Appendix.~\ref{filtering_details})}. All enumerations are performed once per parameter tuple and cached for fast querying. 



\noindent\textbf{Step 2: Populate the scene graph structure with metadata.}
Given a generated scene graph structure, the next step involves populating the graph with metadata. For each object node, attribute node, and relation edge, we sample the corresponding content from our metadata. This process is highly customizable and controllable: users can define the topics and types of metadata to include, for instance, by selecting only commonsense metadata or specifying relationships between particular objects. By determining the scope of metadata sampling, we can precisely control the final content of the captions and easily extend the diversity and richness of scene graphs by adding new metadata.


\noindent\textbf{Step 3: Sample scene attributes.}
We also include scene attributes that describe aspects such as the art style, viewpoint, time span (for video), and 3D attributes (for 3D content). These scene attributes are sampled directly from our metadata, creating a list that provides contextual details to enrich the description of the visual content.

\noindent\textbf{Step 4: Translate scene graph to caption.}
We introduce a deterministic and programmatic algorithm that converts scene graphs with scene attributes into captions. It traverses scene graphs by converting objects/attributes/relations into descriptive text in topological order, while tracking each object's references to ensure coherence. Programmatic grammar rules are employed (e.g., disambiguating identical objects with “the first/second” and skipping already mentioned objects) to prevent duplication and misreference, resulting in clear captions. We also provide LLM paraphrasing as an optional step to diversify wording; however, our studies (see Appendix~\ref{appendix:Phrasing-Robustness}) show that paraphrasing does not materially affect results. We adopt the programmatic caption converter as the default for its speed and low hallucination rate.

\noindent\textbf{Step 5: Convert scene graph to a series of question-answer pairs.} \label{step5}
Given a synthetic scene graph, \name automatically enumerates exhaustive QA pairs using templates that query object attributes (e.g., What color is the sphere?), spatial relations (e.g., What is to the left of the cube?), and other compositional elements. Each answer maps directly to an object, attribute, or edge, ensuring full coverage of the graph at minimal cost. This enables both VQA-based evaluation of generated images and the construction of fine-grained reward models without manual labeling or costly LLM inference.


\section{Self-Improving models with synthetic captions}

\begin{figure}[h!]
    \centering
    \includegraphics[width=\linewidth,page=1]{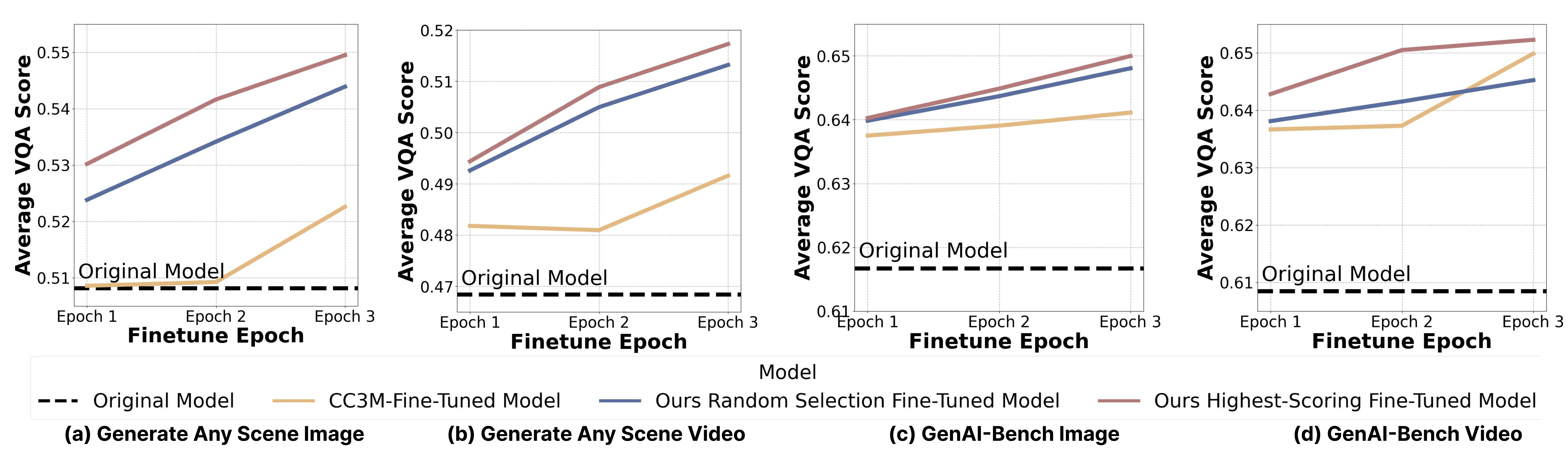}
    \caption{\textbf{Results for Self-Improving Models}.
    Average VQA score of \sdonefive fine-tuned on different data across 1K \name image/video evaluation set and GenAI-Bench image/video benchmark~\cite{li2024genai}.}
    \label{general-bench}
\end{figure}


With \name, we develop a self-improvement framework to improve generative capabilities. By generating scalable compositional captions from scene graphs, \name expands the textual and visual space, allowing for a diversity of synthetic images that extend beyond real-world scenes. Our goal is to utilize these richly varied synthetic images to further boost model performance.

\noindent\textbf{Iterative self-improving framework.}
Inspired by DreamSync~\cite{Sun2023DreamSyncAT}, we designed an iterative self-improving framework using \name with \sdonefive as the baseline model. With \vqascore, which shows strong correlation with human evaluations on compositional images~\cite{Lin2024EvaluatingTG}, we guide the model's improvement throughout the process.
Specifically, \name generates 3 $\times$ 10K captions across three epochs. For each caption, \sdonefive generates 8 images, and the image with the highest \vqascore is selected. From each set of 10K optimal images, we then select the top 25\% (2.5K image-caption pairs) as the training data for each epoch. In subsequent epochs, we use the fine-tuned model from the prior iteration to generate new images. We employ LoRA~\cite{Hu2021LoRALA} for parameter-efficient fine-tuning.

\noindent\textbf{Baselines.} We conduct comparative experiments with the CC3M dataset, which comprises high-quality and diverse real-world image-caption pairs~\cite{sharma2018conceptual}. We randomly sample 3 $\times$ 10K captions from CC3M, applying the same top-score selection strategy for iterative fine-tuning of \sdonefive. Additionally, we include a baseline using random-sample fine-tuning strategy to validate the advantage of our highest-scoring selection-based strategy. We evaluate our self-improving pipeline on \vision benchmarks, including GenAI Bench~\cite{li2024genai}. For the \video task, we use \texttovideozero as the baseline model, substituting its backbone with the original \sdonefive and our fine-tuned \sdonefive models.

\begin{table}[!htp]
\centering
\small
\setlength{\tabcolsep}{3pt} 
\noindent
\begin{minipage}{0.48\linewidth}
    \centering
    {
    \caption{\textbf{Quality and diversity comparison on GenAI-Bench}. Fine-tuning with \name captions improves global semantic fidelity and perceptual quality without reducing generation diversity.}
    \label{tab:quality-diversity}
    \begin{tabular}{@{}lccc@{}}
    \toprule
            & SDv1.5 & CC3M-FT & GAS-FT \\
    \midrule
    CLIPScore   & 0.3167 & 0.3196 & 0.3206 \\
    ImageReward & 0.2056 & 0.3842 & 0.3927 \\
    LPIPS & 0.7297 & 0.7356 & 0.7329 \\
    \bottomrule
    \end{tabular}
    }
\end{minipage}
\hfill 
\noindent
\begin{minipage}{0.48\linewidth}
    \centering
    {
    \caption{\textbf{Generalization to unseen compositions.}
On a 400-caption test set containing only unseen combinations of seen elements, the model fine-tuned with \name achieves the best compositional generalization.}
    \label{tab:unseen-composition}
    \begin{tabular}{@{}lccc@{}}
    \toprule
            & SDv1.5 & CC3M-FT & GAS-FT \\
    \midrule
    VQAScore     & 0.5823 & 0.6044 & 0.6109 \\
    CLIPScore    & 0.2876 & 0.2927 & 0.2938 \\
    ImageReward  & \-0.4861 & \-0.2602 & -0.2497 \\
    \bottomrule
    \end{tabular}
    }
\end{minipage}
\end{table}

\textbf{Fine-tuning with our synthetic captions can outperform fine-tuning with high-quality real-world image-caption data.} Our results show that fine-tuning with \name-generated synthetic data consistently outperforms CC3M-based fine-tuning across \vision tasks (Figure~\ref{general-bench}), achieving the highest gains with our highest-scoring selection strategy. This highlights \name's scalability and compositional diversity, enabling models to effectively capture complex scene structures.

In Table~\ref{tab:quality-diversity}, we further evaluate SDv1.5, the CC3M-finetuned model, and the model finetuned with \name captions on additional metrics from GenAI-Bench. Fine-tuning with \name yields higher CLIPScore and ImageReward while preserving LPIPS, demonstrating that our method not only strengthens compositional alignment but also improves global semantic fidelity and perceptual quality without reducing generation diversity.

In Table~\ref{tab:unseen-composition}, we additionally evaluate whether our self-improving framework enhances combinatorial generalization. We extract all objects, attributes, and relations from the CC3M fine-tuning data and retain the metadata sampled by \name. Using the same element set as in the fine-tuning data, we synthesize 200 CC3M-element-based and 200 \name-element-based captions while excluding all seen combinations, forming a 400-caption test set of unseen compositions. The model fine-tuned with \name achieves the highest VQAScore, CLIPScore, and ImageReward, indicating stronger compositional generalization than both SDv1.5 and the CC3M-finetuned baseline.
Additional experiment settings and results are in Appendix~\ref{app:appone}.


    

\section{Distilling targeted capabilities}

Although self-improving with \name shows clear advantages over high-quality real-world datasets, its efficiency is inherently limited by the model's own generation capabilities. To address this, we leverage the taxonomy and 
systematical generation capabilities within \name to identify specific strengths of proprietary models (\dalle), and distill these capabilities into open-source models. More details are in Appendix~\ref{app:apptwo}.

\begin{figure}[!b]
\centering
    \begin{subfigure}[t]{0.32\textwidth}
        \centering
        \includegraphics[width=\linewidth]{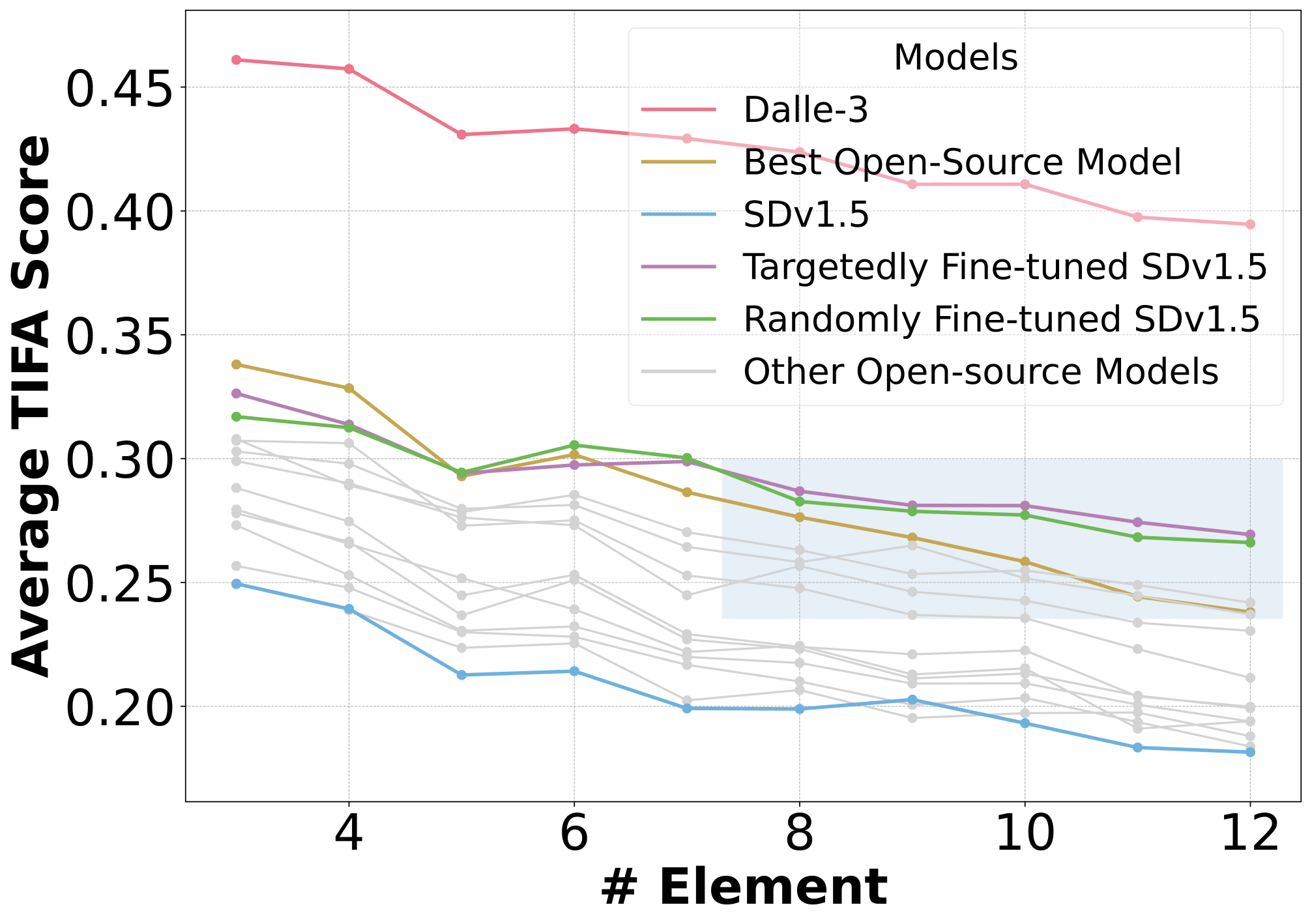}
        \caption{\textbf{Distilling compositionality from DaLL-E 3}: Model results on TIFA vs. total element numbers in captions in 10K general \name captions. ("Best Open-Source Model" refers to Flux.1-schnell)}
        \label{fig:10K_tifa}
    \end{subfigure}
    ~
    \begin{subfigure}[t]{0.32\textwidth}
        \centering
        \includegraphics[width=\linewidth]{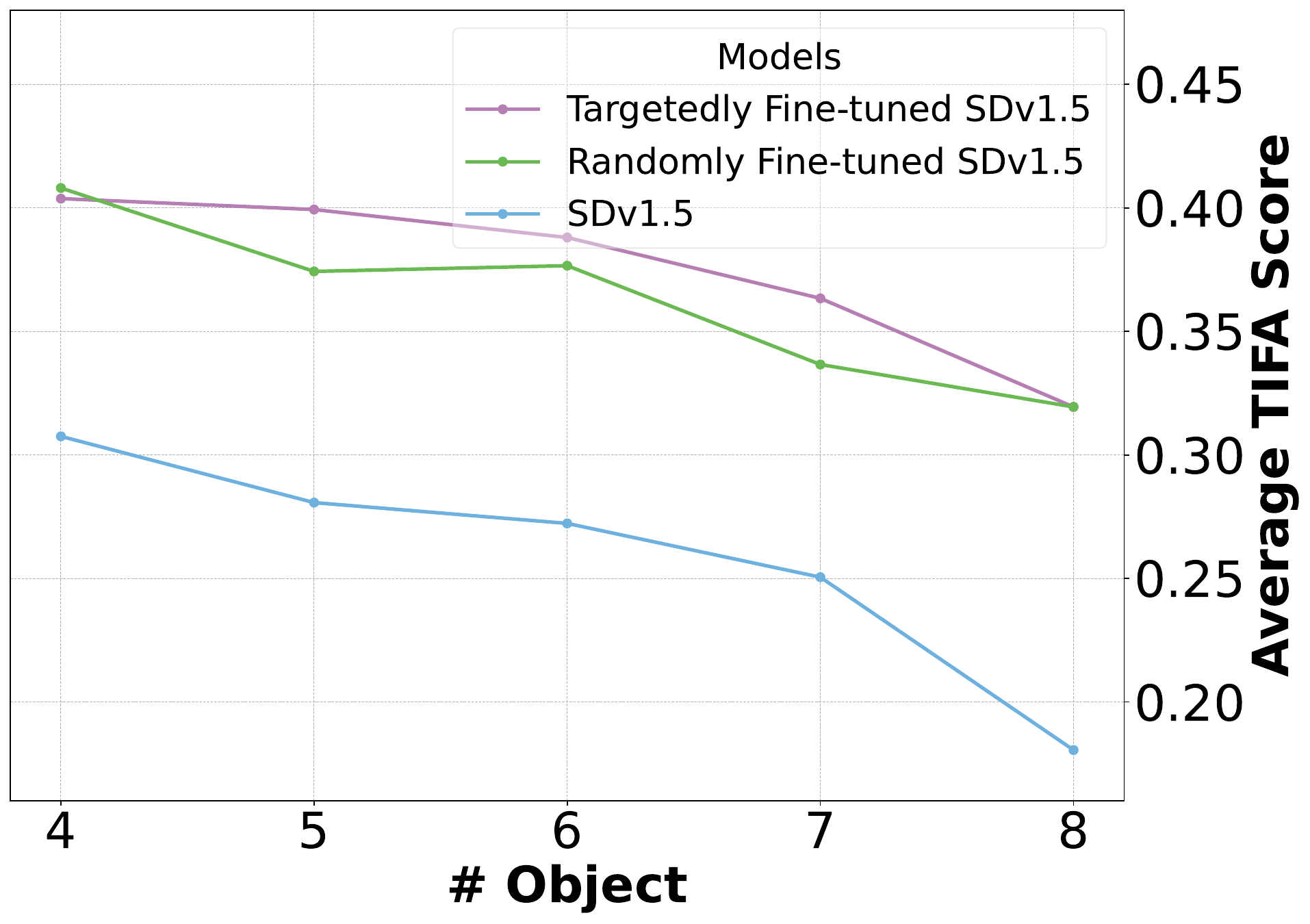}
        \caption{
        \textbf{Distilling compositionality from DaLL-E 3}: Model results on TIFA vs. total element numbers in captions in 1K multi-object \name captions.}
        \label{fig:1K_tifa}
    \end{subfigure}
    ~
    \begin{subfigure}[t]{0.32\textwidth}
        \centering
            \includegraphics[width=\linewidth]{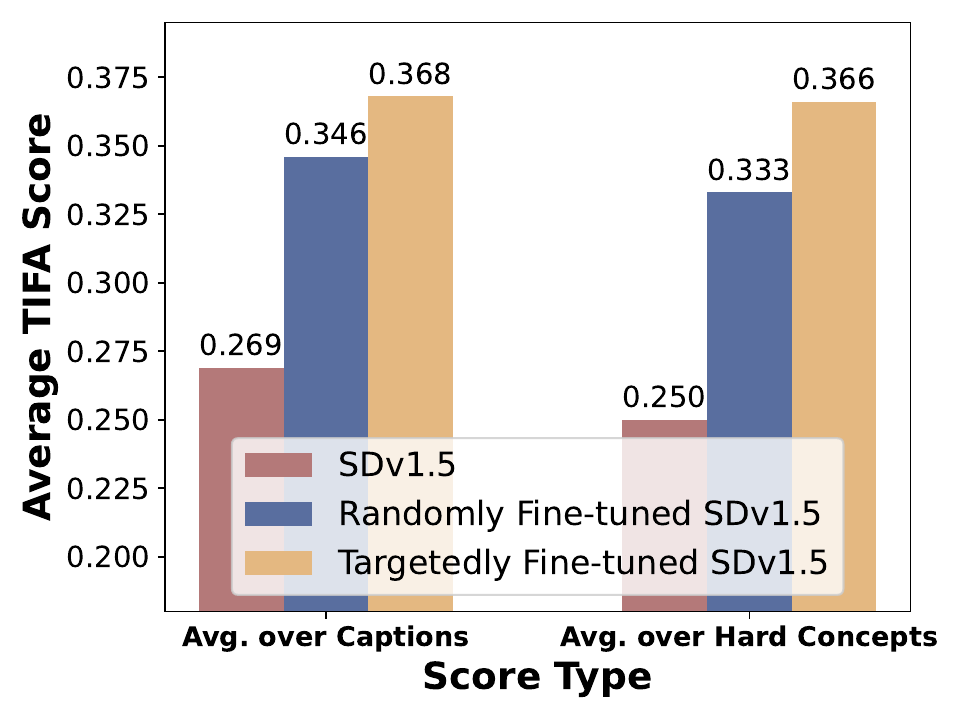}
        \caption{\textbf{Distilling hard concepts understanding from DALL-E 3}: Models' average \tifa performance over captions and hard concepts in 1K hard concepts \name captions.}
        \label{fig:complex_concept}
    \end{subfigure}
    \vspace{-0.5em}
    \caption{\textbf{Results for Distilling Capabilities}. The left two figures show the results for \textbf{Distilling compositionality}, while the rightmost figure shows the results for \textbf{Distilling hard concepts understanding from DALL-E 3}.}
    \label{fig:three_tifa_scores}
\end{figure}

We evaluate multiple models using captions controllably generated by \name and observe that \dalle achieves \tifa \textbf{1.5} to \textbf{2} times higher than those of other models. As shown in Figure~\ref{fig:10K_tifa}, when comparing \tifa across captions with varying numbers of elements (objects, relations, and attributes), \dalle \textbf{counterintuitively} maintains consistent performance regardless of element count. The performance of other models declines as the element count increases, which aligns with expected compositional challenges. We suspect that these differences are primarily due to \dalle's advanced capabilities in compositionality and \textbf{understanding hard concepts}, which ensures high faithfulness across diverse combinations of element types and counts.



\begin{figure*}[!t]
    \centering
    \includegraphics[width=1\textwidth]{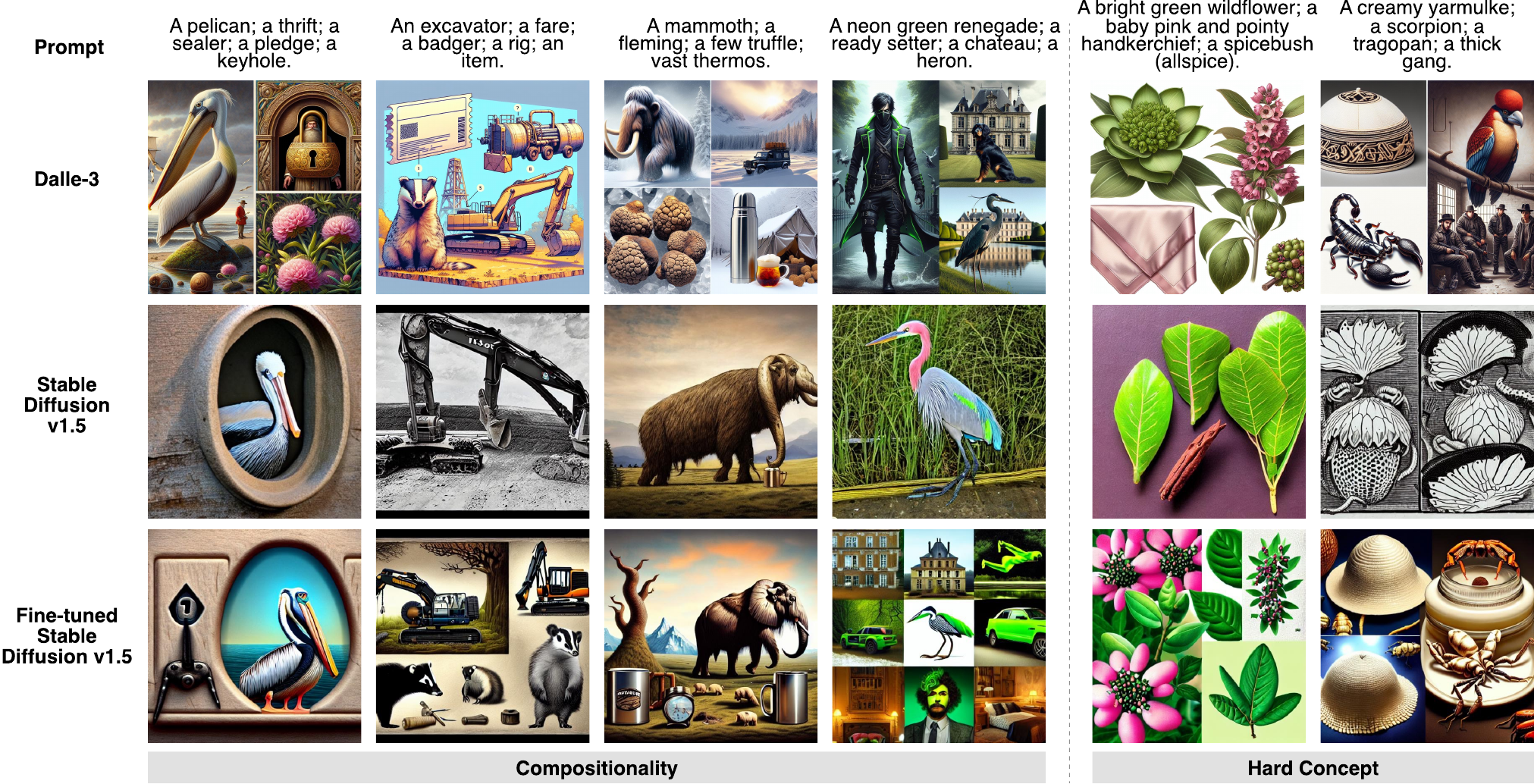}
    \caption{\textbf{Examples for Distilling Capabilities.} Examples of images generated by \dalle, the original \sdonefive, and the fine-tuned versions. The left four captions demonstrate fine-tuning with multi-object captions generated by \name for better compositionality, while the right two columns focus on understanding hard concepts.}
    \label{multiobj_example}
    \vspace{-1em}
\end{figure*}


\noindent\textbf{Distilling compositionality from DaLL-E 3.}
When analyzing images generated from our synthetic captions, we find that \dalle tends to produce straightforward combinations of multiple objects (Figure~\ref{multiobj_example}). In contrast, open-source models like \sdonefive often omit objects from the captions, despite being capable of generating each one individually. This difference suggests that \dalle may benefit from training data emphasizing multi-object presence, even without detailed layout or object interaction. Such training likely underpins \dalle's stronger performance on metrics like \tifa and \vqascore that prioritize object inclusion.
To effectively distill these compositional abilities into \sdonefive, we employ \name for targeted synthesis of 778 multi-object captions, paired with images generated by \dalle, for finetuning \sdonefive. 

\noindent\textbf{Distilling hard concepts understanding from DaLL-E 3.}
Figure \ref{multiobj_example} shows that \dalle is capable not only of handling multi-object generation but also of understanding and generating rare and hard concepts, such as a specific species of flower. We attribute this to its training with proprietary real-world data.
Using the taxonomy of \name, we evaluate both models on 10K \name captions that broadly cover the taxonomy. For each concept, we gather all captions in which it appears and average their generation scores to obtain a concept-level score for each model. Comparing these concept-level scores lets us identify the 81 concepts where \sdonefive shows the largest gap relative to \dalle; the full list is provided in Appendix~\ref{app:apptwo}.
For distilling, we increase the sampling frequency of these hard concepts and generate 778 captions incorporating these hard concepts with other elements, and use \dalle to produce corresponding images.

\noindent\textbf{Baselines.}
For the baseline, we randomly synthesize 778 captions using \name paired with \dalle-generated images to fine-tune the model. To evaluate model improvements, we generate another 1K multi-object captions and 1K hard-concept captions separately.

\noindent\textbf{Targeted caption synthesis via \name enables effective distillation of compositional abilities and hard concept understanding.}
We analyze images generated by \sdonefive before and after fine-tuning on high-complexity captions (Figure~\ref{multiobj_example}). 
Surprisingly, with fewer than 1K LoRA fine-tuning steps, \sdonefive effectively learns \dalle’s capability to arrange and compose multiple objects within a single image. Quantitatively, Figure~\ref{fig:1K_tifa} shows a 10\% improvement in \tifa after targeted fine-tuning, surpassing the performance of the randomly fine-tuned model. On a broader set of 10K \name-generated captions, the targeted fine-tuned model consistently outperforms randomly fine-tuned and original counterparts across complex scenes (Figure~\ref{fig:10K_tifa}). These results confirm not only the effectiveness but also the scalability and efficiency of \name. Also, the results in Figure~\ref{fig:complex_concept} show that our targeted fine-tuning with hard concepts leads to improved model performance, reflected in higher average scores across captions and increased scores for each challenging concept.

\section{Reinforcement learning with a synthetic reward function}

Reinforcement Learning with Human Feedback (RLHF) has become an increasingly popular fine-tuning strategy in text-to-image generation~\cite{gong2025seedream,jiang2025t2ir1,wang2025simplear}. However, defining an effective reward model that accurately captures semantic alignment for text-to-image generation remains an open challenge.
Existing reward models like CLIP offer only coarse-grained image-text similarity signals, which fall short in assessing compositional correctness and lack interpretability. Alternative approaches have explored using visual question answering (VQA) as a proxy for evaluating semantic alignment, aiming for finer-grained assessments, yet require either labor-intensive datasets with dense annotations or large volumes of contextually relevant questions via advanced LLMs. Leveraging its structured scene graph synthesis capabilities, \name offers a scalable alternative by producing exhaustive semantic queries with negligible overhead, enabling low-cost, compositional reward modeling (Sec~\ref{step5}).


\begin{figure*}[!t]
    \centering
    \includegraphics[width=1\textwidth]{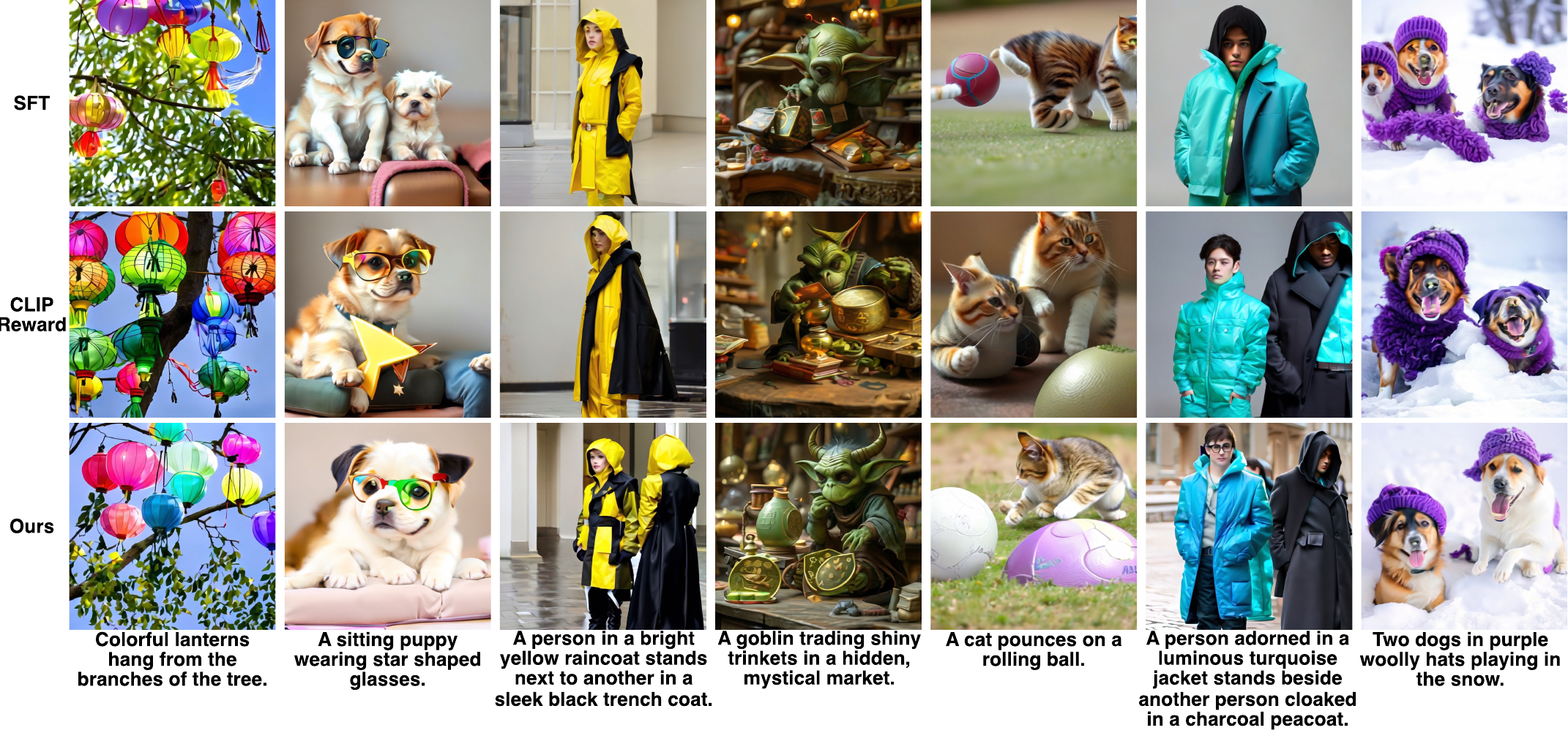}
    \caption{\textbf{Comparison of generated images.} Our reward model enables image generation with better semantic alignment, realism, and visual quality than baselines.}
\end{figure*}

\noindent\textbf{Experiment setup.}
Building on this scene graph-based reward modeling strategy, we adopt Group Relative Policy Optimization (GRPO) as our reinforcement learning algorithm. We fine-tune the SimpleAR-0.5B-SFT model for one epoch using 10K captions generated by \name, each paired with their scene graph-derived QA sets. For reward evaluation, we use Qwen2.5-VL-3B, a lightweight open-source vision-language model, to answer these QA pairs given the model-generated images. The reward is computed as the accuracy across all questions. This fine-grained, scene graph-aligned reward provides precise feedback on compositional faithfulness. As a baseline, we compare against SimpleAR-0.5B-RL, trained with CLIP-based rewards on 11K captions from real world datasets for one epoch. We evaluate our scene graph-based reward model on three benchmarks: DPG-Bench~\cite{hu2024ella}, GenEval~\cite{ghosh2023geneval}, and GenAI-Bench~\cite{li2024genai}. More details are in Appendix~\ref{app:appzero}.


\noindent\textbf{\name rewards outperform CLIP.}
As shown in Table~\ref{tab:reward_exp}, our method outperforms both SFT and CLIP-RL models and achieves a significant improvement, demonstrating superior compositional faithfulness driven by explicit scene graph rewards. Importantly, this performance gain is directly enabled by the \name engine, which constructs explicit scene graphs to generate compositional captions. \name provides a structured and cognitively aligned visual representation, from which we derive exhaustive QA pairs with minimal additional cost. Combined with lightweight VLM judge, this approach offers a scalable, low-cost solution for semantic-level reward modeling.

\begin{table}[h]
\caption{Evaluation on the DPG, GenEval and GenAI benchmark. GRPO training with our reward model outperforms both SFT baseline and CLIP-RL models. TO: two objects, P: position, CA: color attribute.}
\label{tab:reward_exp}
\resizebox{\columnwidth}{!}{%
\begin{tabular}{@{}llcccccccccc@{}}
\toprule
\multirow{2}{*}{Method} &
  \multicolumn{3}{c}{DPG-Bench} &
  \multicolumn{4}{c}{GenEval} &
  \multicolumn{3}{c}{GenAI-Bench} \\ \cmidrule(lr){2-4} \cmidrule(lr){5-8}  \cmidrule(lr){9-11} 
 &
  Global &
  Relation &
  Overall &
  TO &
  P &
  CA &
  Overall &
  Basic &
  Advanced &
  All \\ \midrule
\begin{tabular}[c]{@{}l@{}}SimpleAR-0.5B-SFT\end{tabular} &
  85.02 &
  86.59 &
  78.48 &
  0.73 &
  0.22 &
  0.23 &
  0.53 &
  0.74 &
  0.60 &
  0.66 \\
\begin{tabular}[c]{@{}l@{}}SimpleAR-0.5B-RL (Clip)\end{tabular} &
  86.64 &
  88.51 &
  79.66 &
  \textbf{0.82} &
  0.26 &
  \textbf{0.38} &
  0.59 &
  \textbf{0.75} &
  0.60 &
  0.67 \\
\textbf{\begin{tabular}[c]{@{}l@{}}SimpleAR-0.5B-RL (Ours)\end{tabular}} &
  \textbf{88.46} &
  \textbf{90.13} &
  \textbf{80.50} &
  0.81 &
  \textbf{0.31} &
  \textbf{0.38} &
  \textbf{0.61} &
  \textbf{0.75} &
  \textbf{0.61} &
  \textbf{0.68} \\ \bottomrule
\end{tabular}%
}
\end{table}

\section{Improving generated-content detection}


Advances in \vision underscore the need for effective content moderation~\cite{pei2024deepfake}. Major challenges include the lack of high-quality and diverse datasets and the difficulty of generalizing detection across models \vision~\cite{wang2024deepfake,kaur2024deepfake}. \name addresses these issues by enabling scalable, systematical generation of compositional captions, increasing the diversity and volume of synthetic data. This approach enhances existing datasets by compensating for their limited scope-from realistic to imaginative-and variability.

\textbf{Experiment setup.} To demonstrate \name's effectiveness in training generated content detectors, we used the $D^3$ dataset~\cite{elsa} as a baseline. We sampled 5K captioned real and SDv1.4-generated image pairs from $D^3$ and generated 5K additional images with \name captions.
We trained a ViT-T~\cite{tinyvit} model with a single-layer linear classifier and compared models trained with samples solely from $D^3$ against those trained with samples from both \name and $D^3$.

\begin{figure}[h]
\centering
    \begin{subfigure}[t]{0.32\textwidth}
        \centering
        \includegraphics[width=\linewidth]{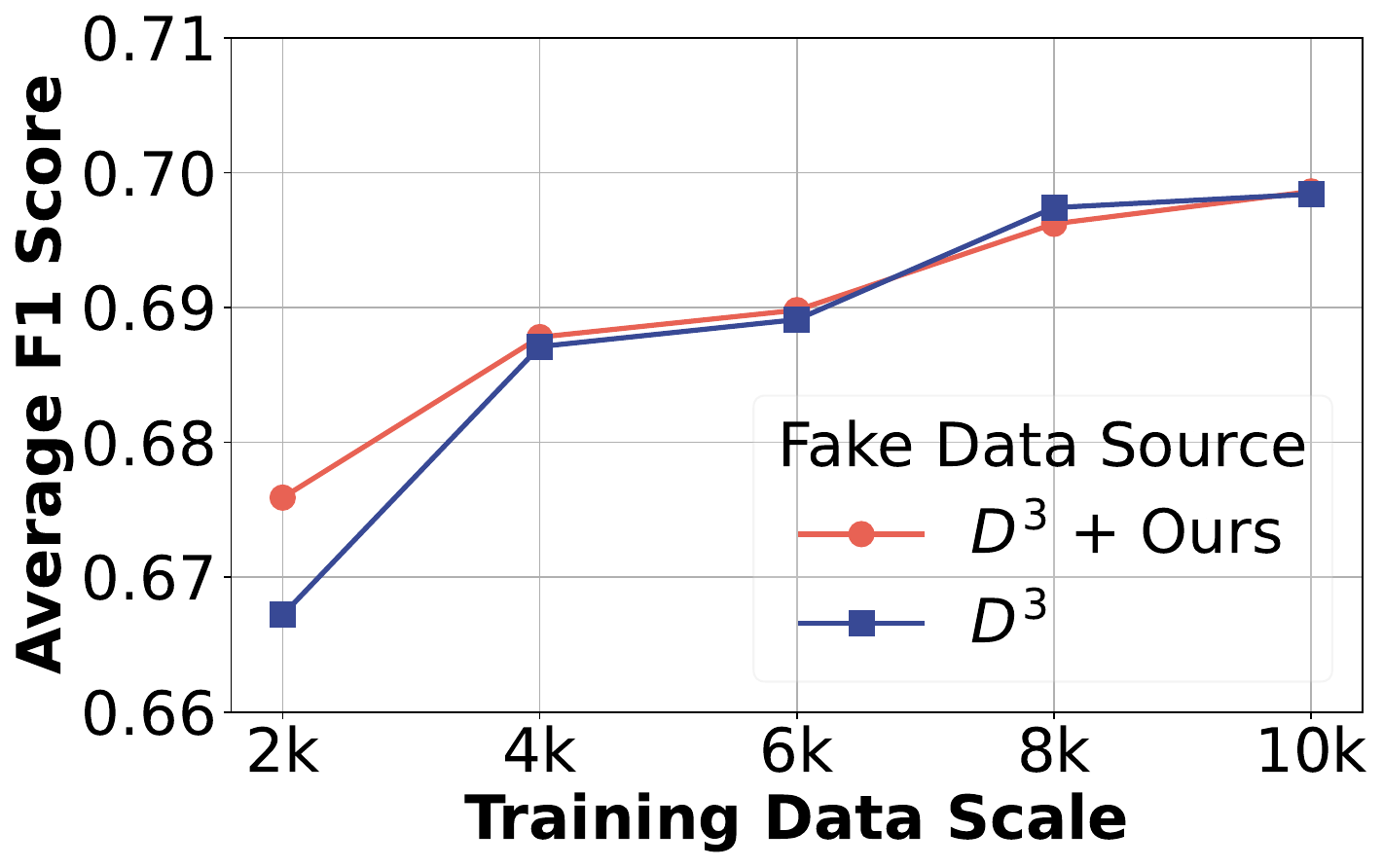}
        \caption{\textbf{In-domain testing (Same Model - SD v1.4)}: Detection results on images generated by SD v1.4 using the GenImage dataset.}
        \label{fig:in_sd4}
    \end{subfigure}
    ~
    \begin{subfigure}[t]{0.32\textwidth}
        \includegraphics[width=\linewidth]{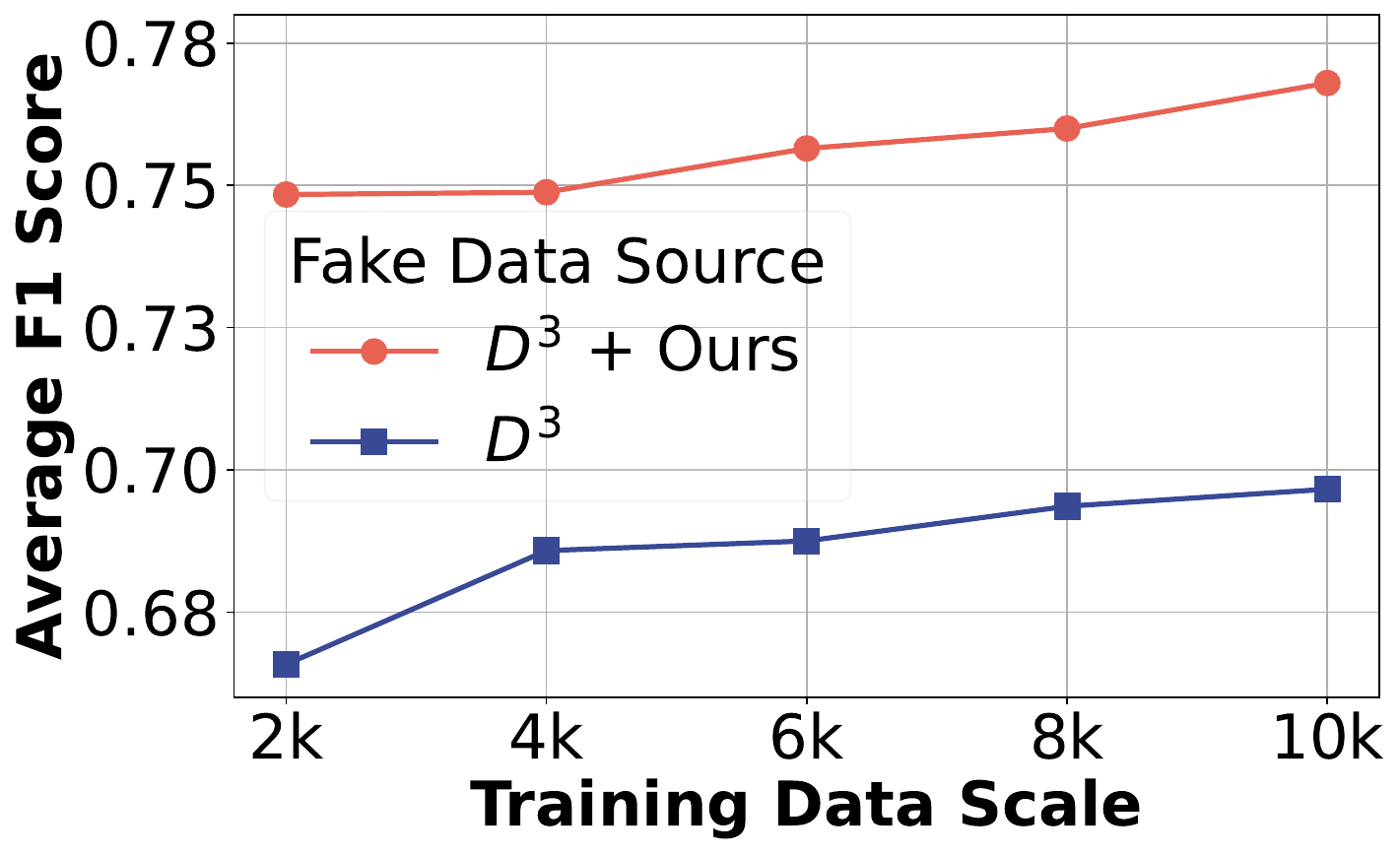}
        \caption{\textbf{In domain testing (cross-model)}:Average detection results on images generated by multiple models using our captions.}
        \label{fig:in_caption}
    \end{subfigure}
    ~
    \begin{subfigure}[t]{0.32\textwidth}
        \includegraphics[width=\linewidth]{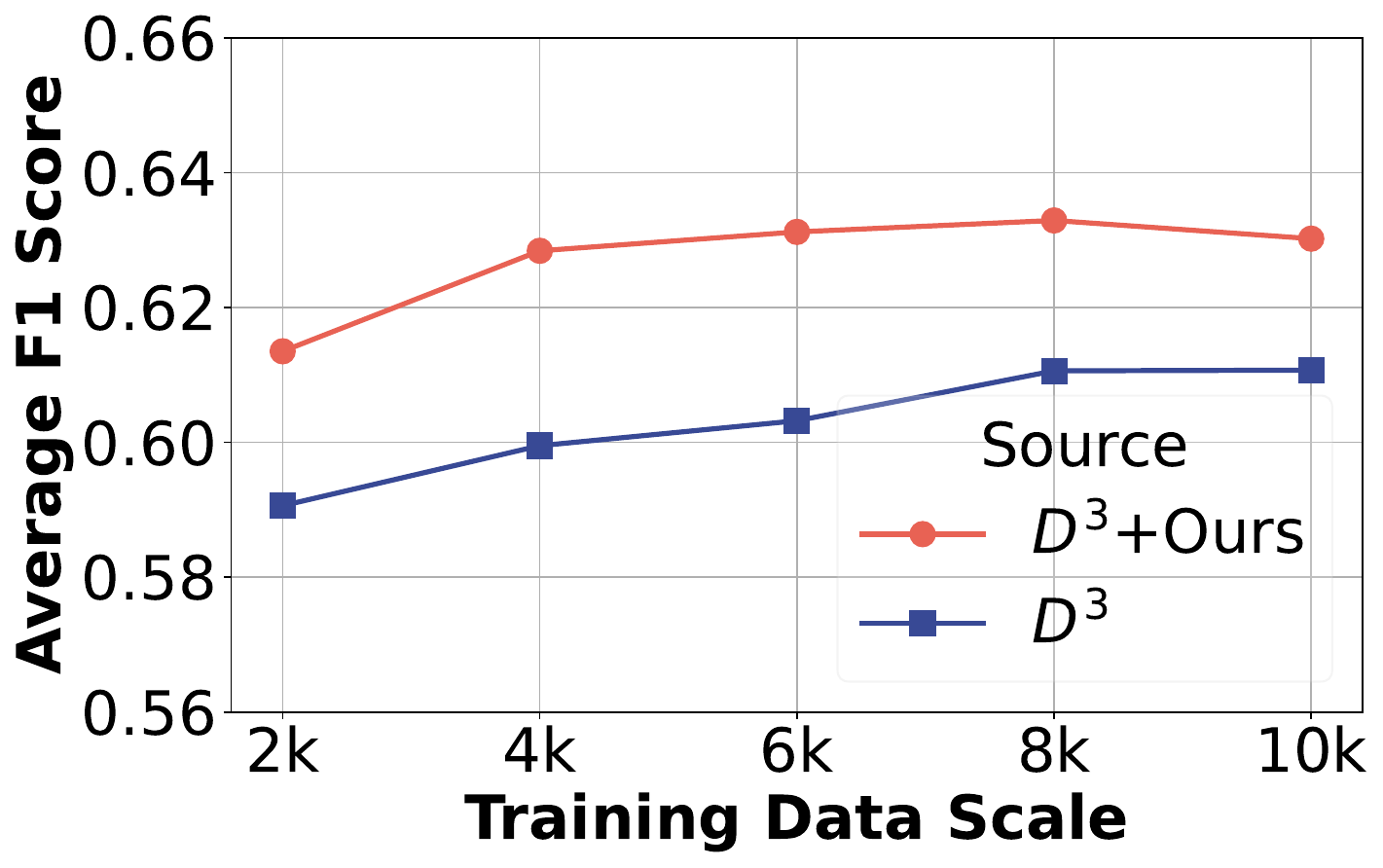}
        \caption{\textbf{Out of domain}: Average detection results on images generated by multiple models using captions from the GenImage dataset.}
        \label{fig:out}
    \end{subfigure}
    \caption{\textbf{Results for Application 4: Generated content detector}. Comparison of detection performance across different data scales using $D^3$ alone versus the combined $D^3$ + \name training set in cross-model and cross-dataset scenarios.}

    \label{fig:deepfake}
\end{figure}

\textbf{\name improves generated content detectors. }We evaluate the detector's generalization on the GenImage~\cite{genimage} validation set and images generated using \name captions. Figure~\ref{fig:deepfake} demonstrates that combining \name-generated images with real-world captioned images consistently enhances detection performance, particularly across cross-model scenarios and diverse visual scenes. More details are in Appendix~\ref{app:appthree}.

    

\section{Comprehensive evaluation with \name}
Beyond showcasing \name in model training, we also show that \name is a valuable resource for comprehensive and compositional evaluation. Specifically, we synthesize 10K captions for text-to-image, 10K for text-to-video, and 1K for text-to-3D, covering diverse scene structures and content topics. We evaluate 12 text-to-image, 9 text-to-video, and 5 text-to-3D models. Evaluations combine \name synthetic scene graphs with existing metrics (e.g., CLIP Score~\cite{Chakrabarty2023LearningTF}, VQA Score~\cite{Lin2024EvaluatingTG}, TIFA Score~\cite{hu2023tifaaccurateinterpretabletexttoimage, Cho2023DavidsonianSG}) to assess semantic similarity, faithfulness, and human preference alignment.
Our key findings include: (1) DiT-backbone text-to-image models align more closely with input captions than UNet-backbone models. (2) Text-to-video models struggle with balancing dynamics and consistency, while both text-to-video and text-to-3D models show notable gaps in human preference alignment. Except for aggregating quantitative results, we also leverage \name’s controllable captioning to evaluate models on fine-grained factors: perplexity, scene complexity, commonsense reasoning, and content category variation for case study.

Overall, \name yields stable, human-aligned rankings across T2I/T2V/T2-3D. Through broad, controllable coverage of objects, attributes, relations, and categories, it serves as a compositional stress test that reliably exposes plausibility gaps, category brittleness, and long-tail concept failures in current models (see Appendix~\ref{appendix:eval}).
\section{Related work}
\paragraph{\vision models.}
\image advances are driven by diffusion models and LLMs. Some open-source models~\cite{stablediffusion2,SDXL,playground,wurstchen,Rombach_2022_CVPR,DeepFloydIF} use UNet backbones to refine images iteratively. In parallel, Diffusion Transformers (DiTs) architectures\cite{stablediffusion3,pixartalpha,pixartsigma,blackforestlabs2024flux1} have emerged as a better alternative in capturing long-range dependencies and improving coherence. Proprietary models like DALL-E 3~\cite{dalle3} and Imagen 3~\cite{baldridge2024imagen} still set the state-of-the-art. Based on \image method, \video models typically utilize time-aware architectures to ensure temporal coherence across frames~\cite{animatediff,animatelcm,text2videozero,modelscope,freeinit,videocrafter2,cogvideox,opensora}. In \threed, recent proposed models~\cite{prolificdreamer,dreamfusion,sjc,latentnerf,magic3d} integrate the diffusion models with Neural Radiance Fields (NeRF) rendering to generate diverse 3D objects. Recent studies~\cite{wang2025simplear, jiang2025t2ir1,wu2024janus,chen2025janus} have also explored the integration of image generation into a unified multimodal language model (MLM) framework based on auto-regressive transformer architectures, demonstrating promising improvements in both performance and efficiency.


\paragraph{Synthetic captions for \vision.}
Captions for \vision models vary greatly in diversity, complexity, and compositionality. This variation makes it challenging and costly to collect large-scale and diverse captions written by humans. Consequently, synthetic captions have been widely used for both training~\cite{Lian2023LLMgroundedDE, Sun2023DreamSyncAT, Li2024SELMALA, Zhao2024EvolveDirectorAA, Sun2024T2VCompBenchAC, Park2021BenchmarkFC, Wen2023ImprovingCT, Wu2024ConceptMixAC} and evaluation purposes~\cite{Huang2023T2ICompBenchAC}. For example, training methods like LLM-Grounded Diffusion~\cite{Lian2023LLMgroundedDE} leverage LLM-generated captions to enhance the model's understanding and alignment with human instruction. For evaluation, benchmarks such as T2I-CompBench~\cite{Huang2023T2ICompBenchAC} and T2V-CompBench~\cite{Sun2024T2VCompBenchAC} utilize benchmarks generated by LLMs. However, LLMs are hard to control and may introduce exhibit systematic bias. In this work, we propose a programmatic scene graph-based data engine that can generate infinitely diverse captions for improving \vision models.

\paragraph{Finetuning techniques for \vision.}
To accommodate the diverse applications and personalization needs in text-to-vision models, numerous fine-tuning techniques have been developed. LoRA~\cite{Hu2021LoRALA} reduces fine-tuning costs via low-rank weight updates, while Textual Inversion~\cite{Mokady2022NulltextIF, Gal2022AnII} introduces new word embeddings for novel concepts without altering core parameters.
DreamBooth~\cite{Ruiz2022DreamBoothFT} adapts models to specific subjects or styles using a few personalized images, and DreamSync~\cite{Sun2023DreamSyncAT} enables models to self-improve by learning from their own high-quality outputs. Recently, RLHF~\cite{wang2025simplear,gong2025seedream,jiang2025t2ir1} in \vision has shown promise as an efficient fine-tuning strategy. In this work, we use several fine-tuning techniques with \name to improve \vision models.
\section{Conclusion}

We present \name, a system leveraging scene graph programming to generate diverse and compositional synthetic captions for \vision tasks. It extends beyond existing real-world caption datasets to include comprehensive scenes and even implausible scenarios. To demonstrate the effectiveness of \name, we explore four applications: (1) self-improvement by iteratively optimizing models, (2) distillation of proprietary model strengths into open-source models, (3) a scene-graph-based efficient reward model within the GRPO, and (4) robust content moderation with diverse synthetic data. \name highlights the importance of synthetic data in improving \vision, and addresses the need to systematically define and scalably produce the space of visual scenes.

\newpage

{
    \small
    \bibliographystyle{unsrtnat}
    \bibliography{main}
}


\appendix
\newpage

\section{Evaluating \vision models with \name}
\label{appendix:eval}
\subsection{Experiment Settings}

\noindent\textbf{Models.} We conduct experiments on 12 \textit{Text-to-image} models~\cite{DeepFloydIF,SDXL,stablediffusion2,playground,wurstchen,stablediffusion3,pixartalpha,pixartsigma,blackforestlabs2024flux1,dalle3}, 9 \textit{Text-to-Video} models~\cite{modelscope,zeroscope,text2videozero,animatediff,animatelcm,freeinit,opensora,cogvideox,videocrafter2}, and 5 \textit{Text-to-3D} models~\cite{dreamfusion,magic3d,sjc,prolificdreamer,latentnerf}. 
\begin{itemize}
    \item[$\bullet$] For \image, we select a range of open-source models, including those utilizing UNet backbones, such as \deepfloyd~\cite{DeepFloydIF}, \sdtwoone~\cite{stablediffusion2}, \sdxl~\cite{SDXL}, \playground~\cite{playground}, and \wuerstchen~\cite{wurstchen}, as well as models with DiT backbones, including \sdthree~\cite{stablediffusion3}, \pixartalpha~\cite{pixartalpha}, \pixartsigma~\cite{pixartsigma}, \fluxschnell~\cite{blackforestlabs2024flux1}, \fluxdev~\cite{blackforestlabs2024flux1}, and FLUX 1. Closed-source models, such as \dalle~\cite{dalle3} and \flux~\cite{blackforestlabs2024flux1}, are also assessed to ensure a comprehensive comparison. All models are evaluated at a resolution of 1024 × 1024 pixels.
    
    \item[$\bullet$] For \video, we select nine open-source models: \modelscope~\cite{modelscope}, \zeroscope~\cite{zeroscope}, \texttovideozero~\cite{text2videozero}, \cogvideox~\cite{cogvideox}, \VideoCrafter~\cite{videocrafter2}, \AnimateLCM~\cite{animatelcm}, \AnimateDiff~\cite{animatediff}, \FreeInit~\cite{freeinit}, and \OpenSora~\cite{opensora}. We standardize the frame length to 16 across all video models for fair comparisons.

    \item[$\bullet$] For \threed, we evaluate five recently proposed models: \SJC~\cite{sjc}, \DreamFusion~\cite{dreamfusion}, \Magic~\cite{magic3d}, \Latentnerf~\cite{latentnerf}, and \ProlificDreamer~\cite{prolificdreamer}. We employ the implementation and configurations provided by ThreeStudio~\cite{threestudio} and generate videos by rendering from 120 viewpoints. To accelerate inference, we omit the refinement stage. For \Magic and \DreamFusion, we respectively use \deepfloyd and \sdtwoone as their 2D backbones.

\end{itemize}

\begin{figure}[h]
    \centering
    \includegraphics[width=0.68\linewidth]{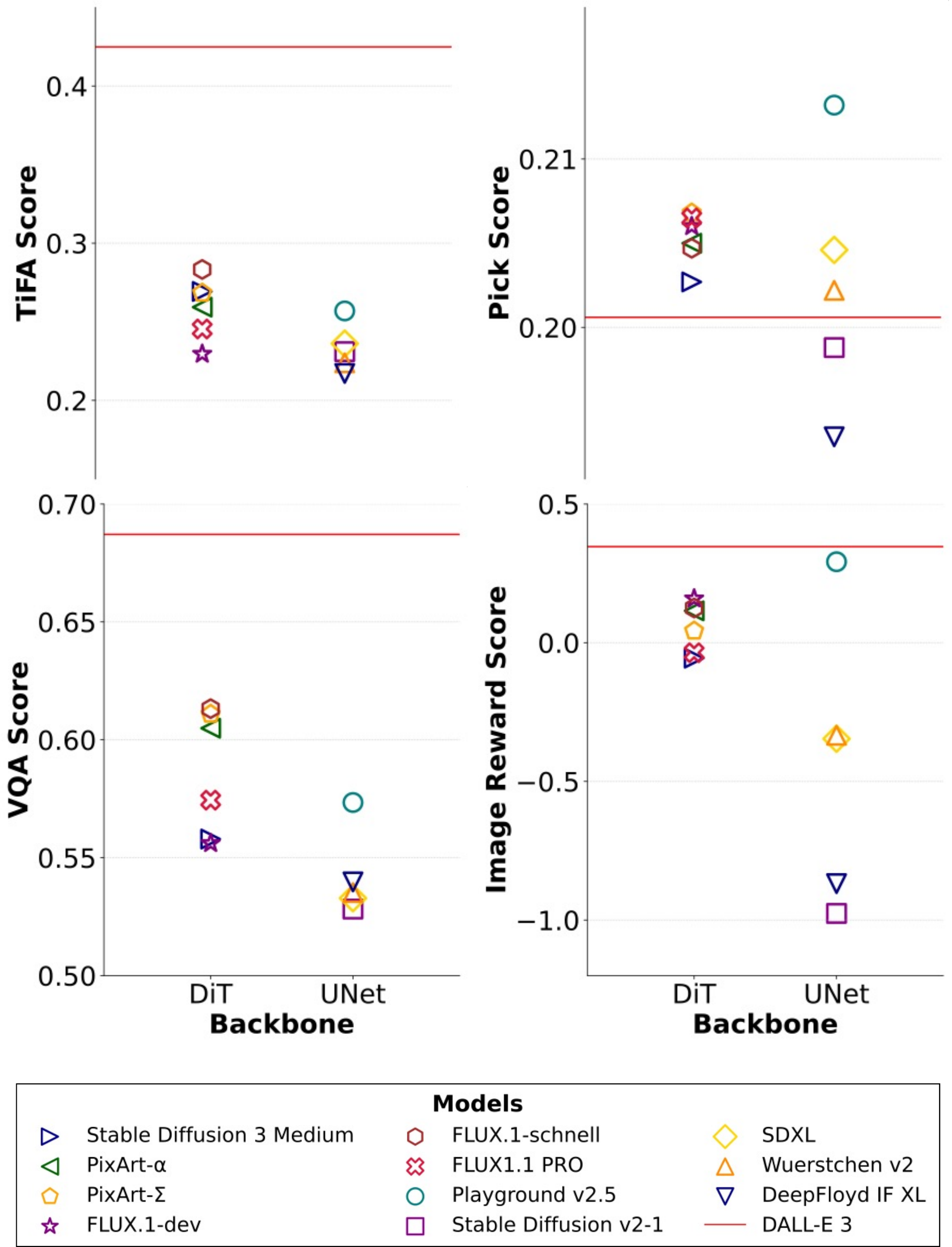}
    \caption{Comparative evaluation of \image models across different backbones (DiT and UNet) using multiple metrics: \tifa, \pickscore, \vqascore, and \imagereward.}
    \label{fig:image_overall}
\end{figure}

\noindent\textbf{Metrics.} Across all \vision tasks, we use \clipscore~\cite{Chakrabarty2023LearningTF} (semantic similarity), \vqascore~\cite{Lin2024EvaluatingTG} (faithfulness), \tifa~\cite{hu2023tifaaccurateinterpretabletexttoimage, Cho2023DavidsonianSG} (faithfulness), \pickscore~\cite{kirstain2023pickapicopendatasetuser} (human preference), and \imagereward~\cite{xu2023imagereward} (human preference) as general metrics:
\begin{itemize}
    \item[$\bullet$] \clipscore: Assesses semantic similarity between images and text.
    \item[$\bullet$] \vqascore and \tifa: Evaluate faithfulness by generating question-answer pairs and measuring answer accuracy from images.
    \item[$\bullet$] \pickscore and \imagereward: Capture human preference tendencies.
\end{itemize}
We also use metrics in VBench~\cite{huang2024vbench} to evaluate \video models on fine-grained dimensions, such as consistency and dynamics, providing detailed insights into video performance.

For \video and \threed tasks:
\begin{itemize}
    \item[$\bullet$] We calculate \clipscore, \pickscore, and \imagereward on each frame, then average these scores across all frames to obtain an overall video score.
    \item[$\bullet$] For \vqascore and \tifa, we handle \video and \threed tasks differently:
        \begin{itemize}
            \item[$\circ$] In \video tasks, we uniformly sample four frames from the 16-frame sequence and arrange them in a 2 × 2 grid image.
            \item[$\circ$] For \threed tasks, we render images at 45-degree intervals from nine different viewpoints and arrange them in a 3 × 3 grid.
        \end{itemize}
\end{itemize}

This sampling approach optimizes inference speed without affecting score accuracy~\cite{Lin2024EvaluatingTG}.

\noindent\textbf{Synthetic captions.}
We evaluate our \image and \video models on 10K randomly generated captions, with scene graph complexity ranging from 3 to 12 and scene attributes from 0 to 5, using unrestricted metadata. The captions exhibit an average graph degree of 1.15, with values spanning from 0.0 to 0.8. The mean number of connected components per scene graph is 3.51, ranging from 1 to 11. For \threed models, due to their limitations in handling complex captions and time-intensive generation, we restrict scene graph complexity to 1-3, scene attributes to 0-2, and evaluate on 1K captions.

\subsection{Overall results}
We evaluate \image, \video, and \threed models on \name. 

\begin{table}[!h]
\centering
\caption{Overall performance of \image models over 10K \name captions. \textsuperscript{†}Evaluated on a 1K caption subset due to inference cost constraints.}

\small
\resizebox{\columnwidth}{!}{%
\begin{tabular}{lccccc}
\toprule
\textbf{Model} & \textbf{clip score} & \textbf{pick score} & \textbf{vqa score} & \textbf{tifa score} & \textbf{image reward score} \\
\midrule
Playground v2.5~\cite{playground}               & 0.2581 & 0.2132 & 0.5734 & 0.2569 & 0.2919 \\
Stable Diffusion v2-1~\cite{stablediffusion2}         & 0.2453 & 0.1988 & 0.5282 & 0.2310 & -0.9760 \\
SDXL~\cite{SDXL}           & 0.2614 & 0.2046 & 0.5328 & 0.2361 & -0.3463 \\
Wuerstchen v2~\cite{wurstchen}                 & 0.2448 & 0.2022 & 0.5352 & 0.2239 & -0.3339 \\
DeepFloyd IF XL~\cite{DeepFloydIF}               & 0.2396 & 0.1935 & 0.5397 & 0.2171 & -0.8687 \\\hline
Stable Diffusion 3 Medium~\cite{stablediffusion3}     & 0.2527 & 0.2027 & 0.5579 & 0.2693 & -0.0557 \\
PixArt-$\alpha$~\cite{pixartalpha}               & 0.2363 & 0.2050 & 0.6049 & 0.2593 & 0.1149 \\
PixArt-$\Sigma$~\cite{pixartsigma}               & 0.2390 & 0.2068 & 0.6109 & 0.2683 & 0.0425 \\
FLUX.1-dev~\cite{blackforestlabs2024flux1}                  & 0.2341 & 0.2060 & 0.5561 & 0.2295 & 0.1588 \\
FLUX.1-schnell~\cite{blackforestlabs2024flux1}              & 0.2542 & 0.2047 & 0.6132 & 0.2833 & 0.1251 \\\hline
FLUX1.1 PRO~\cite{blackforestlabs2024flux1}\textsuperscript{†}                   & 0.2315 & 0.2065 & 0.5744 & 0.2454 & -0.0361 \\
Dalle-3~\cite{dalle3}                       & 0.2518 & 0.2006 & 0.6871 & 0.4249 & 0.3464 \\
\bottomrule
\end{tabular}%
}
\label{table:app-overall-image-perf}
\end{table}


\noindent\textbf{\image results. (Figure~\ref{fig:image_overall}, Table~\ref{table:app-overall-image-perf})}

\begin{enumerate}
    \item DiT-backbone models outperform UNet-backbone models on \vqascore and \tifa, indicating greater faithfulness and comprehensiveness to input captions.
    \item Despite using a UNet architecture, \playground achieves higher \pickscore and \imagereward scores than other open-source models. We attribute this to \playground’s alignment with human preferences achieved during training.
     \item The closed-source model \dalle maintains a significant lead in \vqascore, \tifa, and \imagereward, demonstrating strong faithfulness and alignment with captions across generated content.
\end{enumerate}

\noindent\textbf{\video results. (Table~\ref{overall-video-perf},\ref{overall-video-perf-vbench})}

\begin{table}[htbp]
\centering
\caption{Overall performance of open-source \video models over 10K \name captions.\colorbox[RGB]{255, 200, 200}{Red Cell} is the highest score. \colorbox[RGB]{255, 255, 200}{Yellow Cell} is the second highest score.\textsuperscript{†}Close-source models are evaluated on a 1K caption subset due to high inference cost.}
\resizebox{\columnwidth}{!}{%
\begin{tabular}{lccccc}
\toprule
\textbf{Model} & \textbf{clip score} & \textbf{pick score} & \makecell{\textbf{image reward}\\\textbf{score}} & \textbf{VQA score} & \textbf{TiFA score} \\
\midrule
VideoCraft2~\cite{videocrafter2}          & 0.2398 & 0.1976 & -0.4202 & 0.5018 & 0.2466 \\
AnimateLCM~\cite{animatelcm}           & 0.2450 & \cellcolor[RGB]{255, 200, 200}0.1987 & -0.5754 & 0.4816 & 0.2176 \\
AnimateDiff~\cite{animatediff}          & \cellcolor[RGB]{255, 200, 200}0.2610 & 0.1959 & -0.7301 &  0.5255 & 0.2208 \\
Open-Sora 1.2~\cite{opensora}          & 0.2259 & 0.1928 & -0.6277 & \cellcolor[RGB]{255, 255, 200}0.5519 & 0.2414 \\
FreeInit~\cite{freeinit}           & \cellcolor[RGB]{255, 255, 200}0.2579 & 0.1950 & -0.9335 & 0.5123 & 0.2047 \\
ModelScope~\cite{modelscope}           & 0.2041 & 0.1886 & -1.9172 & 0.3840 & 0.1219 \\
Text2Video-Zero~\cite{text2videozero}   & 0.2539 & 0.1933 & -1.2050 & 0.4753 & 0.1952 \\

CogVideoX-2B~\cite{cogvideox}         & 0.2038 & 0.1901 & -1.2301 & 0.4585 & 0.1997 \\
ZeroScope~\cite{zeroscope}            & 0.2289 & 0.1933 & -1.1599 & 0.4892 & 0.2388 \\\hline
KLING 1.6~\cite{klingai2025}\textsuperscript{†}       & 0.2215 & \cellcolor[RGB]{255, 255, 200}0.1985 & \cellcolor[RGB]{255, 255, 200}-0.3419 & 0.5307 & \cellcolor[RGB]{255, 255, 200}0.2802  \\
Wanx 2.1~\cite{wan2025}\textsuperscript{†}      & 0.2308 & 0.1969 & \cellcolor[RGB]{255, 200, 200}-0.1418 & \cellcolor[RGB]{255, 200, 200}0.5970 & \cellcolor[RGB]{255, 200, 200}0.3328  \\
\bottomrule
\end{tabular}%
}
\label{overall-video-perf}
\end{table}


\begin{table}[htbp]
\centering
\caption{Overall performance of open-source \video models over 10K \name captions with VBench metrics. \colorbox[RGB]{255, 200, 200}{Red Cell} is the highest score. \colorbox[RGB]{200, 220, 255}{Blue Cell} is the lowest score.}
\resizebox{1\columnwidth}{!}{%
\begin{tabular}{lcccccc}
\toprule
\textbf{Model} & \makecell{\textbf{subject} \\ \textbf{consistency}} & \makecell{\textbf{background} \\ \textbf{consistency}} & \makecell{\textbf{motion}\\\textbf{smoothness}} & \makecell{\textbf{dynamic}\\\textbf{degree}} &
\makecell{\textbf{aesthetic}\\\textbf{quality}} & \makecell{\textbf{imaging}\\\textbf{quality}} \\
\midrule
Open-Sora 1.2          & \cellcolor[RGB]{255, 200, 200}0.9964 & \cellcolor[RGB]{255, 200, 200}0.9907 & \cellcolor[RGB]{255, 200, 200}0.9973 & \cellcolor[RGB]{200, 220, 255}0.0044 & 0.5235 & 0.6648\\
Text2Video-Zero   & \cellcolor[RGB]{200, 220, 255}0.8471 & \cellcolor[RGB]{200, 220, 255}0.9030 & \cellcolor[RGB]{200, 220, 255}0.8301 & \cellcolor[RGB]{255, 200, 200}0.9999 & 0.4889 & 0.7018\\
VideoCraft2          & 0.9768 & 0.9688 & 0.9833 & 0.3556 & 0.5515 & 0.6974\\
AnimateDiff          & 0.9823 & 0.9733 & 0.9859 & 0.1406 & 0.5427 & 0.5830\\
FreeInit           & 0.9581 & 0.9571 & 0.9752 & 0.4440 & 0.5200 & \cellcolor[RGB]{200, 220, 255}0.5456\\
ModelScope           & 0.9795 & 0.9831 & 0.9803 & 0.1281 & \cellcolor[RGB]{200, 220, 255}0.3993 & 0.6494\\
AnimateLCM           & 0.9883 & 0.9802 & 0.9887 & 0.0612 & \cellcolor[RGB]{255, 200, 200} 0.6323 & \cellcolor[RGB]{255, 200, 200} 0.6977\\
CogVideoX-2B         & 0.9583 & 0.9602 & 0.9823 & 0.4980 & 0.4607 & 0.6098\\
ZeroScope            & 0.9814 & 0.9811 & 0.9919 & 0.1670 & 0.4582 & 0.6782\\
\bottomrule
\end{tabular}%
}
\label{overall-video-perf-vbench}
\end{table}



\begin{enumerate}

    \item Open-source text-to-video models face challenges in balancing dynamics and consistency (Table~\ref{overall-video-perf-vbench}). This is especially evident in \OpenSora, which achieves high consistency but minimal dynamics, and \texttovideozero, which excels in dynamics but suffers from frame inconsistency.
    \item All models exhibit negative \imagereward (Table~\ref{overall-video-perf}), suggesting a lack of human-preferred visual appeal in the generated content, even in cases where certain models demonstrate strong semantic alignment.
    \item As expected, SOTA close-source text-to-video models outperform others overall, particularly in image reward, VQA score, and TIFA score. This indicates their superior alignment with human preferences, as well as stronger faithfulness and compositional capabilities in generation.
    \item Among open-source models, \VideoCrafter strikes a balance across key metrics, leading in human-preference alignment, faithfulness, consistency, and dynamic.
\end{enumerate}

\noindent\textbf{\threed results. (Table~\ref{overall-threed-perf})}


\begin{table}[htbp]
\centering
\caption{Overall performance of \threed models over 1K \name captions. \textsuperscript{†}Evaluated on a 100 caption subset due to high inference cost.}
\resizebox{\columnwidth}{!}{%
\begin{tabular}{lccccc}
\toprule
\textbf{Model} & \textbf{clip score} & \textbf{pick score} & \textbf{vqa score} & \textbf{tifa score} & \makecell{\textbf{image reward}\\\textbf{score}}  \\
\midrule
Latent-NeRF~\cite{latentnerf}           & 0.2115 & 0.1910 & 0.4767 & 0.2216 & -1.5311 \\
DreamFusion-sd~\cite{dreamfusion}      & 0.1961 & 0.1906 & 0.4421 & 0.1657 & -1.5582 \\
Magic3D-sd~\cite{magic3d}          & 0.1947 & 0.1903 & 0.4193 & 0.1537 & -1.6327 \\
SJC~\cite{sjc}                  & 0.2191 & 0.1915 & 0.5015 & 0.2563 & -1.4370 \\
DreamFusion-IF~\cite{dreamfusion}      & 0.1828 & 0.1857 & 0.3872 & 0.1416 & -1.9353 \\
Magic3D-IF~\cite{magic3d}         & 0.1919 & 0.1866 & 0.4039 & 0.1537 & -1.8465 \\
ProlificDreamer~\cite{prolificdreamer}      & 0.2125 & \textbf{0.1940} & \textbf{0.5411} & 0.2704 & -1.2774 \\
Meshy-4~\cite{meshy2025}\textsuperscript{†}   & \textbf{0.2163 }  & 0.1922 & 0.5290 & \textbf{0.2908} & \textbf{-1.0496} \\
\bottomrule
\end{tabular}%
}
\label{overall-threed-perf}
\end{table}

\begin{enumerate}
    \item Among open-source models, \ProlificDreamer outperforms other models, particularly in \imagereward, \vqascore and \tifa. 
    \item  All models receive negative \imagereward scores, highlighting a significant gap between human preference and current \threed generation capabilities.
    \item Meshy-4 demonstrates overall superior performance compared to all open-source models, especially in terms of \clipscore, \tifa and \imagereward, reflecting its strengths in semantic generation and human preference alignment.
\end{enumerate}

\subsection{Validation of Phrasing Robustness and Human Alignment}
\label{appendix:Phrasing-Robustness}

To assess robustness to linguistic variation and to verify that automated metrics reflect human preferences, we conduct two focused studies.

\subsubsection{Phrasing Robustness via Paraphrasing}
\label{app:template-generalization}
\textbf{Setup.} We sample 100 scene graphs from the 10K benchmark while preserving the distribution of object counts, relation density, and attribute complexity. For each graph, GPT-4o generates a linguistically varied yet graph-faithful caption using the prompt below.
\begin{tcolorbox}[colback=gray!3,colframe=gray!50,boxrule=0.4pt,title={Paraphrasing Prompt}]
\footnotesize\ttfamily
You are given a scene graph in JSON format, where:\\
-- "nodes" contain objects and their attributes,\\
-- "edges" describe relationships between objects or link attributes to objects.\\[2pt]

Your task:\\
1. Understand the semantic meaning of each node and edge.\\
2. Convert the graph into a natural language caption that describes the entire scene.\\
3. Include all objects, attributes, and relations from the graph, and strictly follow the graph structure.\\
4. Do not introduce new objects or relationships not present in the graph.\\[2pt]
Input: \{scene\_graph\}
\end{tcolorbox}
We then re-score all models with \vqascore under these paraphrased captions. Results are listed in Table~\ref{tab:Paraphrase}.
\begin{table}[h]
\centering
\caption{Paraphrase robustness: VQA Score and ranks on 100 graphs.}
\begin{tabular}{lccccc}
\toprule
Model & Orig.\ Score & Para.\ Score & Diff & Orig.\ Rank & Para.\ Rank \\
\midrule
DALLE-3 & 0.6871 & 0.7542 & +0.0671 & 1 & 1 \\
FLUX.1-schnell & 0.6132 & 0.6648 & +0.0516 & 2 & 2 \\
PixArt-$\Sigma$ & 0.6109 & 0.6159 & +0.0050 & 3 & 3 \\
PixArt-$\alpha$ & 0.6049 & 0.6043 & -0.0006 & 4 & 4 \\
Playground v2.5 & 0.5734 & 0.5075 & -0.0659 & 5 & 8 \\
Stable Diffusion 3 & 0.5579 & 0.5140 & -0.0439 & 6 & 7 \\
FLUX.1-dev & 0.5561 & 0.5024 & -0.0537 & 7 & 9 \\
DeepFloyd IF XL & 0.5397 & 0.5606 & +0.0209 & 8 & 5 \\
Wuerstchen v2 & 0.5352 & 0.5014 & -0.0338 & 9 & 10 \\
SDXL & 0.5328 & 0.5322 & -0.0006 & 10 & 6 \\
SD v2-1 & 0.5282 & 0.4961 & -0.0321 & 11 & 11 \\
\bottomrule
\end{tabular}
\label{tab:Paraphrase}
\end{table}

\paragraph{Findings.}
The Pearson correlation coefficient between model rankings on programmatic versus paraphrased captions is \textbf{0.9232}, indicating a very strong positive correlation.

This validation study demonstrates strong consistency between the two approaches. Importantly, the top-performing models (\dalle, \fluxschnell, \pixartsigma, \pixartalpha) maintain their rankings across both evaluation conditions, while the relative ordering of models remains largely consistent. This high correlation validates that our programmatic approach produces rankings that are generalizable and not artifacts of the templated caption generation. The slight variations observed (e.g., some mid-tier models showing small rank changes) are within expected bounds and do not affect the overall conclusions about model capabilities.

\subsubsection{Human Alignment Study}
\label{app:human-eval}

\noindent\textbf{Setup. } We evaluate six representative models (\dalle, \fluxschnell, \pixartsigma, \playground, \sdthree, \sdtwoone) with diverse performance characteristics and recruit 3 human evaluators. Three independent evaluators each assess 40 caption–image groups, with 10 shared overlapping groups across all evaluators to measure inter-annotator agreement. Evaluators ranked the generated images based on both relevance to the caption and overall visual quality. We show the rankings in Table~\ref{tab:human_rank}.

\noindent\textbf{Findings}

\begin{table}[!h]
\centering
\caption{Human vs.\ VQA rankings (lower is better).}
\begin{tabular}{lcc}
\toprule
Model & VQA Rank & Human Avg.\ Rank \\
\midrule
\dalle & 1 & 1 \\
\fluxschnell & 2 & 2 \\
\pixartsigma & 3 & 4 \\
\playground & 4 & 3 \\
\sdthree & 5 & 5 \\
\sdtwoone & 6 & 6 \\
\bottomrule
\end{tabular}
\label{tab:human_rank}
\end{table}

\noindent\textit{Inter-annotator reliability.}
The 3 evaluators showed strong agreement on the 10 shared samples, with a Spearman correlation coefficient of \textbf{0.962}, demonstrating consistent human judgment criteria.

\medskip

\noindent\textit{Human–metric alignment.} 
The correlation between human rankings and our \vqascore rankings is \textbf{0.918}, indicating strong alignment between automated and human evaluation:

This study validates that our VQA Score-based rankings closely align with human preferences. The consistency between automated metrics and human judgment strengthens confidence in our benchmark's ability to assess model performance in a manner that reflects human perception.

\subsection{More Analysis with \name}
With \name, we can generate infinitely diverse and highly controllable captions. Using \name, we conduct several analyses to provide insights into the performance of today’s \vision models.

\subsubsection{Performance analysis across caption properties}
\label{app:caption_prop}
In this section, we delve into how model performance varies with respect to distinct properties of \name captions. While \name is capable of generating an extensive diversity of captions, these outputs inherently differ in key characteristics that influence model evaluation. Specifically, we examine three properties of the caption: Commonsense, Perplexity, and Scene Graph Complexity (captured as the number of elements in the captions). These properties are critical in understanding how different models perform across a spectrum of linguistic and semantic challenges presented by captions with varying levels of coherence, plausibility, and compositional richness.

\begin{figure*}[!htb]
    \centering
    \begin{subfigure}[t]{0.3\textwidth}
        \includegraphics[width=\linewidth]{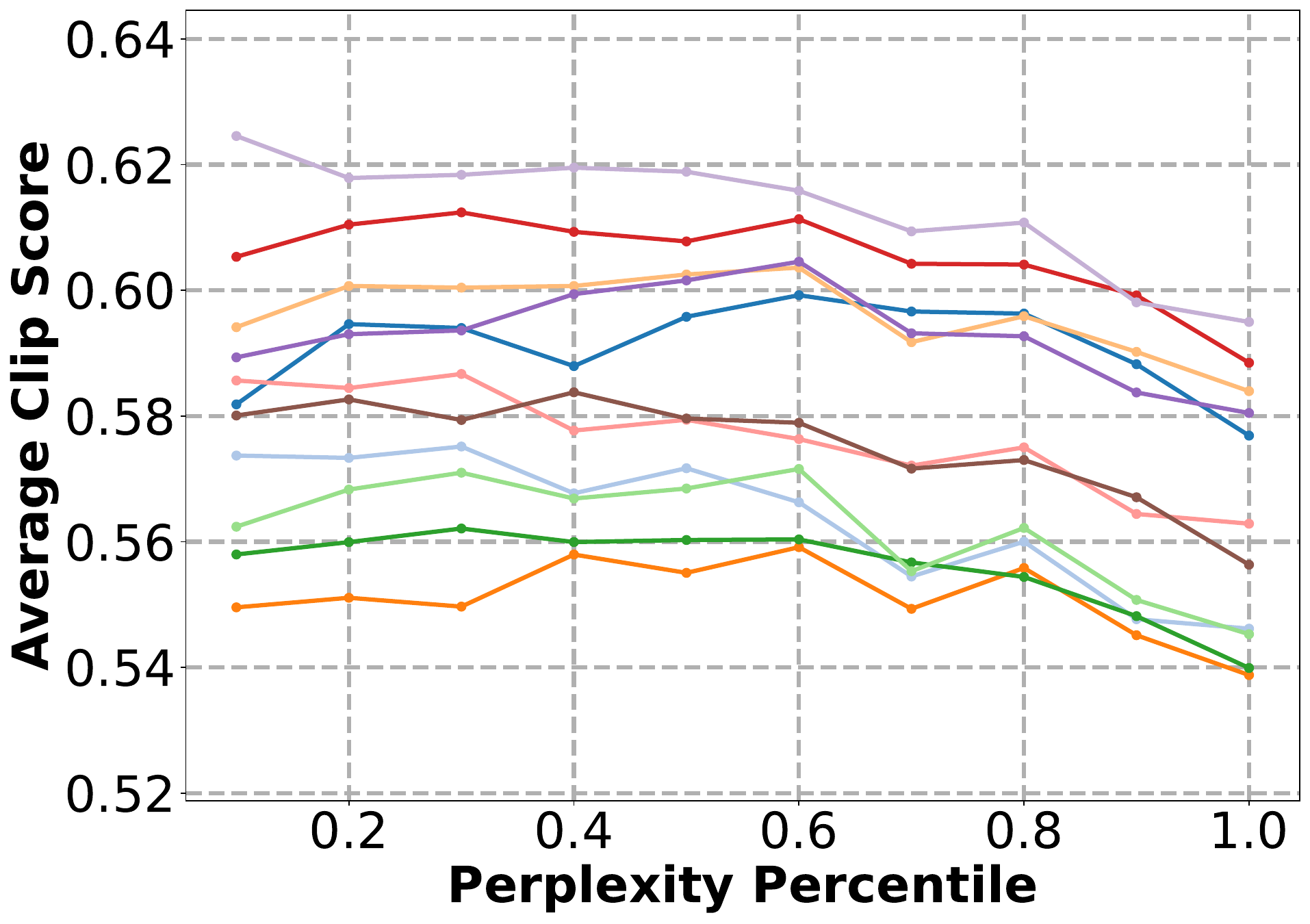}
        \caption{}
    \end{subfigure}
    \hfill
    \begin{subfigure}[t]{0.3\textwidth}
        \includegraphics[width=\linewidth]{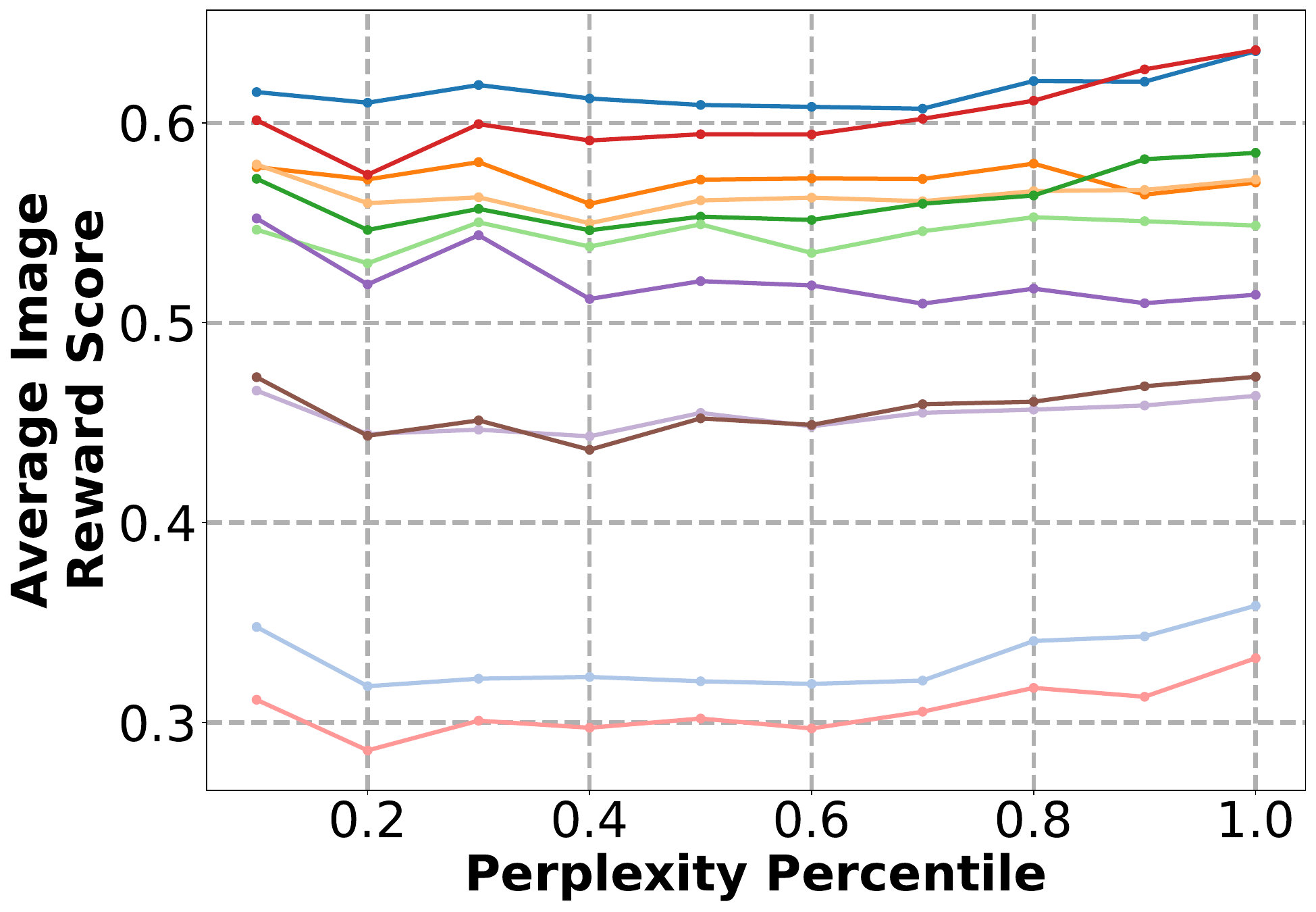}
        \caption{}
    \end{subfigure}
    \hfill
    \begin{subfigure}[t]{0.3\textwidth}
        \includegraphics[width=\linewidth]{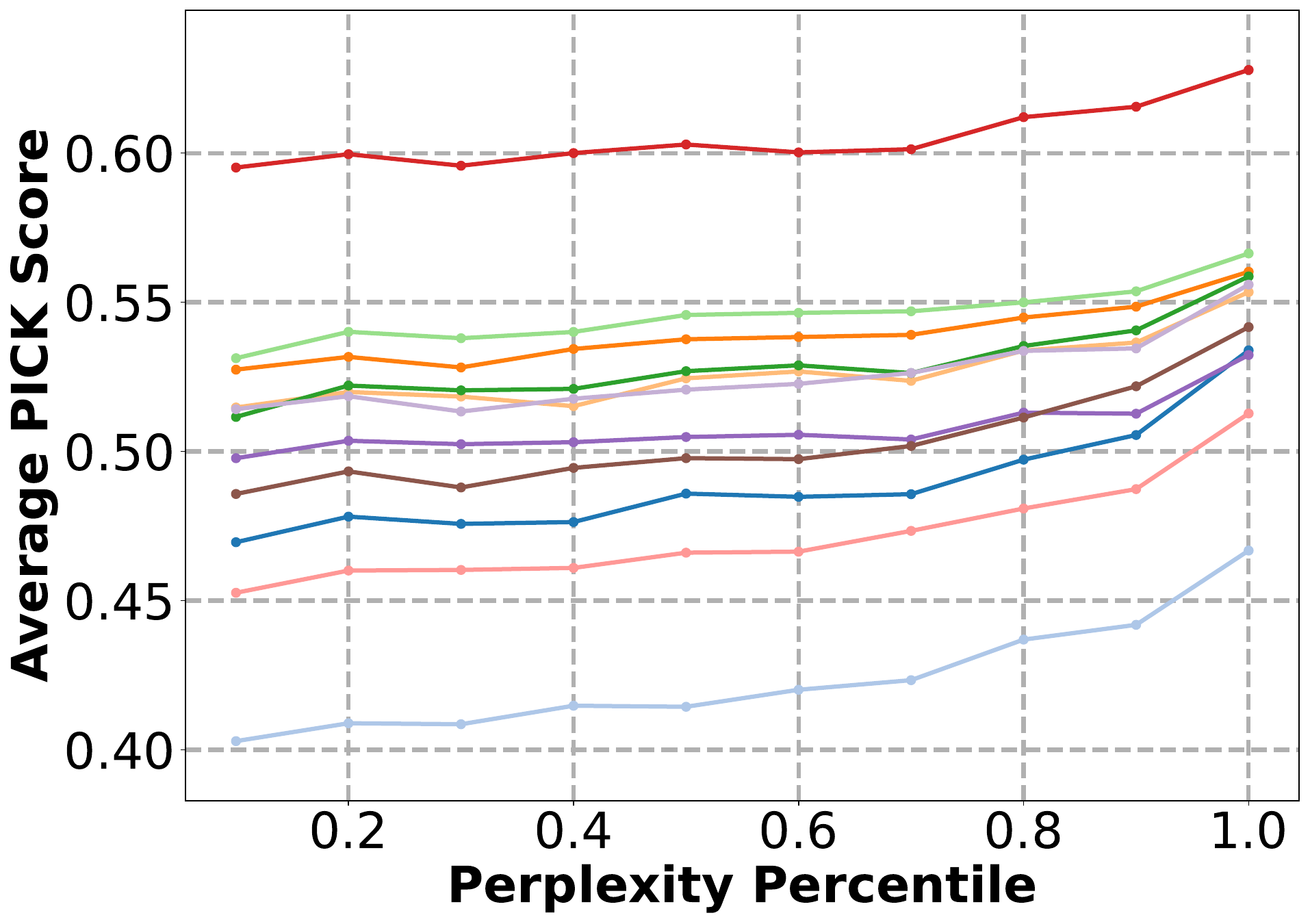}
        \caption{}
    \end{subfigure}
    \begin{subfigure}[t]{0.3\textwidth}
        \includegraphics[width=\linewidth]{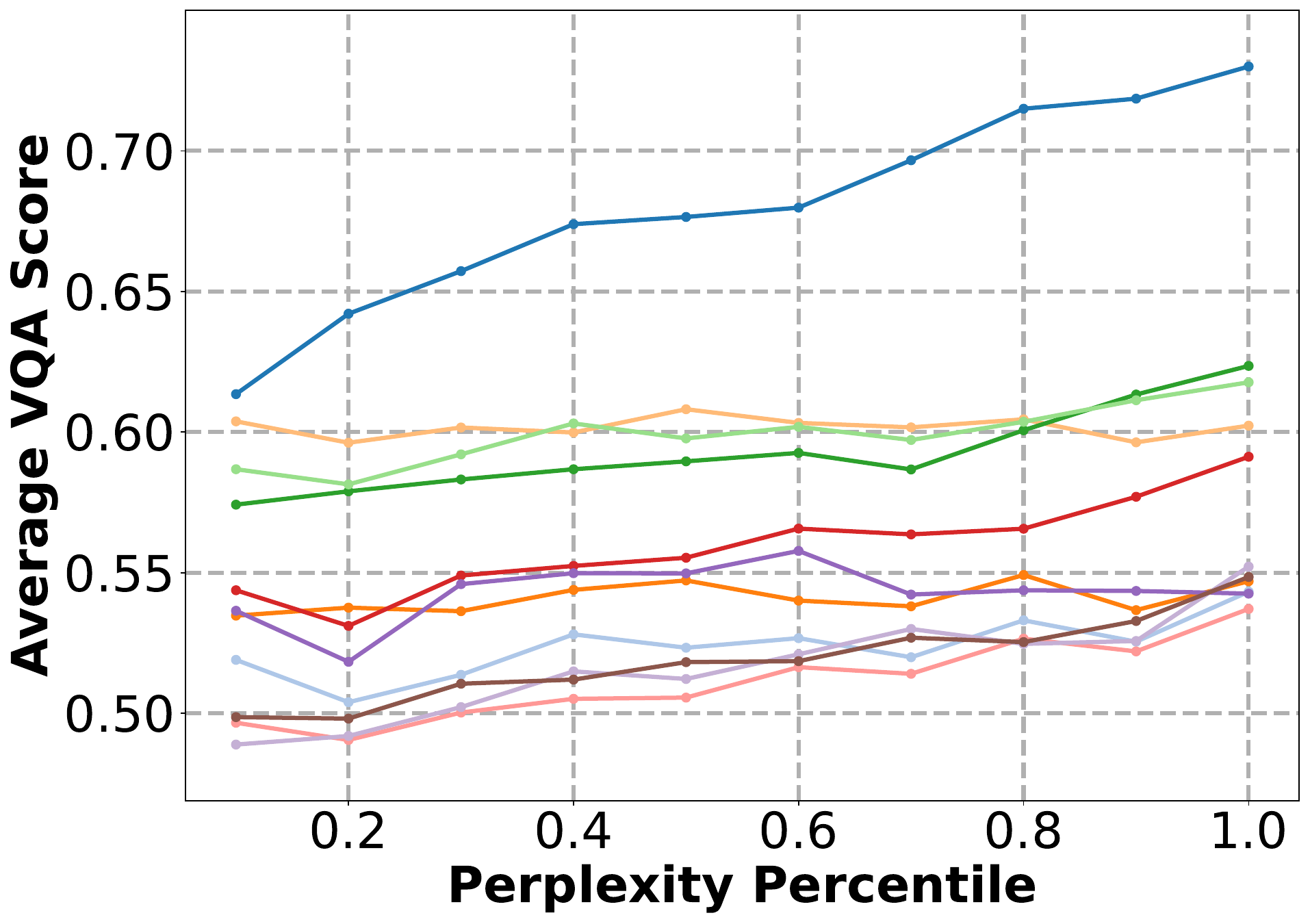}
        \caption{}
    \end{subfigure}
    \hfill
    \begin{subfigure}[t]{0.3\textwidth}
        \includegraphics[width=\linewidth]{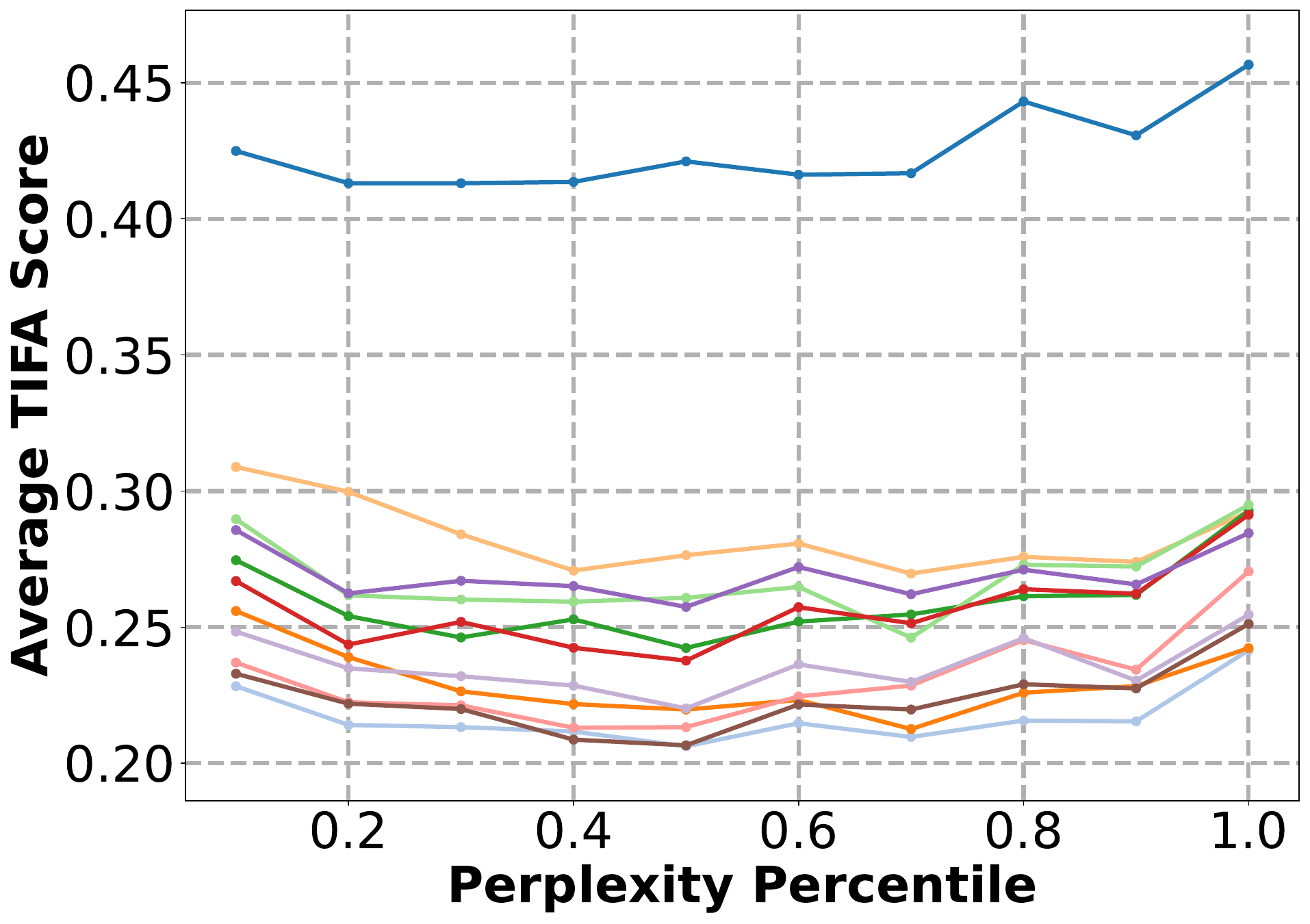}
        \caption{}
    \end{subfigure}
    \hfill
    \begin{subfigure}[t]{0.3\textwidth}
        \includegraphics[width=\linewidth]{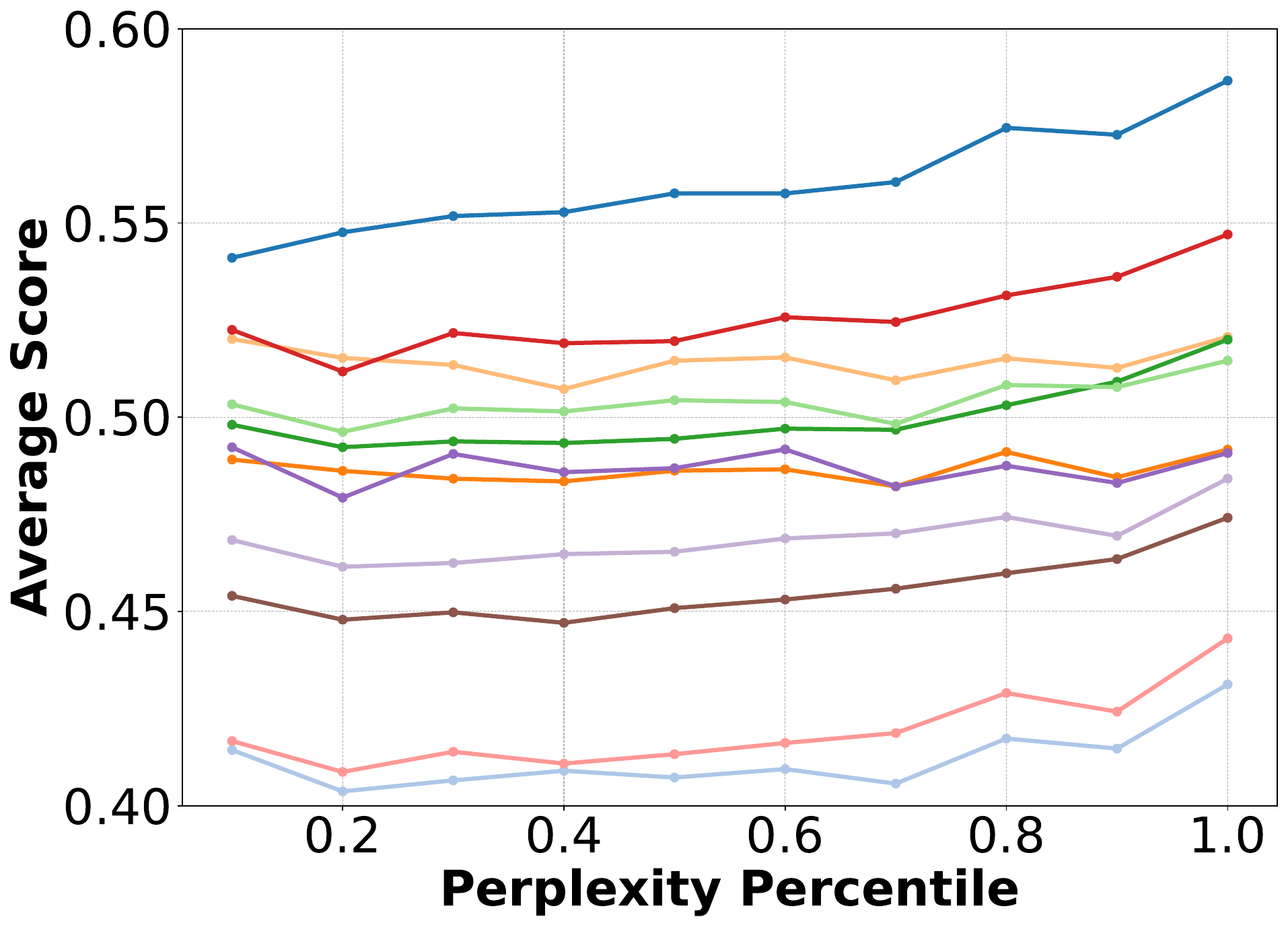}
        \caption{}
    \end{subfigure}

    \begin{subfigure}[b]{\textwidth}
        \includegraphics[width=\linewidth]{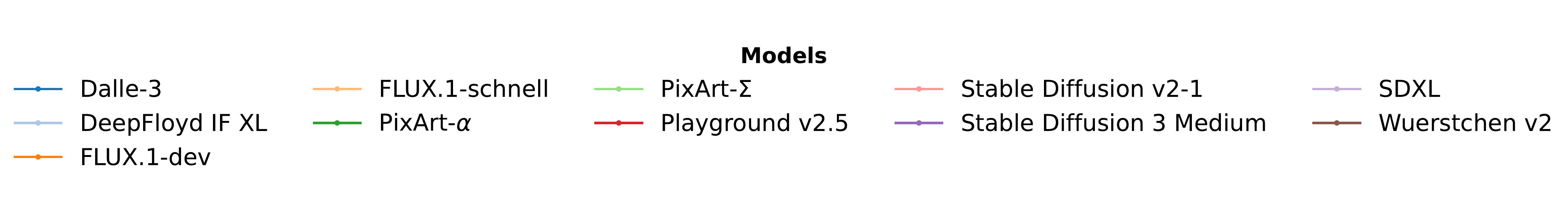}
    \end{subfigure}

    \caption{Average performance of models across different percentiles of perplexity of captions, evaluated on various metrics. From left to right, the perplexity decreases, indicating captions that are progressively more reasonable and easier for the LLM to generate.}
    \label{fig:app-perplexity}
\end{figure*}

\paragraph{Perplexity.  (Figure~\ref{fig:app-perplexity})}
Perplexity is a metric used to measure a language model's unpredictability or uncertainty in generating a text sequence. A higher perplexity value indicates that the sentences are less coherent or less likely to be generated by the model. 

As shown in Figure~\ref{fig:app-perplexity}, From left to right, when perplexity increases, indicating that the sentences become less reasonable and less typical of those generated by a language model, we observe no clear or consistent trends across all models and metrics. This suggests that the relationship between perplexity and model performance varies depending on the specific model and evaluation metric.

\begin{figure*}[!htb]
    \centering
    \begin{subfigure}[t]{0.3\textwidth}
        \includegraphics[width=\linewidth]{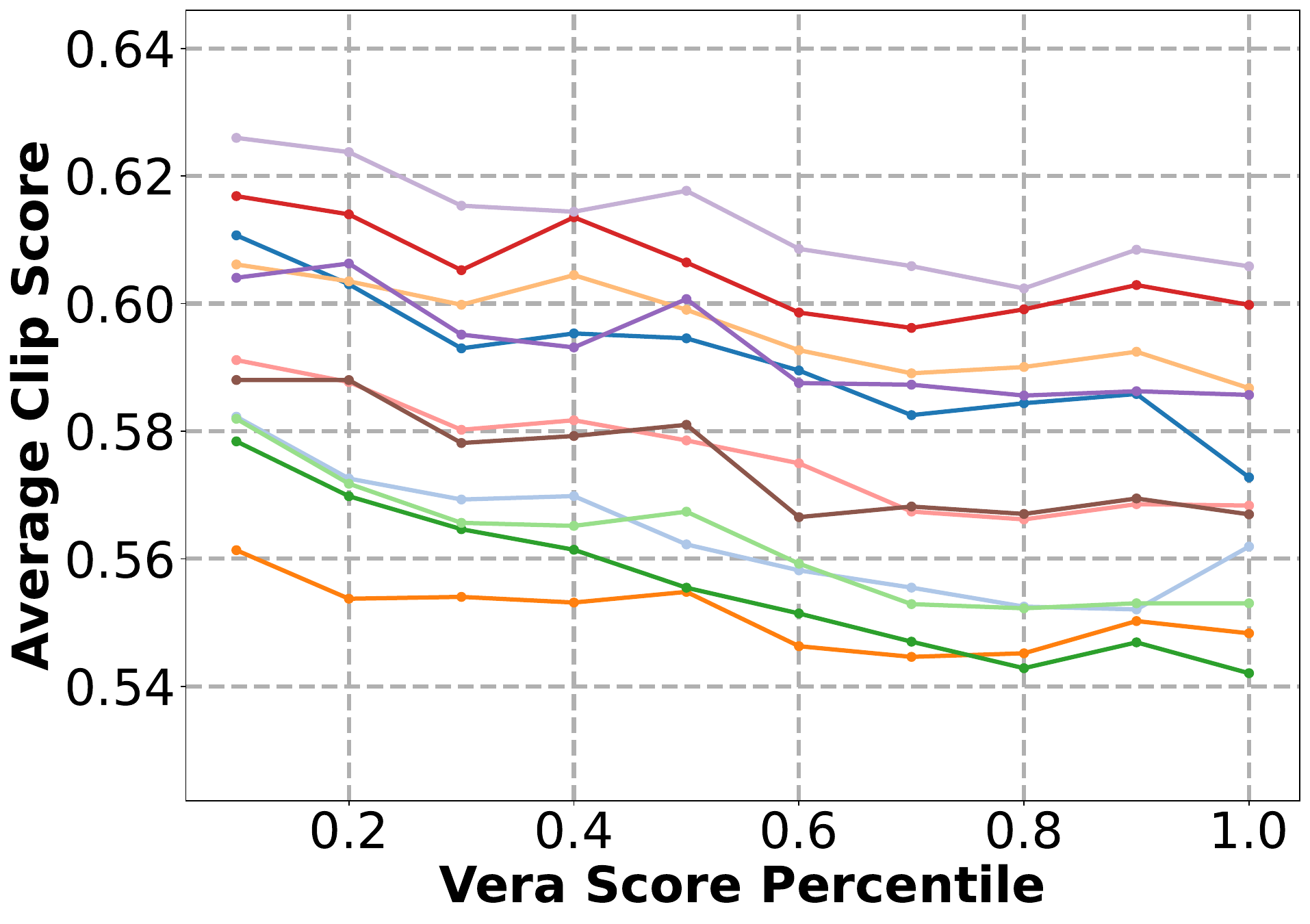}
        \caption{}
    \end{subfigure}
    \hfill
    \begin{subfigure}[t]{0.3\textwidth}
        \includegraphics[width=\linewidth]{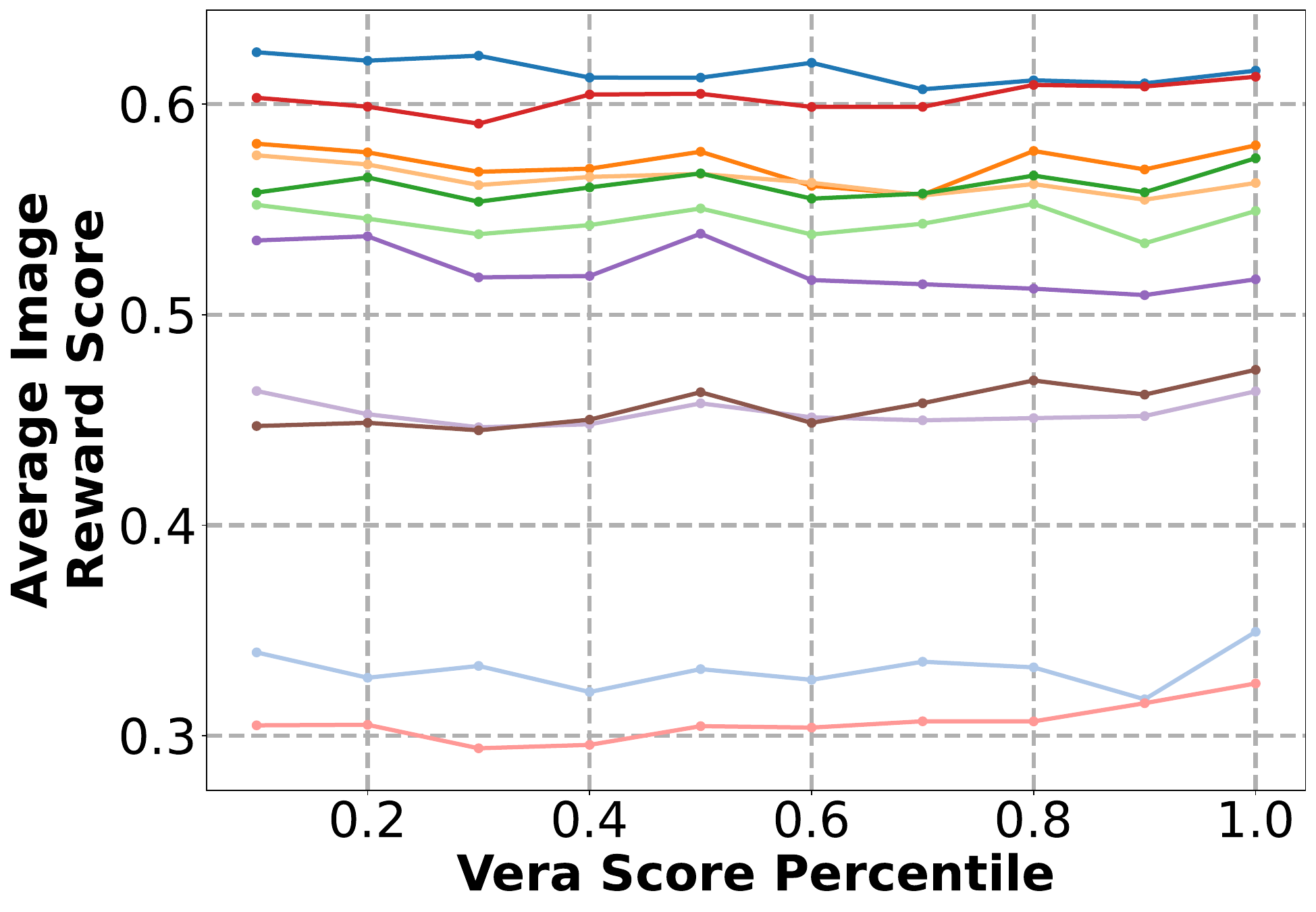}
        \caption{}
    \end{subfigure}
    \hfill
    \begin{subfigure}[t]{0.3\textwidth}
        \includegraphics[width=\linewidth]{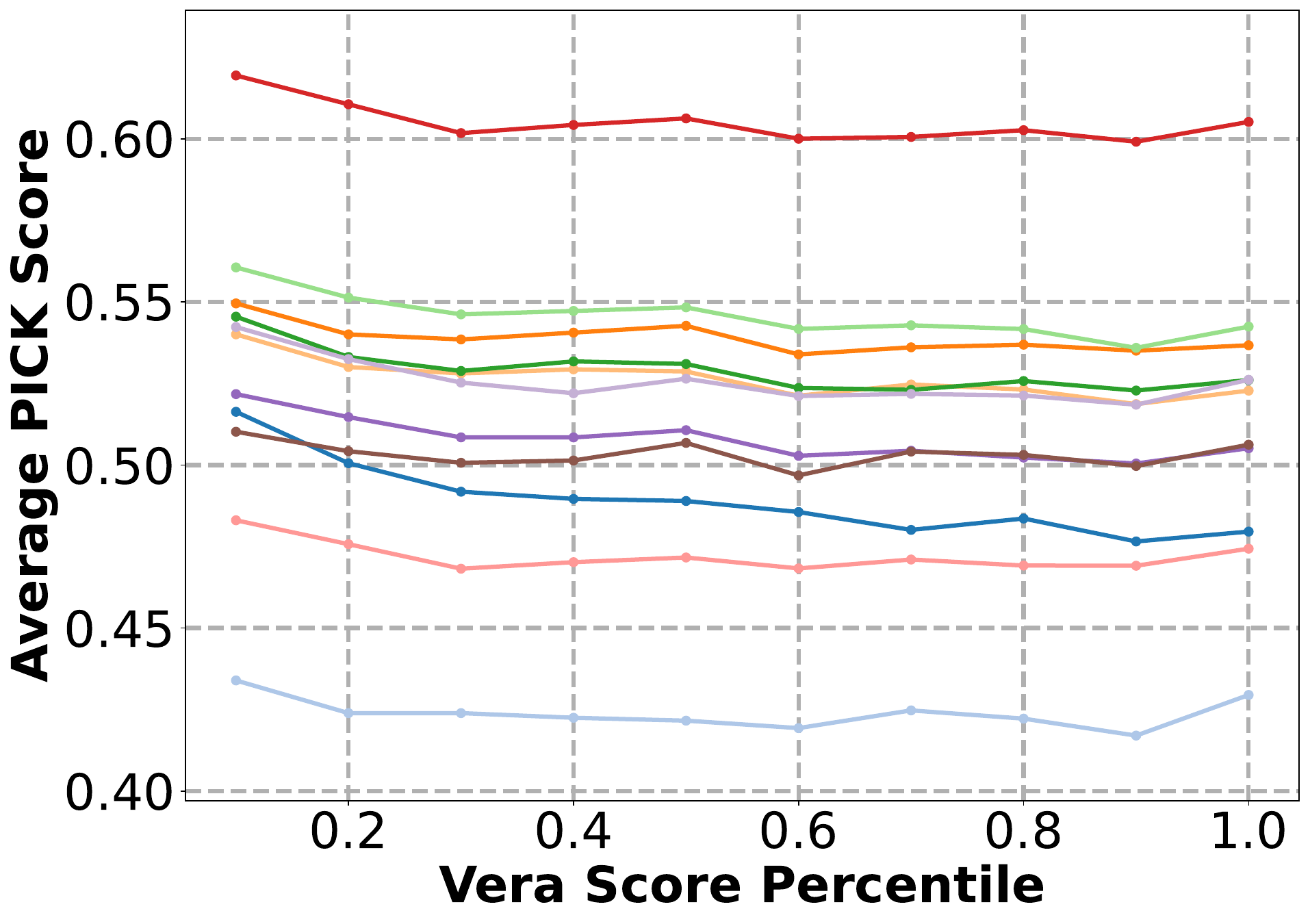}
        \caption{}
    \end{subfigure}

    \begin{subfigure}[t]{0.3\textwidth}
        \includegraphics[width=\linewidth]{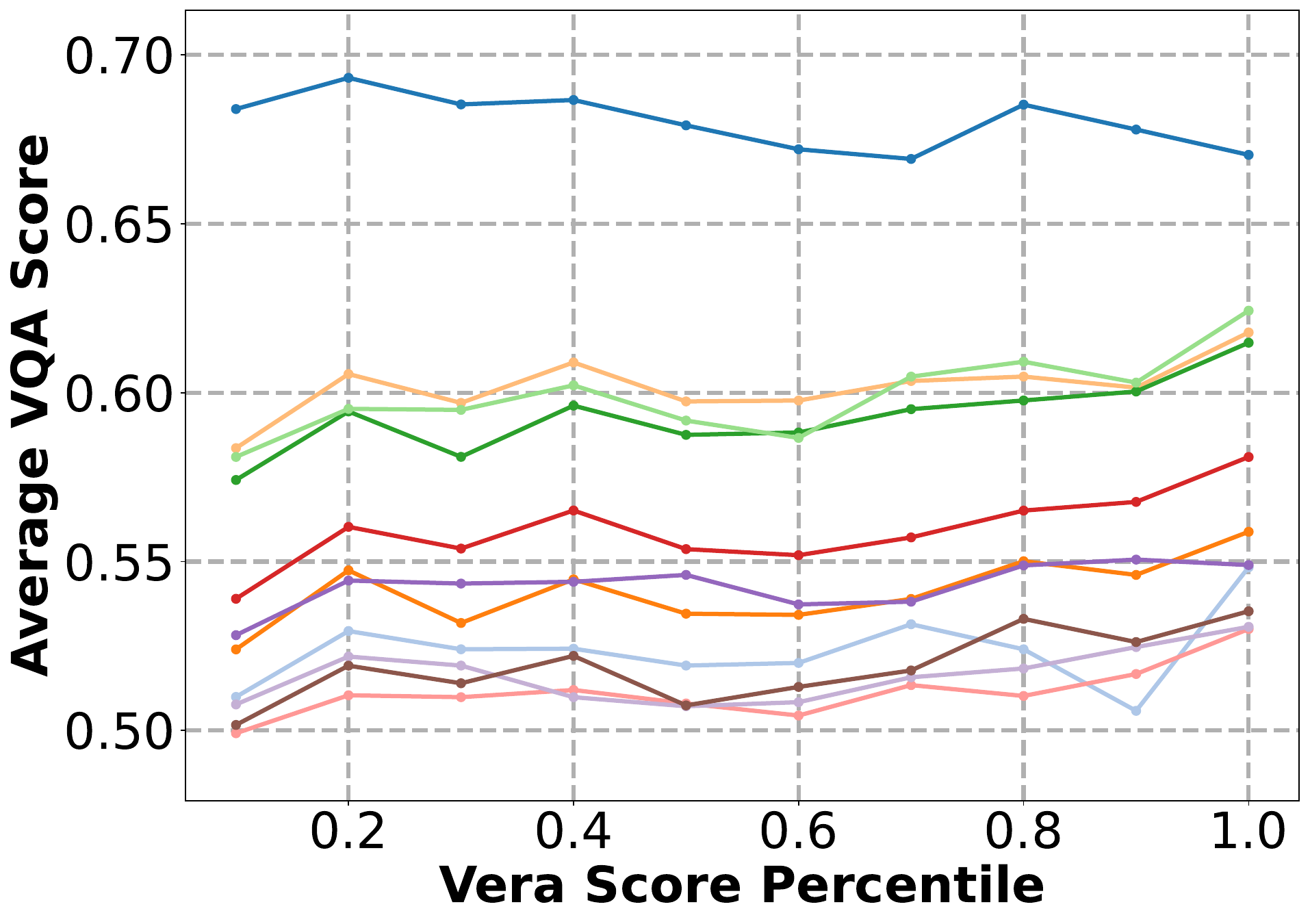}
        \caption{}
    \end{subfigure}
    \hfill
    \begin{subfigure}[t]{0.3\textwidth}
        \includegraphics[width=\linewidth]{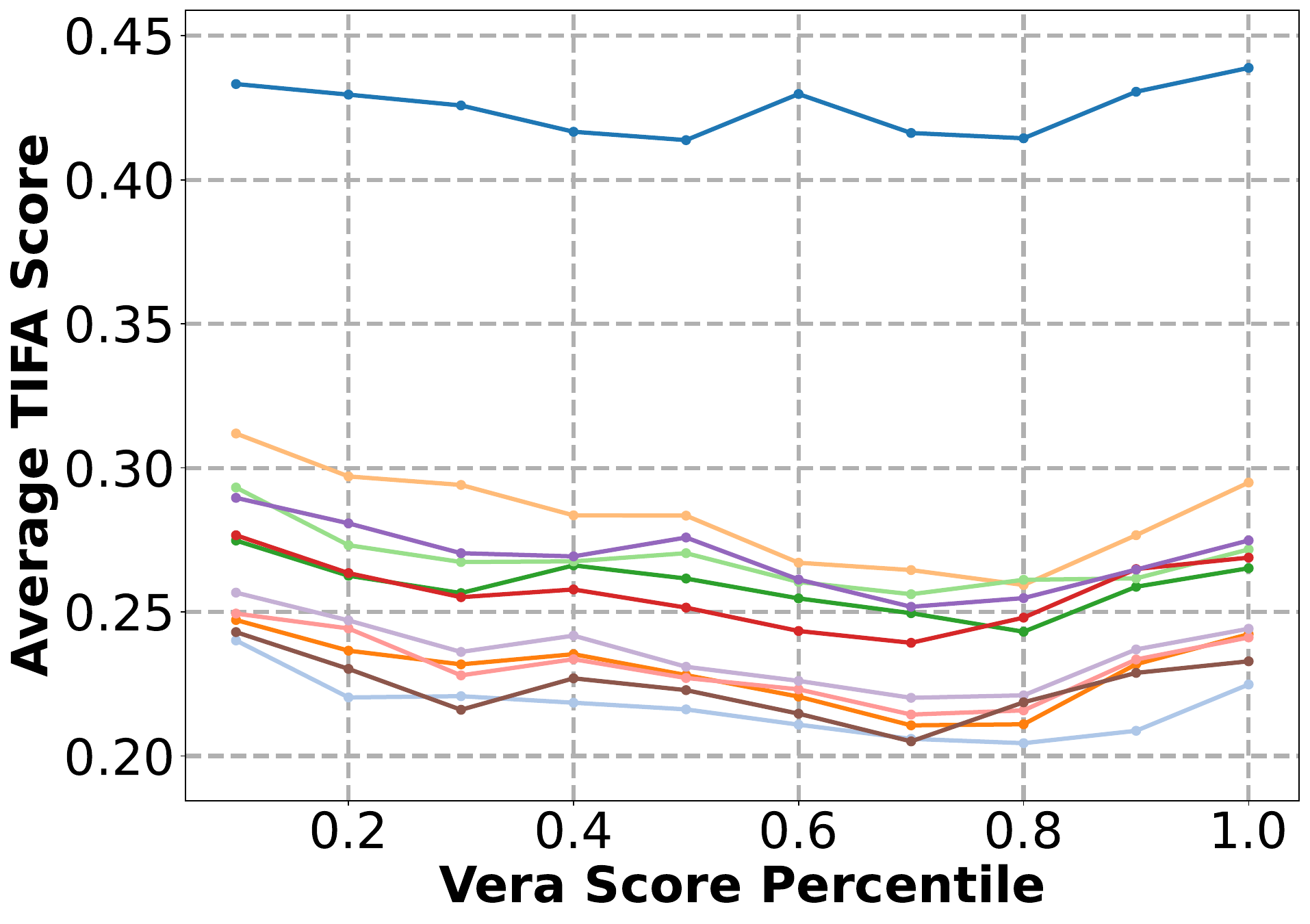}
        \caption{}
    \end{subfigure}
    \hfill
    \begin{subfigure}[t]{0.3\textwidth}
        \includegraphics[width=\linewidth]{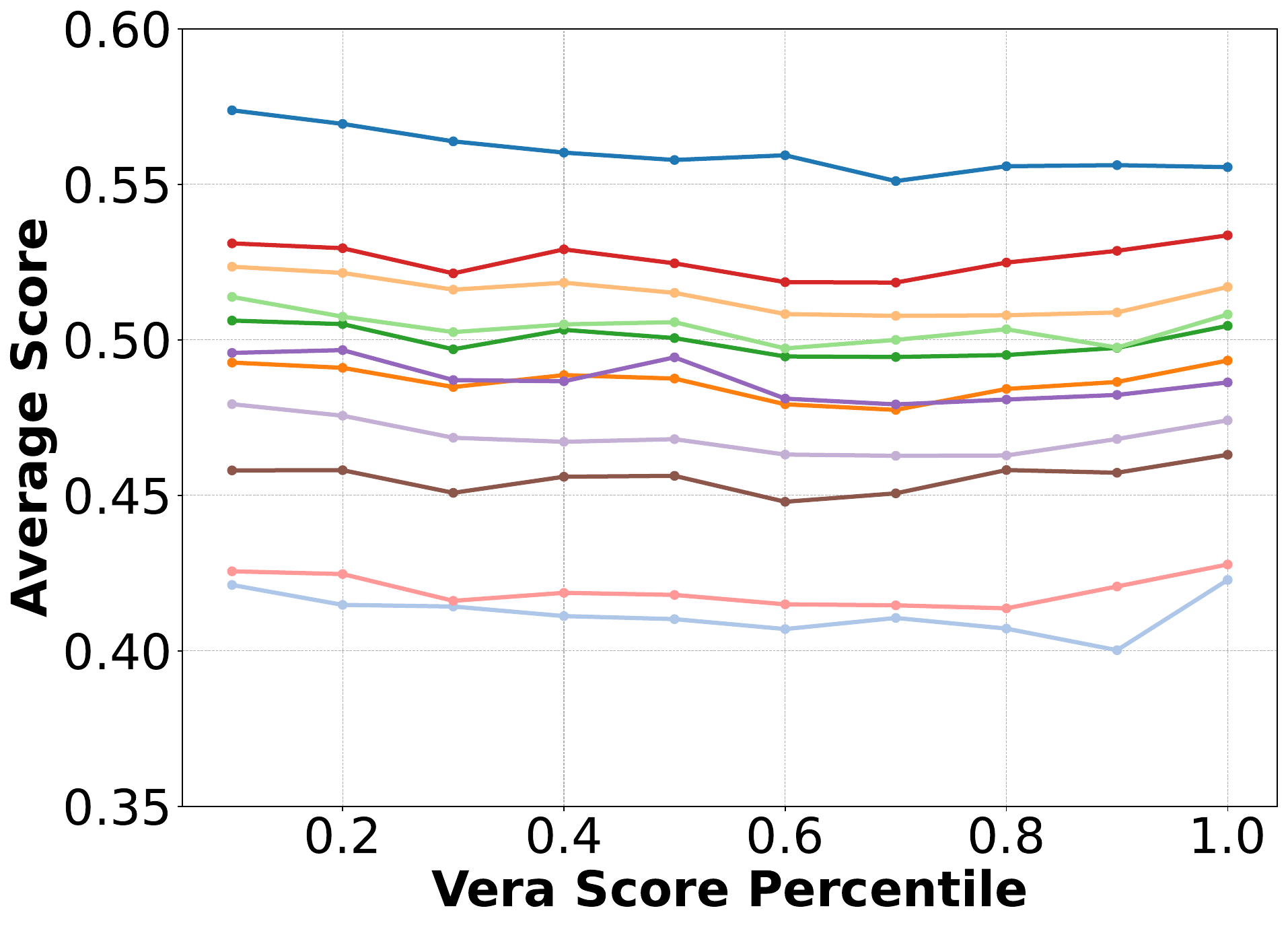}
        \caption{}
    \end{subfigure}

    \begin{subfigure}[b]{\textwidth}
        \includegraphics[width=\linewidth]{imgs/appendix/overall/perplexity/legend_only.pdf}
    \end{subfigure}

    \caption{Average performance of models across different percentiles of Vera Score for captions, evaluated on various metrics. From left to right, the Vera Score decreases, indicating captions that exhibit less commonsense reasoning and are more likely to describe implausible scenes.}
    \label{fig:app-commonsense}
\end{figure*}

\paragraph{Commonsense. (Figure~\ref{fig:app-commonsense})}

\begin{figure*}[!htb]
    \centering
    \begin{subfigure}[t]{0.3\textwidth}
        \includegraphics[width=\linewidth]{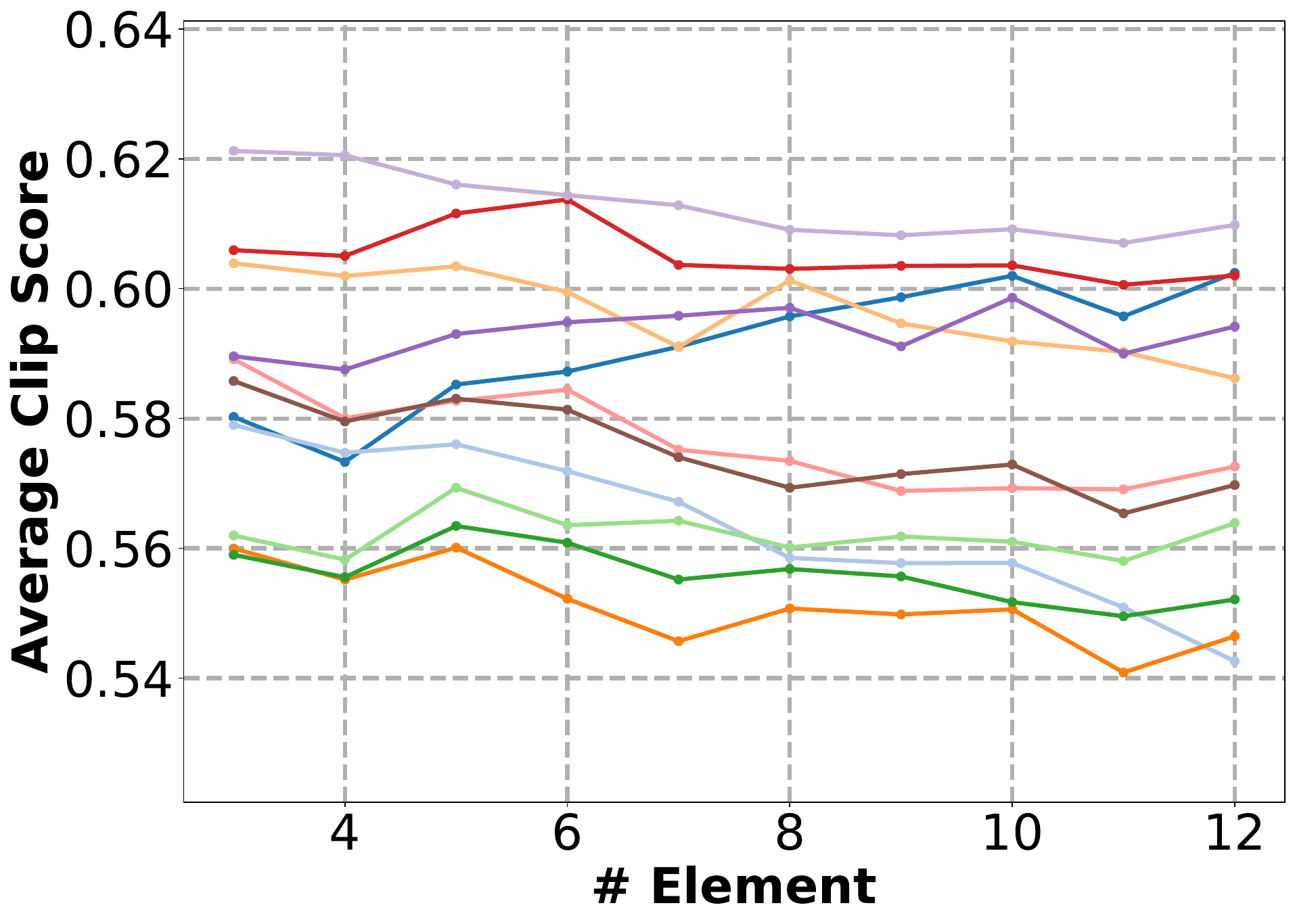}
        \caption{}
    \end{subfigure}
    \hfill
    \begin{subfigure}[t]{0.3\textwidth}
        \includegraphics[width=\linewidth]{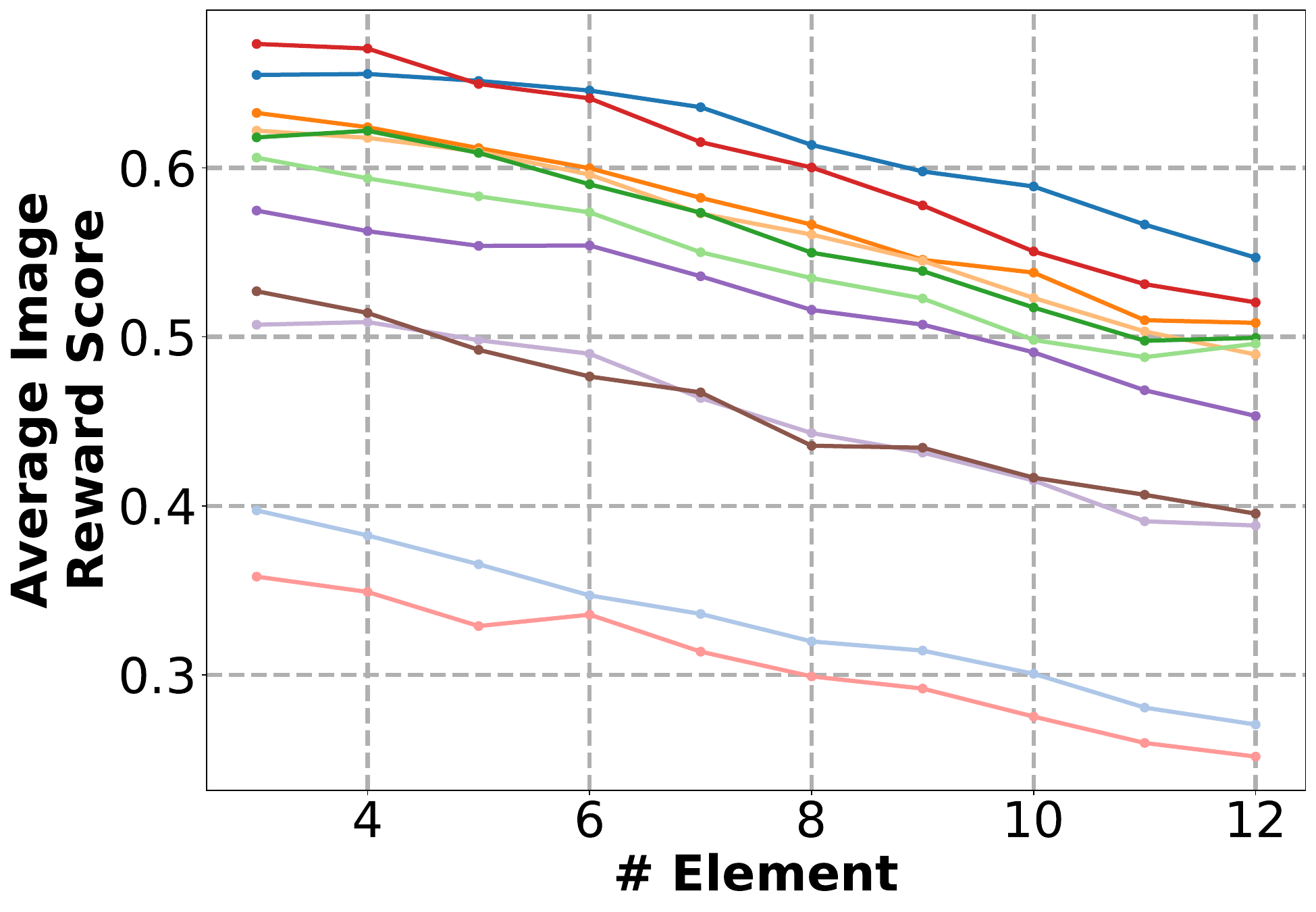}
        \caption{}
    \end{subfigure}
    \hfill
    \begin{subfigure}[t]{0.3\textwidth}
        \includegraphics[width=\linewidth]{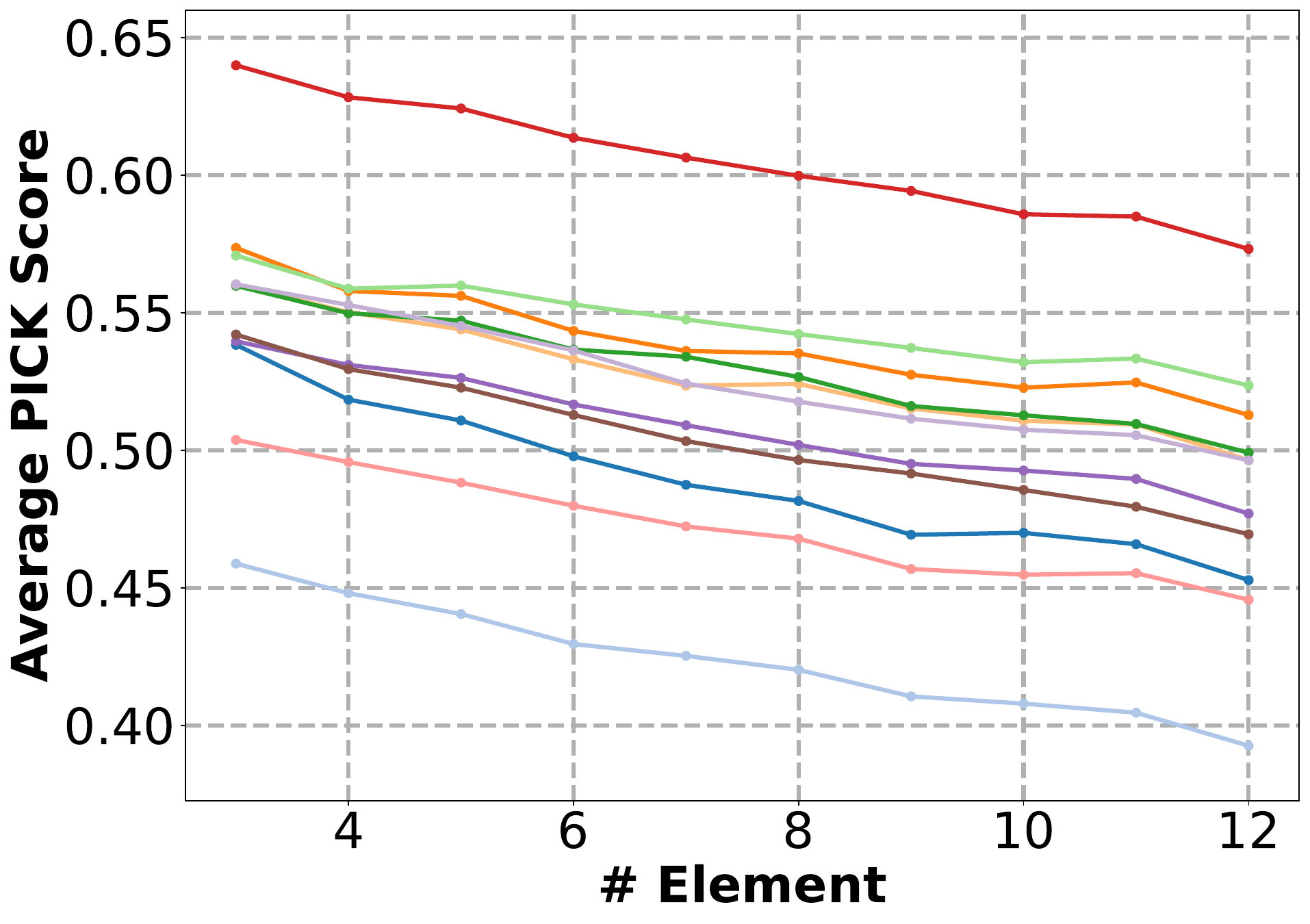}
        \caption{}
    \end{subfigure}

    \begin{subfigure}[t]{0.3\textwidth}
        \includegraphics[width=\linewidth]{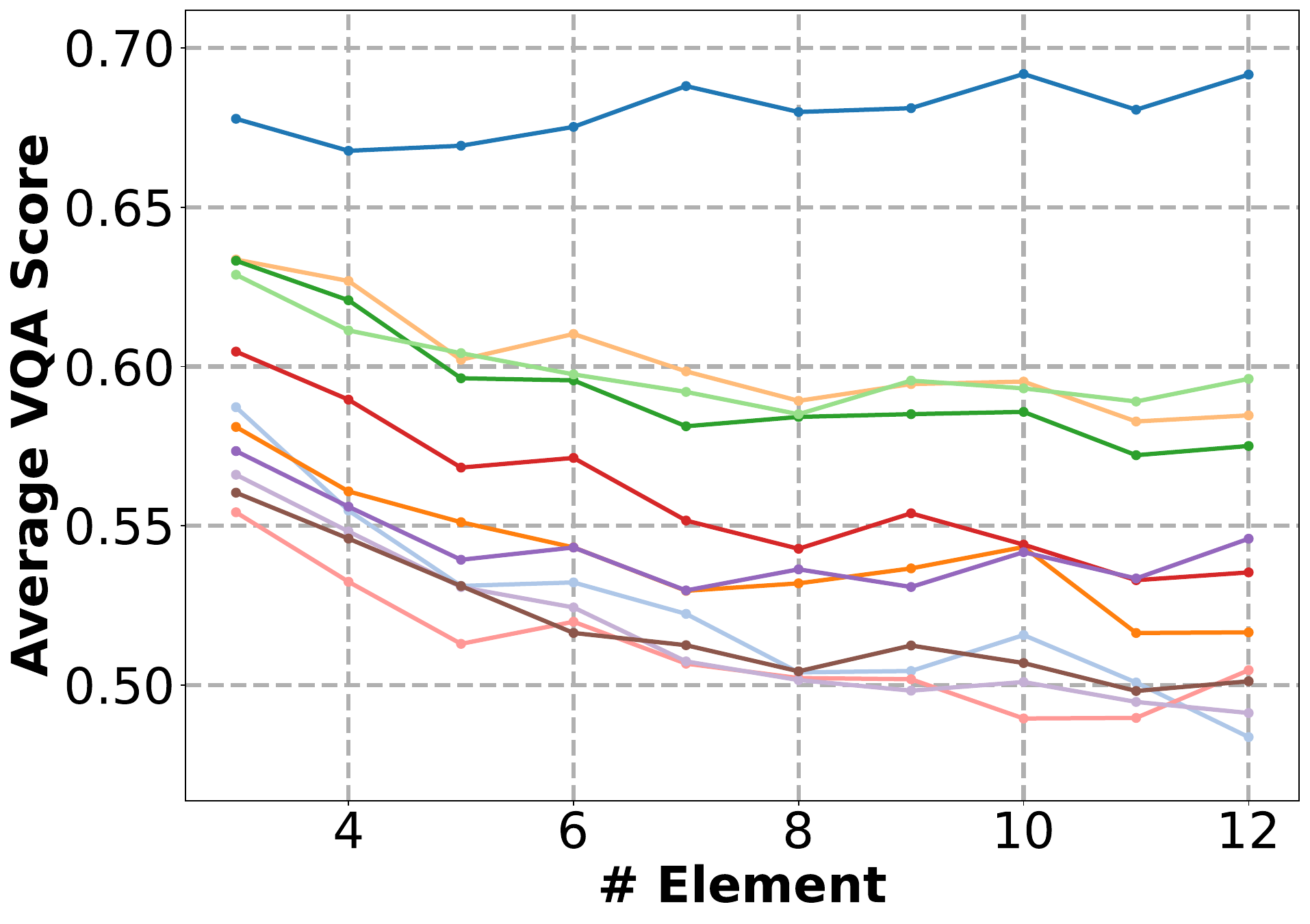}
        \caption{}
    \end{subfigure}
    \hfill
    \begin{subfigure}[t]{0.3\textwidth}
        \includegraphics[width=\linewidth]{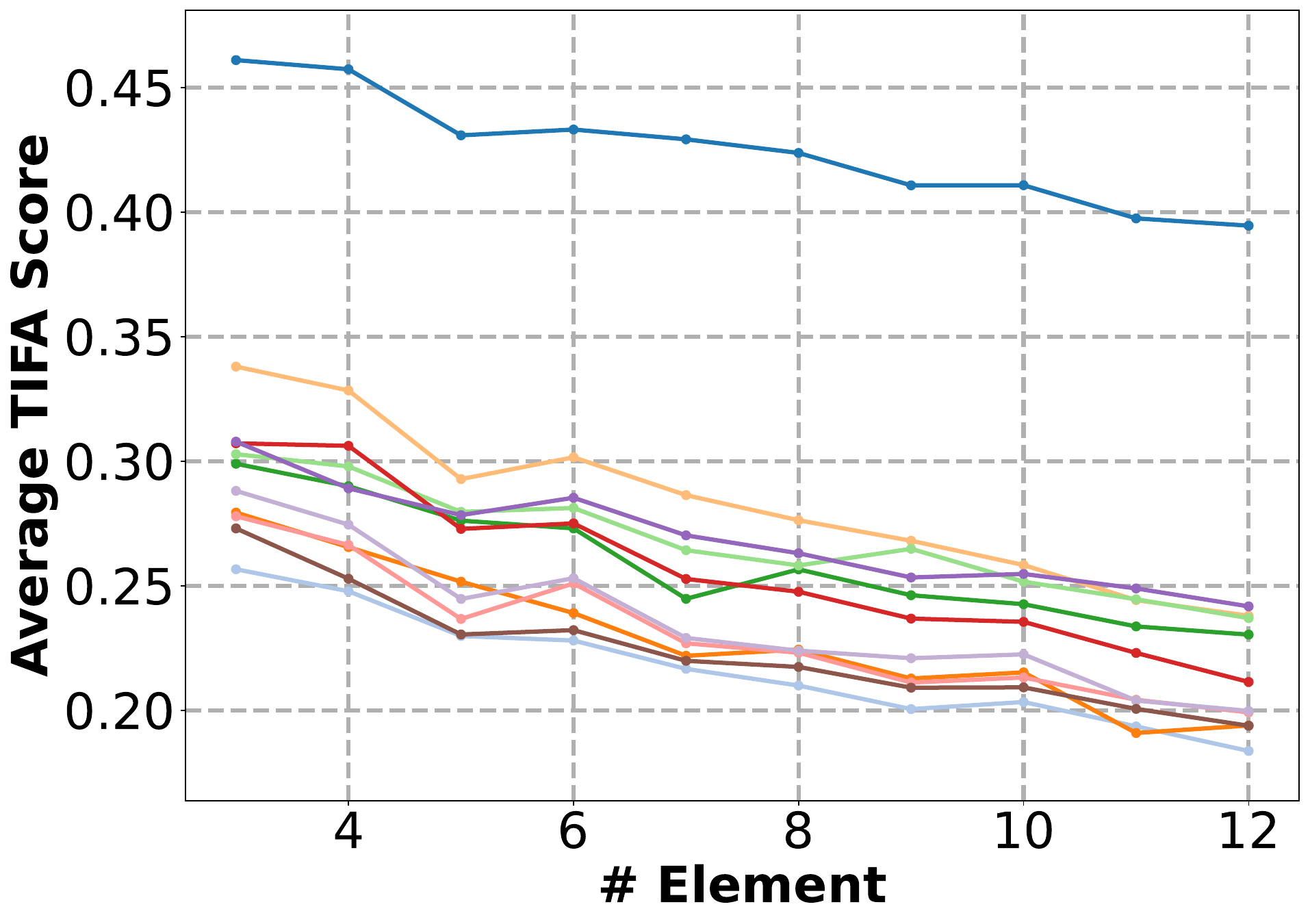}
        \caption{}
    \end{subfigure}
    \hfill
    \begin{subfigure}[t]{0.3\textwidth}
        \includegraphics[width=\linewidth]{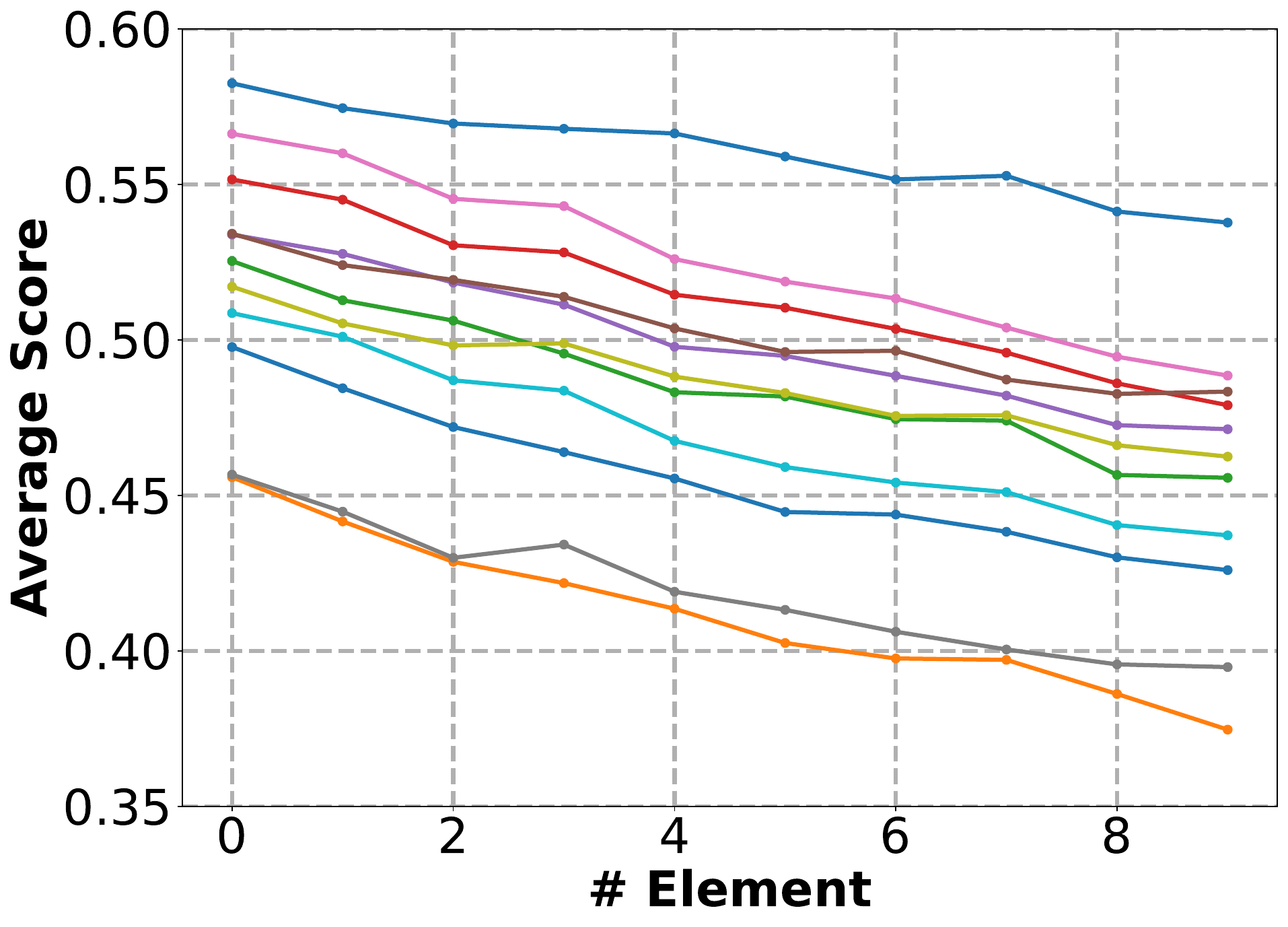}
        \caption{}
    \end{subfigure}

    \begin{subfigure}[b]{\textwidth}
        \includegraphics[width=\linewidth]{imgs/appendix/overall/perplexity/legend_only.pdf}
    \end{subfigure}
    \caption{Average performance of models across different numbers of elements (objects + attributes + relations) in the scene graph (complexity of the scene graph) of the captions, evaluated on various metrics. From left to right, as the number of elements (complexity) increases, the scene graphs become more complicated and compositional.}
    \label{fig:app-complexity}
\end{figure*}

Commonsense is an inherent property of text. We utilize the Vera Score~\cite{liu2023verageneralpurposeplausibilityestimation}, a metric generated by a fine-tuned LLM to evaluate the text's commonsense level.

As shown in Figure~\ref{fig:app-commonsense}, from left to right, as the Vera Score increases—indicating that the captions exhibit greater commonsense reasoning—we observe a general improvement in performance across all metrics and models, except for \clipscore. This trend underscores the correlation between commonsense-rich captions and enhanced model performance.

\paragraph{Element Numbers (Complexity of Scene Graph).  (Figure~\ref{fig:app-complexity})}





Finally, we evaluate model performance across total element numbers in the captions, which represent the complexity of scene graphs (objects + attributes + relations). 

From left to right, the complexity of scene graphs becomes higher, reflecting more compositional and intricate captions. Across most metrics and models, we observe a noticeable performance decline as the scene graphs become more complex. However, an interesting exception is observed in the performance of \dalle. Unlike other models, \dalle performs exceptionally well on \vqascore and \tifa, particularly on \vqascore, where it even shows a slight improvement as caption complexity increases. This suggests that \dalle may have a unique capacity to handle complex and compositional captions effectively.

\subsubsection{Analysis on different metrics}
Compared with most LLM and VLM benchmarks that use multiple-choice questions and accuracy as metrics. There is no universal metric in evaluating \vision models. Researchers commonly used model-based metrics like \clipscore, \vqascore, etc. Each of these metrics is created and fine-tuned for different purposes with bias. Therefore, we also analysis on different metrics.
\paragraph{\clipscore isn't a universal metric.}


\clipscore is one of the most widely used metrics in \vision for evaluating the alignment between visual content and text. However, our analysis reveals that \clipscore is not a perfect metric and displays some unusual trends. For instance, as shown in Figures~\ref{fig:app-perplexity}, \ref{fig:app-commonsense}, and \ref{fig:app-complexity}, we compute the perplexity across 10K captions used in our study, where higher perplexity indicates more unpredictable or disorganized text. Interestingly, unlike other metrics, \clipscore decreases as perplexity lowers, suggesting that \clipscore tends to favor more disorganized text. This behavior is counterintuitive and highlights the potential limitations of using \clipscore as a robust alignment metric.

\paragraph{Limitations of human preference-based metrics.} 
We use two metrics fine-tuned using human preference data: \pickscore and \imagereward. However, we found that these metrics exhibit a strong bias toward the data on which they were fine-tuned. For instance, as shown in Table~\ref{table:app-overall-image-perf}, \pickscore assigns similar scores across all models, failing to provide significant differentiation or meaningful insights into model performance. In contrast, \imagereward demonstrates clearer preferences, favoring models such as \dalle and \playground, which incorporated human-alignment techniques during their training. However, this metric shows a significant drawback: it assigns disproportionately large negative scores to models like \sdtwoone, indicating a potential over-sensitivity to alignment mismatches. Such behavior highlights the limitations of these metrics in providing fair and unbiased evaluations across diverse model architectures.

\paragraph{\vqascore and \tifa are relative reliable metrics.}
Among the evaluated metrics, \vqascore and \tifa stand out by assessing model performance on VQA tasks, rather than relying solely on subjective human preferences. This approach enhances the interpretability of the evaluation process. Additionally, we observed that the results from \vqascore and \tifa show a stronger correlation with other established benchmarks. Based on these advantages, we recommend prioritizing these two metrics for evaluation. However, it is important to note that their effectiveness is constrained by the limitations of the VQA models utilized in the evaluation.







\subsubsection{Fairness analysis}
We evaluate fairness by examining the model’s performance across different genders and races. Specifically, we calculate the average performance for each node and its associated child nodes within the taxonomy tree constructed for objects. For example, the node “females” includes child nodes such as “waitresses,” and their combined performance is considered in the analysis.

\paragraph{Gender.}

\begin{figure}[]
  \centering
  \begin{minipage}[t]{0.48\linewidth}
    \centering
    \includegraphics[width=\linewidth]{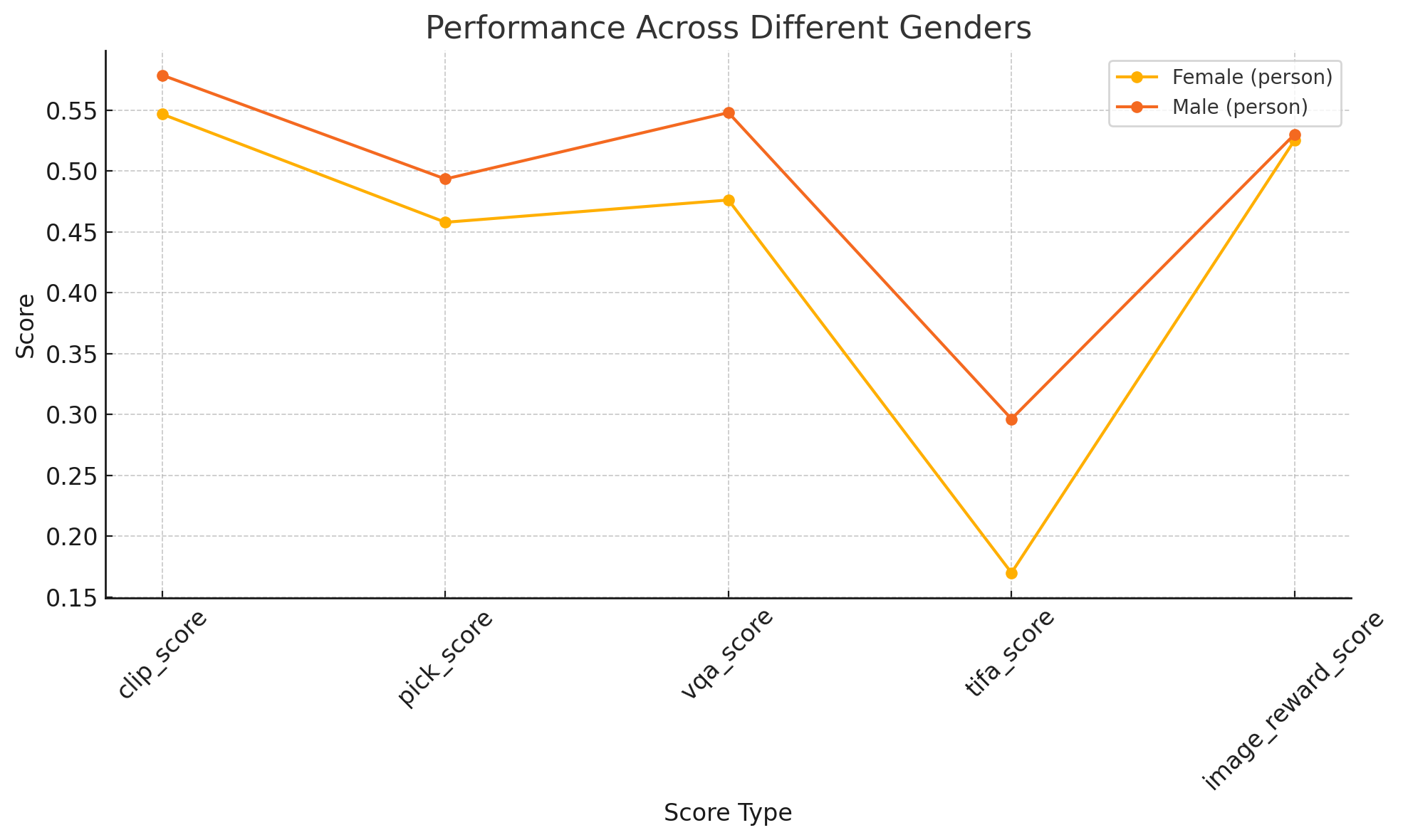}
    \caption{Average performance scores of all models across different genders evaluated using various metrics.}
    \label{fig:gender}
  \end{minipage}
  \hfill
  \begin{minipage}[t]{0.48\linewidth}
    \centering
    \includegraphics[width=\linewidth]{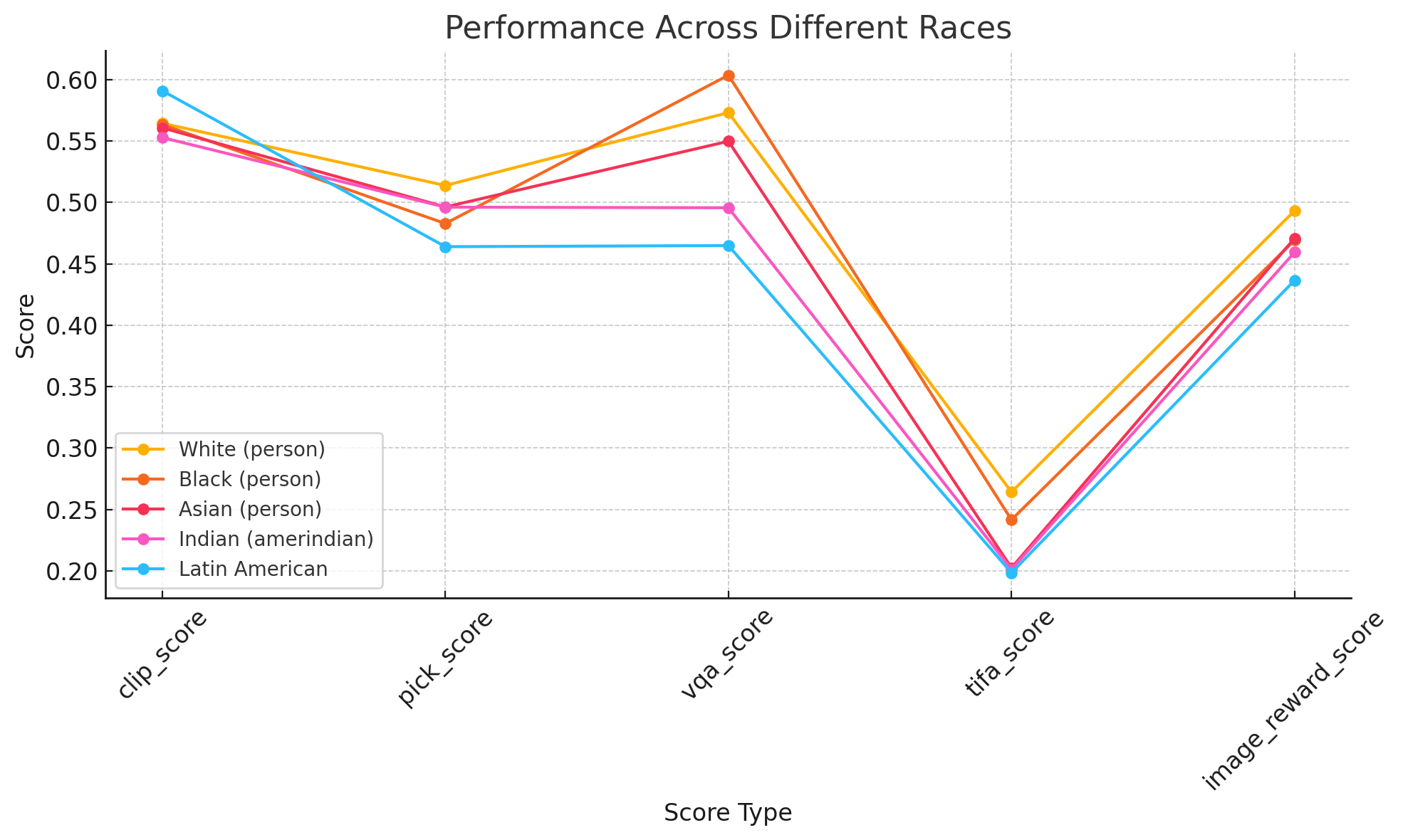}
    \caption{Average performance scores of all models across different races evaluated using various metrics.}
    \label{fig:race}
  \end{minipage}
\end{figure}
In gender, we observe a notable performance gap between females and males, as could be seen from Figure~\ref{fig:gender}, Models are better at generating male concepts.

\paragraph{Race.}
There are also performance gaps in different races. From Figure~\ref{fig:race}, we found that "white (person)" and "black (person)" perform better than "asian (person)", "Indian (amerindian)", and "Latin American".


\subsubsection{Correlation of \name with other \vision benchmarks}

\begin{figure}[]
  \centering
  \includegraphics[width=0.5\linewidth]{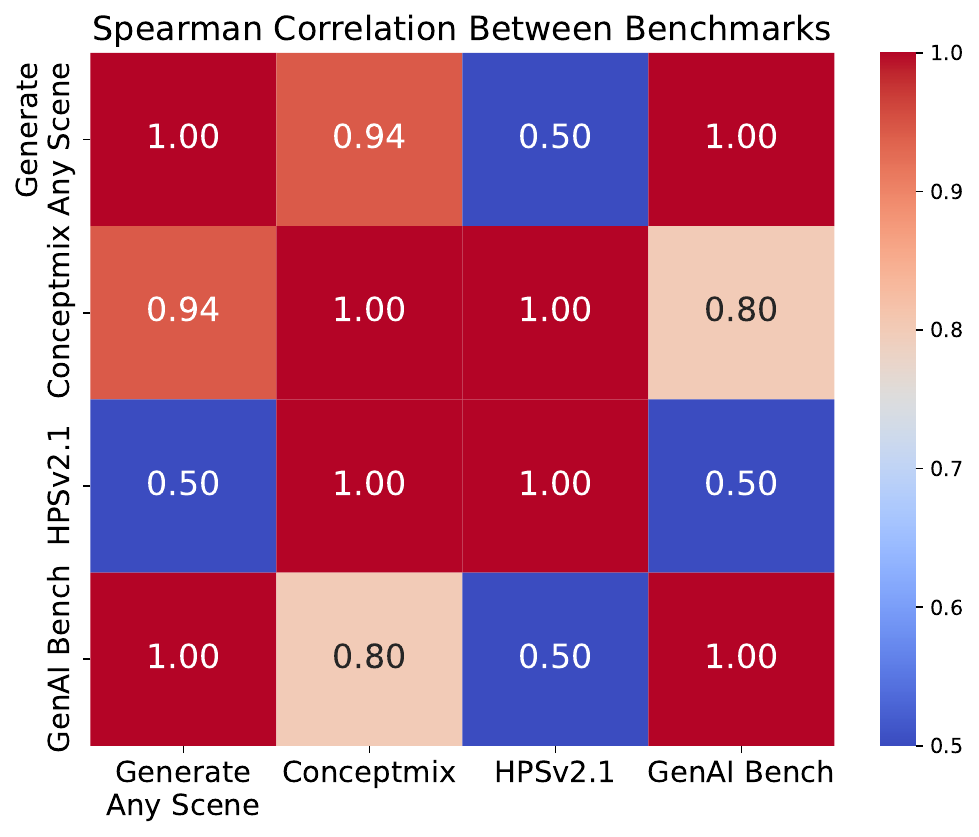}
  \caption{Correlation of \name with other popular \vision benchmarks.}
  \label{fig:heatmap}
\end{figure}
The \name benchmark uniquely relies entirely on synthetic captions to evaluate models. To assess the transferability of these synthetic captions, we analyzed the consistency in model rankings across different benchmarks~\cite{Wu2024ConceptMixAC,li2024genai,wu2023human}. Specifically, we identified the overlap of models evaluated by two benchmarks and computed the Spearman correlation coefficient between their rankings.

As shown in the figure~\ref{fig:heatmap}, \name demonstrates a strong correlation with other benchmarks, such as Conceptmix~\cite{Wu2024ConceptMixAC} and GenAI Bench~\cite{li2024genai}, indicating the robustness and reliability of \name's synthetic caption-based evaluations. This suggests that the synthetic captions generated by \name can effectively reflect model performance trends, aligning closely with those observed in benchmarks using real-world captions or alternative evaluation methods.

\begin{figure}[]
    \centering
    \begin{subfigure}[t]{0.48\linewidth}
        \centering
        \includegraphics[width=\linewidth]{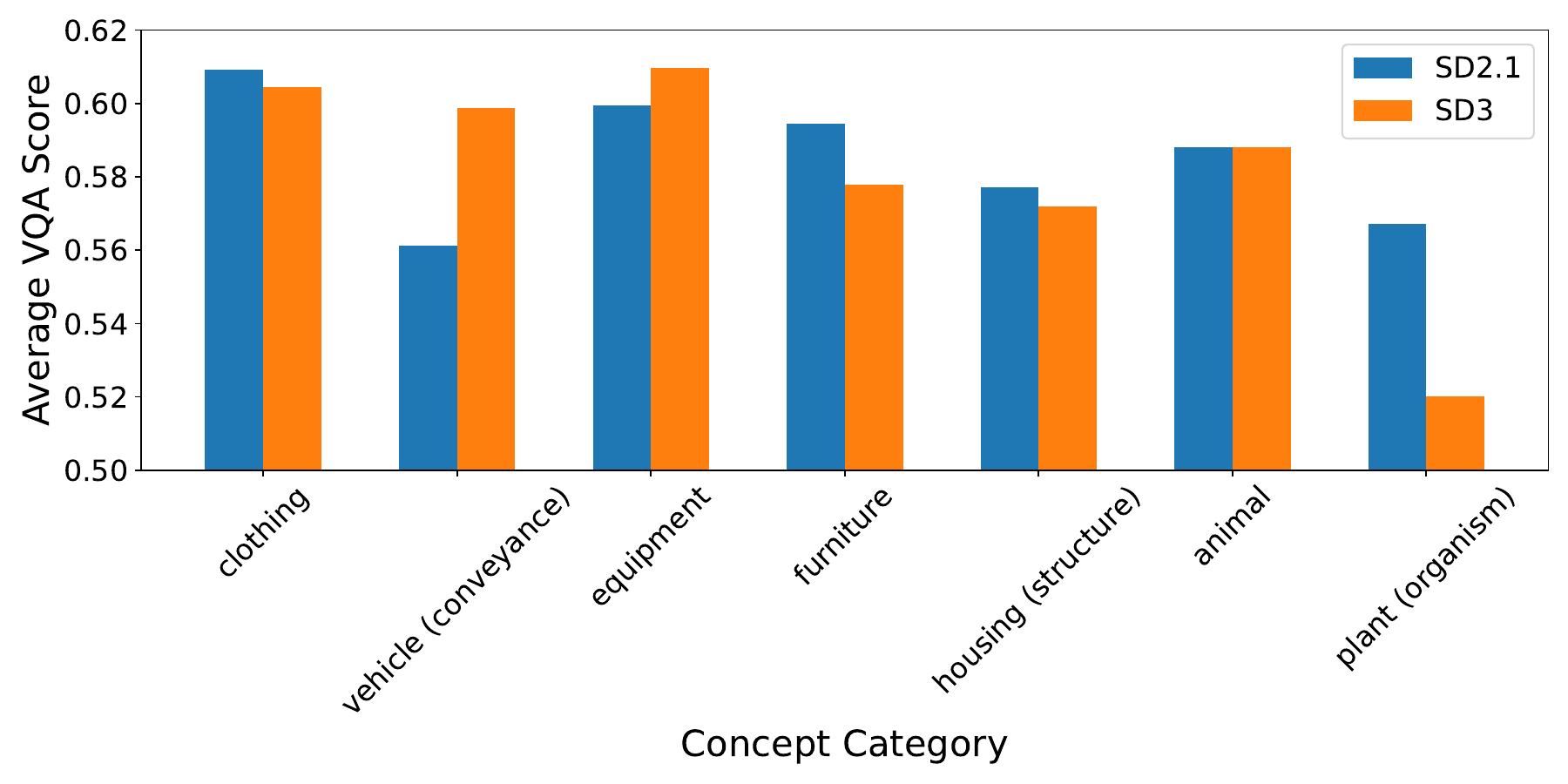}
        \caption{\sdtwoone vs. \sdthree on average \vqascore in fine-grained categories.}
    \end{subfigure}
    \hfill
    \begin{subfigure}[t]{0.48\linewidth}
        \centering
        \includegraphics[width=\linewidth]{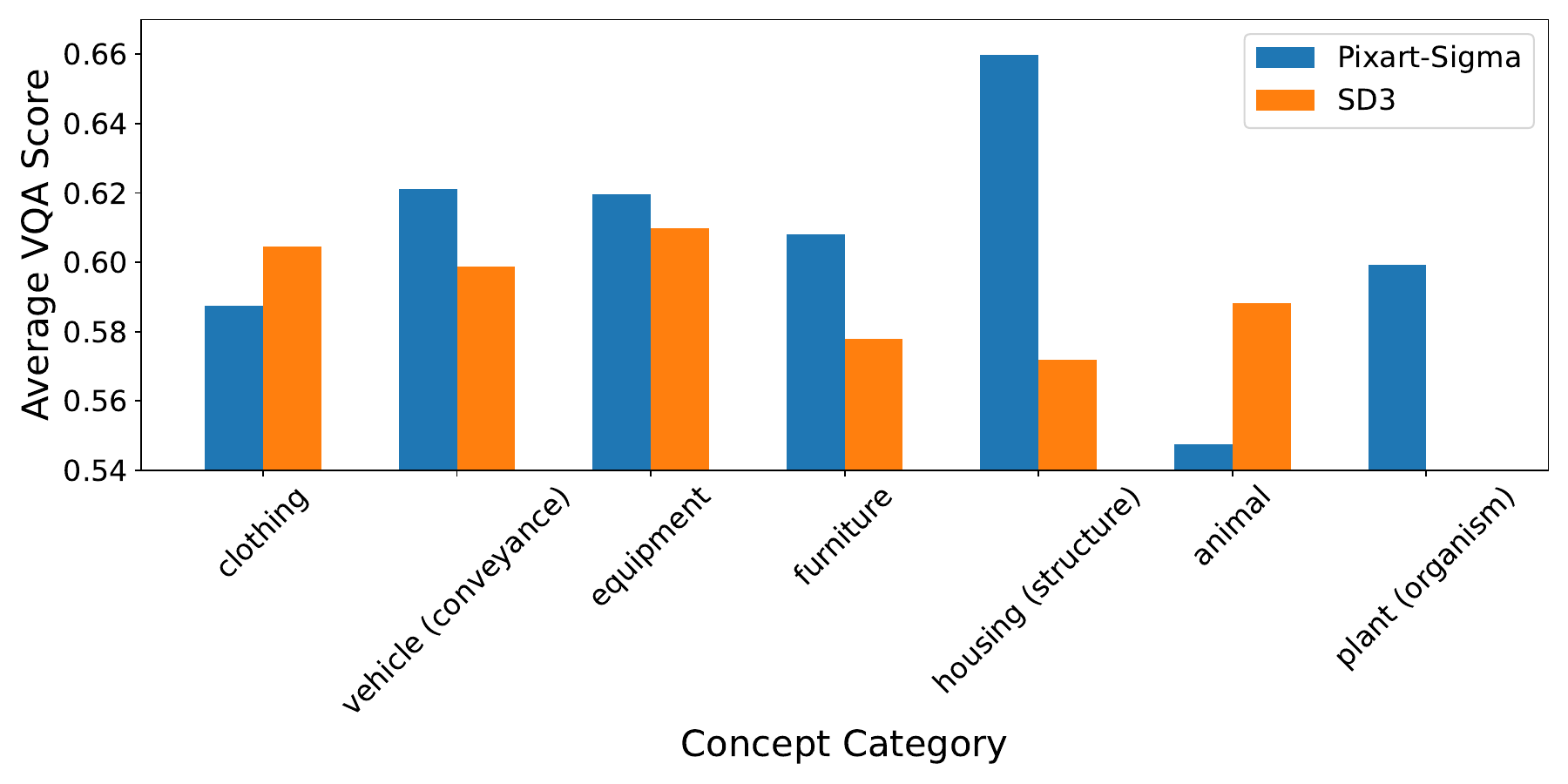}
        \caption{\pixartsigma vs. \sdthree on average \vqascore in fine-grained categories.}
    \end{subfigure}

    \vspace{1em}

    \begin{subfigure}[t]{0.48\linewidth}
        \centering
        \includegraphics[width=\linewidth]{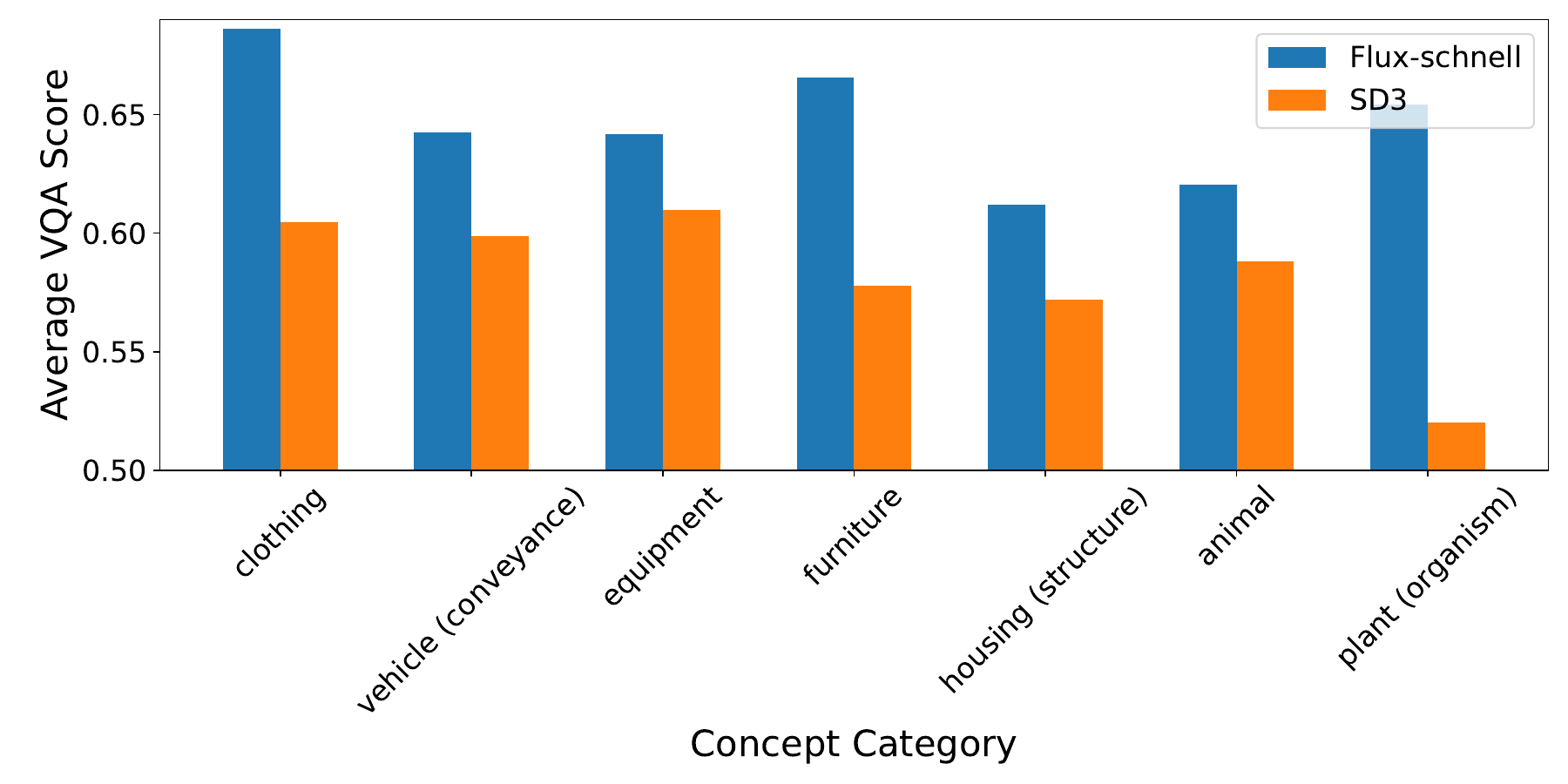}
        \caption{\fluxschnell vs. \sdthree on average \vqascore in fine-grained categories.}
    \end{subfigure}
    \hfill
    \begin{subfigure}[t]{0.48\linewidth}
        \centering
        \includegraphics[width=\linewidth]{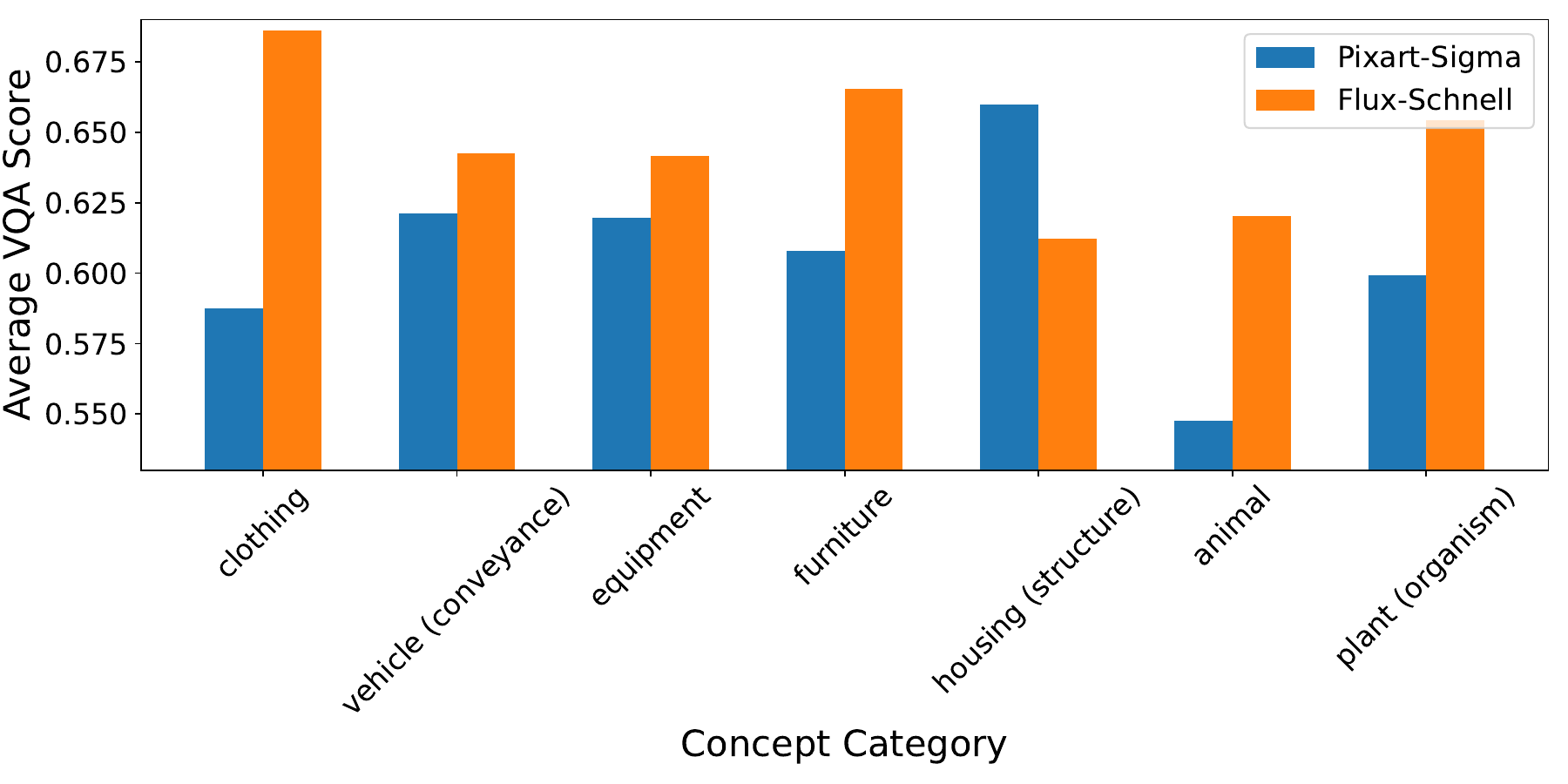}
        \caption{\pixartsigma vs. \fluxschnell on average \vqascore in fine-grained categories.}
    \end{subfigure}
    
    \caption{Pairwise comparison on average \vqascore in fine-grained categories.}
\end{figure}

\subsubsection{Case study: Pairwise fine-grained model comparison}
Evaluating models using a single numerical average score can be limiting, as different training data often lead models to excel in generating different types of concepts. By leveraging the taxonomy we developed for \name, we can systematically organize these concepts and evaluate each model’s performance on specific concepts over the taxonomy. This approach enables a more detailed comparison of how well models perform on individual concepts rather than relying solely on an overall average score. Our analysis revealed that, while the models may achieve similar average performance, their strengths and weaknesses vary significantly across different concepts. Here we present a pairwise comparison of models across different metrics.


\clearpage
\onecolumn
\newpage

\clearpage
\onecolumn
\newpage

\section{Details of Taxonomy of Visual Concepts}
\label{app:taxonomy}

To construct a scene graph, we utilize three primary types of metadata: objects, attributes, and relations, which represent the structure of a visual scene. Additionally, scene attributes—which include factors like image style, perspective, and video time span—capture broader aspects of the visual content. Together, the scene graph and scene attributes form a comprehensive representation of the scene.

Our metadata is further organized using a well-defined taxonomy, enhancing the ability to generate controllable captions. This hierarchical taxonomy not only facilitates the creation of diverse scene graphs, but also enables fine-grained and systematic model evaluation.

\paragraph{Objects.}

To enhance the comprehensiveness and taxonomy of object data, we leverage noun synsets and the structure of WordNet~\cite{wordnet}. In WordNet, a \textit{physical object} is defined as \textit{"a tangible and visible entity; an entity that can cast a shadow."} Following this definition, we designate the \textit{physical object} as the root node, constructing a hierarchical tree with all \textit{28,787} hyponyms under this category as the set of objects in our model.

Following WordNet’s hypernym-hyponym relationships, we establish a tree structure, linking each object to its primary parent node based on its first-listed hypernym. For objects with multiple hypernyms, we retain only the primary parent to simplify the hierarchy. Furthermore, to reduce ambiguity, if multiple senses of a term share the same parent, we exclude that term itself and reassign its children to the original parent node. This approach yields a well-defined and disambiguated taxonomy.

\paragraph{Attributes.}
The attributes of a scene graph represent properties or characteristics associated with each object. We classify these attributes into \textit{nine} primary categories. For \textit{color}, we aggregate \textit{677} unique entries sourced from Wikipedia~\cite{wikipedia_colors}. The \textit{material} category comprises \textit{76} types, referenced from several public datasets~\cite{material1-meta, material2-meta, material3-meta}. The \textit{texture} category includes \textit{42} kinds from the Describable Textures Dataset~\cite{texture-meta}, while the \textit{architectural style} encompasses \textit{25} distinct styles~\cite{archstyle-meta}. Additionally, we collect \textit{85} \textit{states}, \textit{41} \textit{shapes}, and \textit{24} \textit{sizes}. For \textit{human descriptors}, we compile 59 terms across subcategories, including body type and height. Finally, we collect \textit{465} common \textit{adjectives} covering general characteristics of objects to enhance the descriptive richness of our scene graphs.

\paragraph{Relationships.}
We leverage the Robin dataset~\cite{robin} as the foundation for relationship metadata, encompassing six key categories: spatial, functional, interactional, social, emotional, and symbolic. With 10,492 relationships, the dataset provides a comprehensive and systematic repository that supports modeling diverse and complex object interactions. Its extensive coverage captures both tangible and abstract connections, forming a robust framework for accurate scene graph representation.
\paragraph{Scene Attributes.}
In \vision tasks, people mainly focus on creating realistic images and art from a text description~\cite{imagen, dalle2, dalle3}. For artistic styles, we define scene attributes using \textit{76} renowned \textit{artists}, \textit{41} \textit{genres}, and \textit{126} \textit{painting styles} from WikiArt~\cite{style-meta}, along with \textit{29} common \textit{painting techniques}. For realistic imagery, we construct camera settings attributes across 6 categories: camera models, focal lengths, perspectives, apertures, depths of field, and shot scales. The camera models are sourced from the 1000 Cameras Dataset~\cite{camera-meta}, while the remaining categories are constructed based on photography knowledge and common captions in \vision tasks~\cite{sora, wangDiffusionDBLargescalePrompt2022}. To control scene settings, we categorize location, weather and lighting attributes, using 430 diverse locations from Places365~\cite{location-meta}, alongside \textit{76 weathers} and \textit{57 lighting conditions}. For video generation, we introduce attributes that describe dynamic elements. These include 12 types of camera rig, 30 distinct camera movements, 15 video editing styles, and 27 temporal spans. The comprehensive scene attributes that we construct allow for the detailed and programmatic \vision generation.

\clearpage
\onecolumn
\newpage

\section{Details of self-improving models with synthetic captions (Section 3)}
\label{app:appone}
\subsection{Experiment details}
\subsubsection{Captions Preparation}
To evaluate the effectiveness of our iterative self-improving \vision model, we generated three distinct sets of 10K captions using \name, covering a sample complexity range from 3 to 12. These captions were programmatically created to reflect a spectrum of structured scene graph compositions, designed to challenge and enrich the model’s learning capabilities.

For comparative analysis, we leveraged the Conceptual Captions (CC3M)~\cite{changpinyo2021cc12m} dataset, a large-scale benchmark containing approximately 3.3 million image-caption pairs sourced from web alt-text descriptions. CC3M is renowned for its diverse visual content and natural language expressions, encompassing a wide range of styles, contexts, and semantic nuances.

To ensure fair comparison, we randomly sampled three subsets of 10K captions from the CC3M dataset, matching the \name-generated caption sets in size. This approach standardizes data volume while enabling direct performance evaluation. The diversity and semantic richness of the CC3M captions serve as a robust benchmark to assess whether \name-generated captions can match or exceed the descriptive quality of real-world data across varied visual contexts.

\subsubsection{Dataset Construction and Selection Strategies}
For the captions generated by \name, we employed a top-scoring selection strategy to construct the fine-tuning training dataset, using a random selection strategy as a baseline for comparison. Specifically, for each caption, the model generated eight images. Under the top-scoring strategy, we evaluated the generated images using the VQA score and selected the highest-scoring image as the best representation of the caption. This process yielded 10K top-ranked images per iteration, from which the top 25\% (approximately 2.5k images) with the highest VQA scores were selected to form the fine-tuning dataset.

In the random selection strategy, one image was randomly chosen from the eight generated per caption, and 25\% of these 10K randomly selected images were sampled to create the fine-tuning dataset, maintaining parity in data size.

For the CC3M dataset, each caption was uniquely paired with a real image. From the 10K real image-caption pairs sampled from CC3M, the top 25\% with the highest VQA scores were selected as the fine-tuning training dataset. This ensured consistency in data size and selection criteria across all methods, facilitating a rigorous and equitable comparison of fine-tuning strategies.

\subsubsection{Fine-tuning details} \label{Self-improve-lora}
We fine-tuned the \sdonefive using the LoRA technique. The training was conducted with a resolution of 512 $\times$ 512 for input images and a batch size of 8. Gradients were accumulated over two steps. The optimization process utilized the AdamW optimizer with $\beta_1 = 0.9$, $\beta_2 = 0.999$, an $\epsilon$ value of $1 \times 10^{-8}$, and a weight decay of $10^{-2}$. The learning rate was set to $1 \times 10^{-4}$ and followed a cosine scheduler for smooth decay during training. To ensure stability, a gradient clipping threshold of 1.0 was applied. The fine-tuning process was executed for one epoch, with a maximum of 2500 training steps. For the LoRA-specific configurations, we set the rank of the low-rank adaptation layers and the scaling factor $\alpha$ to be 128.

After completing fine-tuning for each epoch, we set the LoRA weight to 0.75 and integrate it into \sdonefive to guide image generation and selection for the next subset. For the CC3M dataset, images from the subsequent subset are directly selected.

In the following epoch, the fine-tuned LoRA parameters from the previous epoch are loaded and used to resume training on the current subset, ensuring continuity and leveraging the incremental improvements from prior iterations.

\begin{figure*}[!t]
    \centering
    \includegraphics[width=1\textwidth]{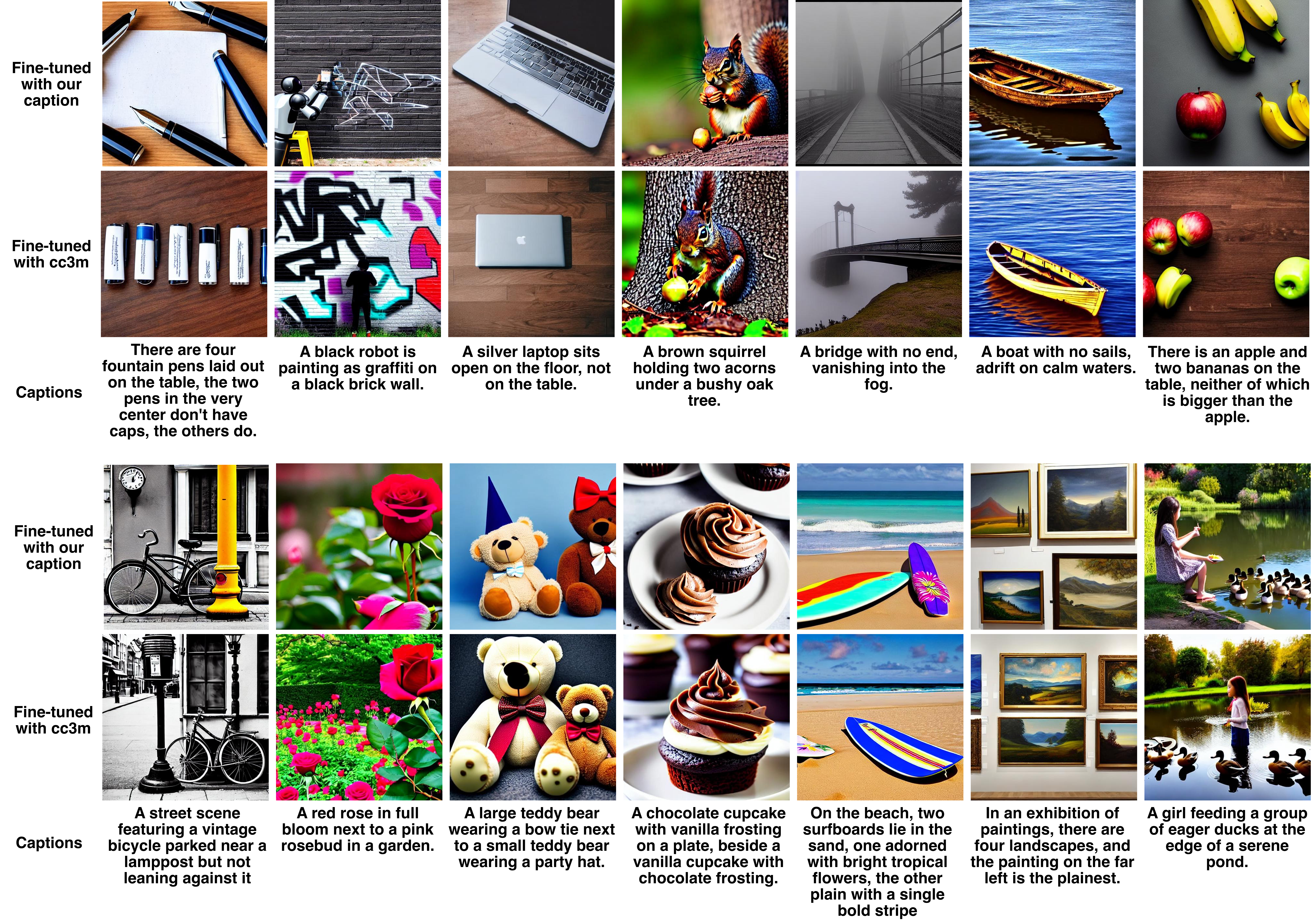}
    \caption{\textbf{Visualization of Different Caption Fine-Tuning.}}
    \label{fig:caption_finetuningg_vis}
\end{figure*}

In Figure~\ref{fig:caption_finetuningg_vis}, we present results using our captions and the CC3M captions. The model fine-tuned with captions generated by \name demonstrates superior performance in terms of text semantic relevance and the generation of complex compositional scenes.

\subsection{Evaluation on TIFA bench}
\begin{figure}[h]
    \centering
    \includegraphics[width=\linewidth]{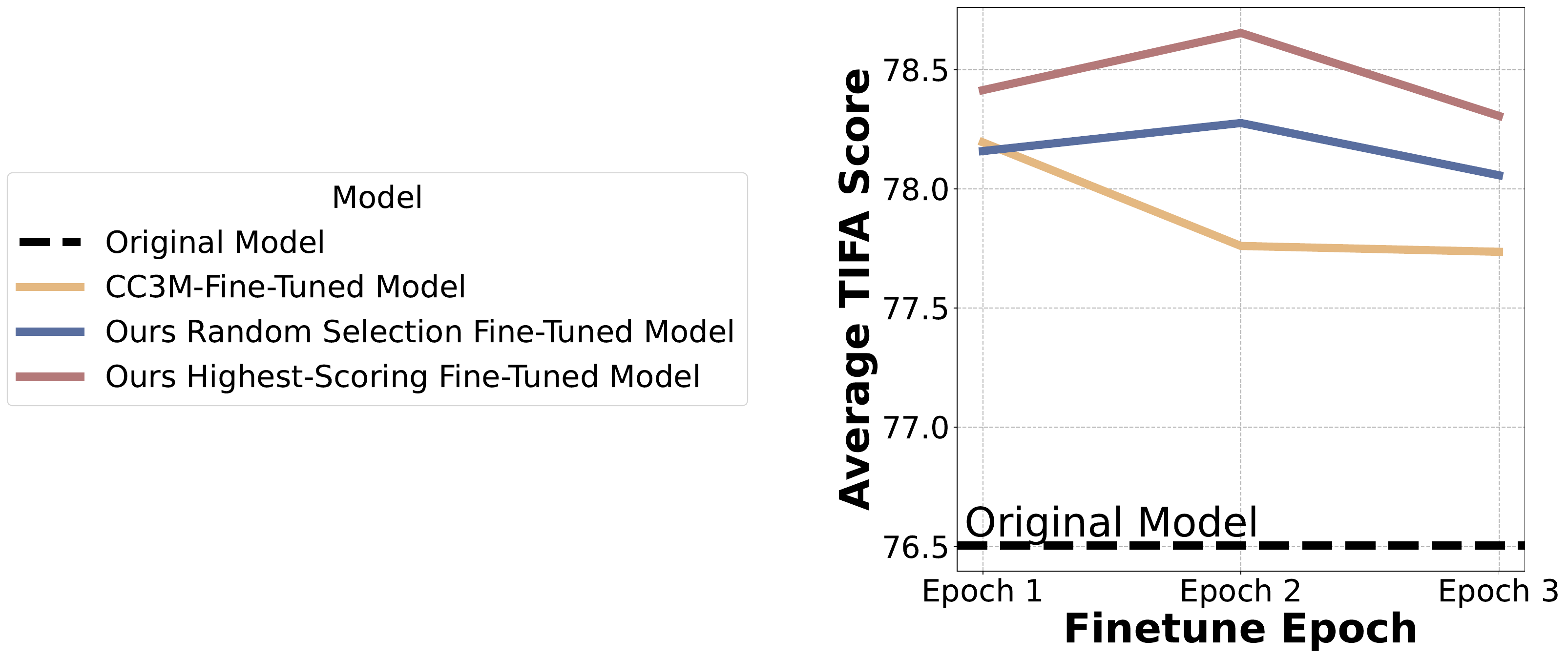}
    \caption{\textbf{Results for Application 1: Self-Improving Models}.
    Average TIFA score of \sdonefive fine-tuned with different data over TIFA Bench.}
    \label{fig:tifa-result}
\end{figure}

Aside from our own test set and GenAI benchmark, we also evaluated our fine-tuned \image models on the Tifa Bench (Figure~\ref{fig:tifa-result}), where we observed the same trend: models fine-tuned with our captions consistently outperformed the original \sdonefive and CC3M fine-tuned models.

\subsection{Additional real-data baselines}
\label{app:real-baselines}

\paragraph{Setup.}
We conduct more experiments comparing \name synthetic captions to other real-world caption sources. 
We sampled 10K captions from MS-COCO-2017 and LAION-COCO for one-epoch LoRA fine-tuning under same experimental settings. The results on \name test set are summarized in Table~\ref{tab:more-real}.
\begin{table}[h]
\centering
\caption{Self-improvement on \name Test (VQA). One-epoch finetuning, equal budget.}
\begin{tabular}{l c}
\toprule
Method & VQA $\uparrow$ \\
\midrule
Baseline (\sdonefive) & 0.508 \\
MS-COCO-2017 & 0.508 \\
LAION-COCO & 0.510 \\
CC3M & 0.508 \\
GAS (Random) & 0.524 \\
GAS (Top-Score) & \textbf{0.530 }\\
\bottomrule
\end{tabular}
\label{tab:more-real}
\end{table}

\paragraph{Findings.}
Fine‑tuning with MS‑COCO‑2017 and LAION‑COCO captions yields results similar to CC3M, with none surpassing the significant improvements achieved by our \name captions. We think that although MS-COCO-2017 and LAION captions are generally high-quality and well‑aligned with images, they offer limited compositional diversity. These additional results confirm that the observed gains are not specific to CC3M but generalize across other widely used real-caption datasets. This further supports our claim that the compositional diversity of \name synthetic captions drives the improvement.

\subsection{Full fine-tuning vs. LoRA fine-tuning}
\label{app:fullft}

\paragraph{Setup.}
We replicate the self-improvement pipeline with \emph{full fine-tuning} and compare three strategies: \name captions with high‑score selection, \name captions with random selection, and CC3M captions as the real‑data baseline. The results are shown in Tables~\ref{tab:full-gas} and~\ref{tab:full-genai}.
\begin{table}[!h]
\centering
\caption{Results on \name test set under full fine-tuning. (\vqascore)}
\begin{tabular}{lccc}
\toprule
Method & Iter-1 & Iter-2 & Iter-3 \\
\midrule
Baseline & 0.508 & --- & --- \\
CC3M (Full FT) & 0.496 & 0.518 & 0.519 \\
GAS (Rand, Full FT) & 0.510 & 0.519 & 0.520 \\
GAS (Top, Full FT) & \textbf{0.510} & \textbf{0.534 }& \textbf{0.540} \\
\bottomrule
\end{tabular}
\label{tab:full-gas}
\end{table}

\begin{table}[!h]
\centering
\caption{Results on GenAI-Bench under full fine-tuning. (\vqascore)}
\begin{tabular}{lccc}
\toprule
Method & Iter-1 & Iter-2 & Iter-3 \\
\midrule
Baseline & 0.617 & --- & --- \\
CC3M (Full FT) & 0.589 & 0.619 & 0.622 \\
GAS (Rand, Full FT) & 0.599 & 0.621 & 0.617 \\
GAS (Top, Full FT) & \textbf{0.620} & \textbf{0.626} & \textbf{0.634} \\
\bottomrule
\end{tabular}
\label{tab:full-genai}
\end{table}
\paragraph{Findings.}
Using our \name captions with high score selection not only improves performance consistently across iterations but also surpasses CC3M at every stage. The full fine-tuning results confirm that our captions and strategy's effectiveness is not dependent on the specific training approach (LoRA vs. full fine-tuning). The consistent improvement patterns across both evaluation benchmarks validate the robustness of our iterative self-improvement framework.





\clearpage
\onecolumn
\newpage

\section{Details of distilling targeted capabilities (Section 4)}
\label{app:apptwo}

\subsection{Collecting hard concepts}
We evaluate both models on 10K \name captions and select 81 challenging object concepts where \sdonefive and \dalle exhibit the largest gap. To determine the score for each concept, we calculated the average \tifa of the captions containing that specific concept. For each targeted-generated caption, we generate four images and use the one with the highest \vqascore. The full list of hard concepts is shown below:

\begin{enumerate}[label=\arabic*.]
    \item cloverleaf
    \item aerie (habitation)
    \item admixture
    \item webbing (web)
    \item platter
    \item voussoir
    \item hearthstone
    \item puttee
    \item biretta
    \item yarmulke
    \item surplice
    \item overcoat
    \item needlepoint
    \item headshot
    \item photomicrograph
    \item lavaliere
    \item crepe
    \item tureen
    \item bale
    \item jetliner
    \item square-rigger
    \item supertanker
    \item pocketcomb
    \item filament (wire)
    \item inverter
    \item denture
    \item lidar
    \item volumeter
    \item colonoscope
    \item synchrocyclotron
    \item miller (shaper)
    \item alternator
    \item dicer
    \item trundle
    \item paddle (blade)
    \item harmonica
    \item piccolo
    \item handrest
    \item rundle
    \item blowtorch
    \item volleyball
    \item tile (man)
    \item shuttlecock
    \item jigsaw
    \item roaster (pan)
    \item maze
    \item belt (ammunition)
    \item gaddi
    \item drawer (container)
    \item tenter
    \item pinnacle (steeple)
    \item pegboard
    \item afterdeck
    \item scaffold
    \item catheter
    \item broomcorn
    \item spearmint
    \item okra (herb)
    \item goatsfoot
    \item peperomia
    \item ammobium
    \item gazania
    \item echinocactus
    \item birthwort
    \item love-in-a-mist (passionflower)
    \item ragwort
    \item spicebush (allspice)
    \item leadplant
    \item barberry
    \item hamelia
    \item jimsonweed
    \item undershrub
    \item dogwood
    \item butternut (walnut)
    \item bayberry (tree)
    \item lodestar
    \item tapa (bark)
    \item epicalyx
    \item blackberry (berry)
    \item stub
    \item shag (tangle)
\end{enumerate}

\subsection{Experiment details}
We conducted targeted fine-tuning experiments on \sdonefive to evaluate \name's effectiveness in distilling model compositionality and learning hard concepts. For each task, we selected a dataset of 778 \name captions paired with images generated by \dalle. 
For compositionality, we selected multi-object captions from the existing dataset of 10K \name captions and paired them with the corresponding images generated by \dalle. To address hard concept learning, we first used \sdonefive to generate images based on the 10K \name captions and identified the hard concepts with the lowest VQA scores. These concepts were then used to create a subset of objects, which we recombined into our scene-graph based captions with complexity levels ranging from 3 to 9. Finally, we used \dalle to generate corresponding images for these newly composed captions.

The fine-tuning configurations were consistent with those used in the self-improving setup (Appendix~\ref{Self-improve-lora}). To accommodate the reduced dataset size, the maximum training steps were set to 1000. 

As a baseline, we randomly selected 778 images from 10K \name-generated images, using captions produced by \name. This ensured a controlled comparison between the targeted and random fine-tuning strategies.

\subsection{Benchmark against web-crawled caption–image pairs}
\label{app:webcrawl}

\paragraph{Setup.}
We conduct additional experiments to benchmark against alternative data sourcing strategies, specifically comparing our \dalle distillation approach with web-scraped real images. Using the Bing Image Search API, we retrieve images matching our multi‑object and hard‑concept captions and constructed two datasets of equivalent scale for comparison. We then apply the same fine‑tuning setup described in Application 2. The results are shown in Table~\ref{tab:Web-crawled}:
\begin{table}[h]
\centering
\caption{Comparison of VQA scores from targeted fine-tuning on different data sources. (\sdonefive)}
\begin{tabular}{lccc}
\toprule
Test Set & Original & \dalle Distill & Web-crawled \\
\midrule
Hard Concept & 0.303 &\textbf{ 0.361} & 0.258 \\
Multi-object & 0.271 &\textbf{ 0.325 }& 0.264 \\
\bottomrule
\end{tabular}
\label{tab:Web-crawled}
\end{table}
\paragraph{Findings.}
The results show that web-scraped images not only failed to improve performance but actually degraded model capabilities.

Upon examination of the retrieved images, we identify several critical issues. The web-crawled images contain significant noise, including watermarks, overlaid text, and irrelevant visual element. Our hard concept and multi-object captions feature high compositional complexity and novel object combinations that rarely exist in real-world photographs. The retrieved images show poor relevance to our systematically designed compositional scenarios, as real-world images cannot adequately represent the diverse and controlled compositional variations we programmatically generate. Thus, training on such misaligned data appears to introduce incorrect visual-textual associations, leading to performance degradation rather than improvement.

\begin{table}[!h]
\centering
\caption{\vqascore of targeted distillation on \textit{FLUX.1-dev}.}
\begin{tabular}{lcc}
\toprule
Test Set & Original & Fine-tuned \\
\midrule
Hard Concept & 0.303 & \textbf{0.361} \\
Multi-object & 0.271 & \textbf{0.325} \\
\bottomrule
\end{tabular}
\label{distillation-flux}
\end{table}
\subsection{Distillation on Flux.1-dev}
\label{app:flux-distill}

\paragraph{Setup.}
We further apply our distillation framework to \textit{FLUX.1-dev}, a current SOTA open-source model, using \dalle‑generated images of hard concepts and multi‑object captions to distill these capabilities into \textit{FLUX.1-dev}. The results are shown in the Table~\ref{distillation-flux}:

\paragraph{Findings.}
The results demonstrate that our approach's effectiveness extends to state-of-the-art models (\textit{FLUX.1-dev}). The distillation approach yields substantial improvements on challenging compositional tasks.

\clearpage
\onecolumn
\newpage

\section{Details of reinforcement learning with a synthetic reward function (Section 5)}
\label{app:appzero}
\subsection{Training data preparation}
We adopt SimpleAR-0.5B-SFT~\cite{wang2025simplear} as our base model. Given that SimpleAR-0.5B-SFT is pretrained on high-quality real image datasets such as LAION~\cite{schuhmann2022laion} and CC3M~\cite{sharma2018conceptual}, we aim to mitigate potential distributional shift between the original training data and the reinforcement learning phase. To this end, we perform metadata pre-selection for \name by analyzing the frequency of each object category appearing in the LAION dataset. Leveraging the controllable compositional capabilities of \name, we filter object categories by selecting the top 10\% most frequent entries and constrain scene complexity to 3–6 objects per scene. Based on these conditions, we synthesize a set of 10K captions, ensuring semantic alignment with the base model’s pretraining distribution while maintaining structural and content diversity.

\subsection{Experiment details}
The detailed training configuration is provided in Table~\ref{tab:training_config}. We utilize 8 $\times$ NVIDIA H100 GPUs (80GB HBM3), with one GPU allocated for online generation using vLLM. The total training time is approximately 14 hours.
\begin{table}[h]
\centering
\caption{Scene-graph based GRPO Fine-tuning Configuration for SimpleAR}
\label{tab:training_config}
\begin{tabular}{ll}
\toprule
\textbf{Component} & \textbf{Details} \\
\midrule
Model Name & SimpleAR-0.5B-SFT \\
Model Size & $\sim$0.5B parameters \\
Training Policy & GRPO  \\
Inference Engine & vLLM (GPU utilization = 0.7) \\
Completion Length & 4096 tokens \\
Training Epochs & 1 \\
Batch Size per Device & 4 \\
Learning Rate & $1 \times 10^{-5}$ \\
Scheduler & Cosine Annealing (min lr rate = 0.1) \\
Warm-up Ratio & 0.1 \\
Gradient Accumulation & 1 \\
\bottomrule
\end{tabular}
\end{table}

\begin{figure}[h]
    \centering
    \includegraphics[width=0.75\linewidth]{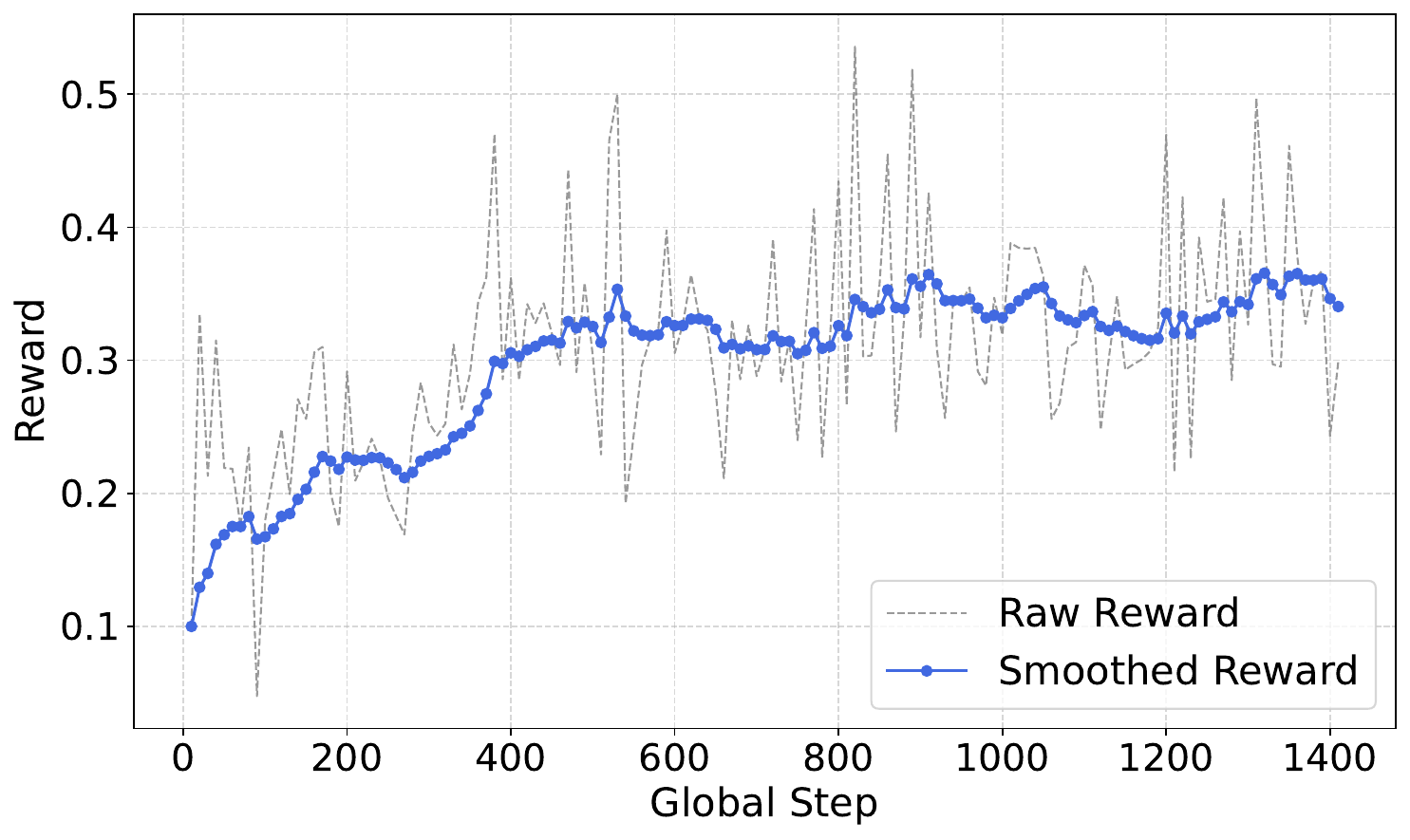}
    \caption{Reward progression during scene-graph based GRPO training.}
    \label{fig:reward_curve}
\end{figure}
Figure~\ref{fig:reward_curve} illustrates the reward progression during training. A noticeable improvement in reward is observed following the application of a learning rate of 1e-5 combined with a warm-up strategy. Overall, the reward increases by approximately 0.2, indicating effective learning under the adjusted training configuration.

In Table~\ref{tab:reward_exp}, we observe that the reproduced results of baseline models on DPG-Bench and GenEval Bench are slightly lower than those reported in the original paper. Considering the inherent stochasticity in generative model outputs, we cite the original results for comparison. For GenAI-Bench, all reported results are based on our own experimental evaluations.

\subsection{Reward variants and ablations}
\label{app:reward-variants}

\paragraph{Setup.}
To verify the observed gains arise specifically from the scene-graph–generated QA reward, rather than simply from using any QA-based reward, 
we conduct experiments incorporating manually annotated QA datasets, VQAv2, as additional reward signals under the same RLHF framework. We sample 10K images from VQAv2, with corresponding QA pairs, matched them to COCO2017 captions, and apply same training frameworks to SimpleAR-0.5B-SFT with RL training. The results on GenAI Bench are shown in the table:
\begin{table}[!h]
\centering
\caption{GenAI Bench performance (VQA) under RLHF with different reward sources. All models start from \emph{SimpleAR-0.5B-SFT}.}
\begin{tabular}{lccc}
\toprule
Method & Basic $\uparrow$ & Advanced $\uparrow$ & All $\uparrow$ \\
\midrule
SimpleAR-0.5B-SFT & 0.74 & 0.60 & 0.66 \\
SimpleAR-0.5B-RL (CLIP) & \textbf{0.75} & 0.60 & 0.67 \\
SimpleAR-0.5B-RL (VQAv2) & 0.73 & 0.59 & 0.66 \\
SimpleAR-0.5B-RL (Ours) & \textbf{0.75} & \textbf{0.61} & \textbf{0.68} \\
\bottomrule
\end{tabular}
\end{table}

\paragraph{Findings.}
The results show that using VQAv2 captions and QA pairs as rewards yields even lower performance than CLIP‑based RL training. Furthermore, we observe minimal reward improvement from VQA signals throughout training. We attribute this to the fact that, although VQAv2 QA pairs are rich, the underlying image captions fail to cover enough visual elements, leading to a mismatch between QA pairs and captions that undermines RLHF reward alignment.

This highlights the inherent difficulty and cost of constructing high-quality image-caption and QA annotations, whereas our method leverages scene-graph structures to systematically generate synthetic caption-QA pairs at minimal cost with unique advantages.

\clearpage
\onecolumn
\newpage

\section{Details of improving generated-content detection (Section 6)}
\label{app:appthree}

\subsection{Experiment details}
In this section, our goal is to validate that the more diverse captions generated by \name can complement existing datasets, which are predominantly composed of real-world images paired with captions. By doing so, we aim to train AI-generated content detectors to achieve greater robustness.

\paragraph{Dataset preparation}
We conducted comparative experiments between captions generated by \name and entries from the $D^3$ dataset. From the $D^3$ dataset, we randomly sampled 10K entries, each including a caption, a link to a real image, and an image generated by SD v1.4. Due to some broken links, we successfully downloaded 8.5K real images and retained 10K SD v1.4-generated images. We also used SD v1.4 to generate images based on 10K \name captions.

We varied the training data sizes based on the sampled dataset. Specifically, we sampled N real images from the 10K $D^3$ real images. For synthetic data, we compared N samples exclusively from $D^3$ with a mixed set of N/2 samples from 10K \name images and N/2 sampled from $D^3$, ensuring a total of N synthetic samples. Combined, this resulted in 2N training images. We tested 2N across various sizes, ranging from 2K to 10K.

\paragraph{Detector architecture and training}
We employed ViT-T~\cite{tinyvit} and ResNet-18~\cite{he2016deep} as backbones for the detection models. Their pretrained parameters on ImageNet-21K were frozen, and the final classification head was replaced with a linear layer using a sigmoid activation function to predict the probability of an image being AI-generated. During training, We used Binary Cross-Entropy (BCE) as the loss function, and the AdamW optimizer was applied with a learning rate of $2e^{-3}$. Training was conducted with a batch size of 256 for up to 50 epochs, with early stopping triggered after six epochs of no improvement in validation performance.
\paragraph{Testing}
To evaluate the performance of models trained with varying dataset sizes and synthetic data combinations, we tested them on both GenImage and \name datasets to assess their in-domain and out-of-domain performance under different settings.

For GenImage, we used validation data from four models: SD v1.4, SD v1.5, MidJourney, and VQDM. Each validation set contained 8K real images and 8k generated images. For \name, we sampled 10K real images from CC3M and paired them with 10K generated images from each of the following models: \sdtwoone, \pixartalpha, \sdthree, and \playground. This created distinct test sets for evaluating model performance across different synthetic data sources.

\begin{table*}[h]
\caption{F1-Score Comparison of ResNet-18 and ViT-T Detectors Trained with $D^3$ and $D^3$+ \name Across In-Domain Settings}
\resizebox{\textwidth}{!}{%
\begin{tabular}{c|c|cc|cccccccccc}
\hline
\multirow{2}{*}{Detector}  & \multirow{2}{*}{\begin{tabular}[c]{@{}c@{}}Data Scale\\ (2N)\end{tabular}} & \multicolumn{2}{c|}{\begin{tabular}[c]{@{}c@{}}SDv1.4\\ (In-domain, same model)\end{tabular}} & \multicolumn{2}{c}{SDv2.1}            & \multicolumn{2}{c}{Pixart-$\alpha$}      & \multicolumn{2}{c}{SDv3-medium}       & \multicolumn{2}{c}{Playground v2.5}   & \multicolumn{2}{c}{\begin{tabular}[c]{@{}c@{}}Average\\ (In-domain, cross model)\end{tabular}} \\ \cline{3-14} 
                           &                                                                            & $D^3$ + Ours                         & $D^3$                                  & $D^3$ + Ours & $D^3$  & $D^3$ + Ours & $D^3$  & $D^3$ + Ours & $D^3$  & $D^3$ + Ours & $D^3$  & $D^3$ + Ours                              & $D^3$                               \\ \hline
\multirow{5}{*}{Resnet-18} & 2K                                                                         & 0.6561                                               & \textbf{0.6663}                        & 0.7682                       & 0.6750 & 0.7379                       & 0.606  & 0.7509                       & 0.6724 & 0.7380                       & 0.5939 & \textbf{0.7488}                                           & 0.6368                              \\
                           & 4K                                                                         & 0.6751                                               & \textbf{0.6812}                        & 0.7624                       & 0.6853 & 0.7328                       & 0.6494 & 0.7576                       & 0.7028 & 0.7208                       & 0.6163 & \textbf{0.7434}                                           & 0.6635                              \\
                           & 6K                                                                         & 0.6780                                               & \textbf{0.6995}                        & 0.7886                       & 0.6870 & 0.7493                       & 0.6586 & 0.7768                       & 0.7285 & 0.7349                       & 0.6335 & \textbf{0.7624}                                           & 0.6769                              \\
                           & 8K                                                                         & 0.6828                                               & \textbf{0.6964}                        & 0.7710                       & 0.6741 & 0.7454                       & 0.6418 & 0.7785                       & 0.7186 & 0.7215                       & 0.6033 & \textbf{0.7541}                                           & 0.6595                              \\
                           & 10K                                                                        & 0.6830                                               & \textbf{0.6957}                        & 0.7807                       & 0.6897 & 0.7483                       & 0.6682 & 0.7781                       & 0.7326 & 0.7300                       & 0.6229 & \textbf{0.7593}                                           & 0.6784                              \\ \hline
\multirow{5}{*}{ViT-T}     & 2K                                                                         & \textbf{0.6759}                                      & 0.6672                                 & 0.7550                       & 0.6827 & 0.7585                       & 0.6758 & 0.7473                       & 0.6941 & 0.7327                       & 0.6106 & \textbf{0.7484}                                           & 0.6658                              \\
                           & 4K                                                                         & \textbf{0.6878}                                      & 0.6871                                 & 0.7576                       & 0.7000 & 0.7605                       & 0.7071 & 0.7549                       & 0.7217 & 0.7221                       & 0.6144 & \textbf{0.7488}                                           & 0.6858                              \\
                           & 6K                                                                         & \textbf{0.6898}                                      & 0.6891                                 & 0.7663                       & 0.6962 & 0.7666                       & 0.7164 & 0.7629                       & 0.7238 & 0.7303                       & 0.6134 & \textbf{0.7565}                                           & 0.6875                              \\
                           & 8K                                                                         & 0.6962                                               & \textbf{0.6974}                        & 0.7655                       & 0.6894 & 0.7712                       & 0.7253 & 0.7653                       & 0.7253 & 0.7381                       & 0.6344 & \textbf{0.7600}                                           & 0.6936                              \\
                           & 10K                                                                        & \textbf{0.6986}                                      & 0.6984                                 & 0.7828                       & 0.6960 & 0.7777                       & 0.7275 & 0.7786                       & 0.7334 & 0.7330                       & 0.6293 & \textbf{0.7680}                                           & 0.6966                              \\ \hline
\end{tabular}
}

\label{appdixtab:deepfake_indomain}
\end{table*}

\subsection{Results}
Table~\ref{appdixtab:deepfake_outdomain} and Table~\ref{appdixtab:deepfake_indomain} evaluate the performance of ResNet-18 and ViT-T detection backbones trained on datasets of varying sizes and compositions across in-domain (same model and cross-model) and out-of-domain settings. While models trained with $D^3$ and \name occasionally underperform compared to those trained solely on $D^3$ in the in-domain same-model setting, they exhibit significant advantages in both in-domain cross-model and out-of-domain evaluations. These results demonstrate that incorporating our data (\name) into the training process enhances the detector's robustness. By supplementing existing datasets with \name under the same training configurations and dataset sizes, detectors achieve stronger cross-model and cross-dataset capabilities, highlighting improved generalizability to diverse generative models and datasets.

\begin{table*}[h]
\caption{F1-Score Comparison of ResNet-18 and ViT-T Detectors Trained with $D^3$ and $D^3$+ \name Across Out-of-Domain Settings}
\resizebox{\textwidth}{!}{%
\begin{tabular}{c|c|cccccccc}
\hline
\multirow{2}{*}{Detector}  & \multirow{2}{*}{\begin{tabular}[c]{@{}c@{}}Data Scale\\ (2N)\end{tabular}} & \multicolumn{2}{c}{SDv1.5}            & \multicolumn{2}{c}{VQDM}              & \multicolumn{2}{c}{Midjourney}        & \multicolumn{2}{c}{\begin{tabular}[c]{@{}c@{}}Average\\ (Out-of-domain)\end{tabular}} \\ \cline{3-10} 
                           &                                                                            & $D^3$ + Ours                         & $D^3$                                  & $D^3$ + Ours & $D^3$  & $D^3$ + Ours & $D^3$  & $D^3$ + Ours                         & $D^3$                          \\ \hline
\multirow{5}{*}{Resnet-18} & 2K                                                                         & 0.6515                       & 0.6591 & 0.5629                       & 0.5285 & 0.5803                       & 0.5647 & \textbf{0.5982}                                      & 0.5841                         \\
                           & 4K                                                                          & 0.6709                       & 0.6817 & 0.5693                       & 0.5428 & 0.6016                       & 0.5941 & \textbf{0.6139}                                      & 0.6062                         \\
                           & 6K                                                                         & 0.6750                       & 0.6963 & 0.5724                       & 0.5327 & 0.6084                       & 0.6072 & \textbf{0.6186}                                      & 0.6121                         \\
                           & 8K                                                                         & 0.6792                       & 0.6965 & 0.5716                       & 0.5282 & 0.6097                       & 0.5873 & \textbf{0.6202}                                      & 0.6040                         \\
                           & 10K                                                                        & 0.6814                       & 0.6955 & 0.5812                       & 0.5454 & 0.6109                       & 0.6040 & \textbf{0.6245}                                      & 0.6150                         \\ \hline
\multirow{5}{*}{ViT-T}     & 2K                                                                         & 0.6755                       & 0.6685 & 0.5443                       & 0.4966 & 0.6207                       & 0.6066 & \textbf{0.6135}                                      & 0.5906                         \\
                           & 4K                                                                         & 0.6845                       & 0.6865 & 0.5591                       & 0.4971 & 0.6416                       & 0.6149 & \textbf{0.6284}                                      & 0.5995                         \\
                           & 6K                                                                         & 0.6900                       & 0.6890 & 0.5580                       & 0.4948 & 0.6455                       & 0.6259 & \textbf{0.6313}                                      & 0.6032                         \\
                           & 8K                                                                         & 0.6940                       & 0.6969 & 0.5553                       & 0.4962 & 0.6495                       & 0.6387 & \textbf{0.6329}                                      & 0.6106                         \\
                           & 10K                                                                        & 0.6961                       & 0.6988 & 0.5499                       & 0.4975 & 0.6447                       & 0.6358 & \textbf{0.6302}                                      & 0.6107                         \\ \hline
\end{tabular}
}
\label{appdixtab:deepfake_outdomain}
\end{table*}

\clearpage
\onecolumn
\newpage

\section{Advantages of \name over LLM-Driven Scene Graph and Caption Generation}
\label{subsec:gas_vs_llm}

\name is conceptually superior to a well-prompted LLM for large-scale scene graph and corresponding captions generation. While modern LLMs are powerful, they do not provide the guarantees required for systematic, controllable, and reproducible enumeration of compositional structures. In contrast, \name explicitly enumerates graph topologies under user-specified constraints (e.g., complexity, topics, connectivity) and then deterministically instantiates them, yielding uniform coverage, strict structural validity, and high efficiency.
\paragraph{Controllability and Diversity.} \name explicitly enumerates scene graph structures and populates with user-specified configuration and taxonomy (e.g., complexity, topics, connectivity, etc.), ensuring systematic coverage of rare or unconventional compositions without requiring users to manually write prompts for desired structures. In contrast, an LLM tends to default to common patterns in its training distribution. For example, given only the metadata \{book, table, on\}, an LLM will prefer the statistically dominant configuration "the book is on the table", and struggle to produce the less common but equally valid “the table is on the book” without extensive prompt engineering. Moreover, such extensive or high-quality prompting for scene graph generation essentially requires the user to manually enumerate graph structures and design multiple templates in natural language, whereas GAS accomplishes this systematically with a single program.
\paragraph{Reduced Bias and Hallucination.} Relying on LLMs to generate large-scale captions inherently inherits their internal biases and increases the likelihood of hallucinating unseen or semantically inconsistent object configurations. \name avoids this by enumerating scene graphs and then deterministically mapping them to captions, producing text that is faithful by construction to the underlying graph structure.
\paragraph{Lower Cost and Higher Reproducibility.}
In \name, once a scene graph is enumerated, it is cached and reused across multiple populations, and it can also serve as a seed graph for controllable topological expansion without re-enumerating the entire structure. Combined with our fully programmatic operations, this makes large-scale generation substantially more cost-efficient. In contrast, relying on an LLM would require repeated API calls or prompt redesign for structural variant and new content, making the process both costly and labor-intensive.

To empirically validate these points, we compare \name against Gemini 2.5-flash on generating 10K scene graphs from our common metadata (3,649 items: 2,591 objects, 551 attributes, 507 relations). Because Gemini becomes increasingly error-prone when prompted with the full metadata list, we adopt a batching strategy: in each batch we randomly sample 5\% of the metadata (~182 items) and prompt the model to generate 20 scene graphs containing 3–12 elements.

Table~\ref{tab:dist_quality} shows the distribution quality and diversity of generated elements. \name achieves near-uniform usage across objects, attributes, and relations, with Gini coefficients between 0.14 and 0.17 and normalized entropy above 99.3\%. Gemini, in contrast, exhibits strong concentration (Gini 0.53–0.66) and substantially lower entropy (79.5–92.5\%), indicating a tendency to overuse a narrow subset of frequent categories. The top-10\% coverage further highlights this imbalance: under Gemini, 37.29\% of object occurrences and 50.38\% of relation occurrences are concentrated in only 10\% of the vocabulary, whereas \name remains close to the uniform ideal.

\begin{table}[t]
\centering
\caption{Distribution quality and diversity of generated scene graphs.}
\label{tab:dist_quality}
\begin{tabular}{lcccc}
\toprule
\multicolumn{2}{c}{Metric} & GAS (Ours) & Gemini 2.5-flash \\
\midrule
\multicolumn{4}{l}{\textbf{Gini Coefficient} ($\downarrow$)} \\
\quad & Objects    & 0.14 & 0.53 \\
\quad & Attributes & 0.14 & 0.57 \\
\quad & Relations  & 0.17 & 0.66 \\
\midrule
\multicolumn{4}{l}{\textbf{Normalized Entropy} ($\uparrow$)} \\
\quad & Objects    & 99.6\% & 92.5\% \\
\quad & Attributes & 99.3\% & 91.7\% \\
\quad & Relations  & 99.3\% & 79.5\% \\
\midrule
\multicolumn{4}{l}{\textbf{Top 10\% Coverage} ($\downarrow$)} \\
\quad & Objects    & 14.68\% & 37.29\% \\
\quad & Relations  & 15.41\% & 50.38\% \\
\bottomrule
\end{tabular}
\end{table}

Beyond distributional properties, we assess structural validity and data quality using strict schema-level checks (Table~\ref{tab:struct_quality}). \name produces 100\% structurally valid graphs with zero hallucinated elements. In contrast, only 49.1\% of Gemini’s outputs satisfy the schema. Common failure modes include treating relations as nodes (34.6\% of graphs), and omitting required \texttt{value} (31.2\%) or \texttt{type} (30.3\%) fields. Gemini also hallucinates 1{,}773 “unknown” objects (4.59\% of all objects) and 3{,}638 “unknown” relations. 

\begin{table}[h]
\centering
\caption{Structural validity and data quality of generated scene graphs.}
\label{tab:struct_quality}
\begin{tabular}{lcc}
\toprule
Metric & GAS (Ours) & Gemini 2.5-flash \\
\midrule
Structurally valid graphs                & 100\%           & 49.1\% \\
Graphs with relations as nodes (error)   & 0\%             & 34.6\% \\
Graphs missing \texttt{value} field      & 0\%             & 31.2\% \\
Graphs missing \texttt{type} field       & 0\%             & 30.3\% \\
Hallucinated ``unknown'' objects         & 0               & 1{,}773 (4.59\%) \\
Hallucinated ``unknown'' relations       & 0               & 3{,}638 \\
\bottomrule
\end{tabular}
\end{table}

Finally, \name is more efficient than LLM-based generation (Table~\ref{tab:efficiency}). Because \name uses programmatic enumeration, it generates 10K scene graphs in under one minute, with negligible cost. In contrast, Gemini requires 1.5 hours and incurs over \$50 of API cost for the same workload. Overall, \name provides a 90$\times$ speedup and near-zero marginal expense.

These results confirm that programmatic enumeration in \name outperforms LLM-based generation, providing the systematic guarantees of uniformity, validity, and efficiency.

\begin{table}[b]
\centering
\caption{Efficiency and cost of generating 10K scene graphs.}
\label{tab:efficiency}
\begin{tabular}{lcc}
\toprule
Metric & GAS (Ours) & Gemini 2.5-flash \\
\midrule
Generation time (10K graphs) & $< 1$ minute & 1.5 hours \\
Monetary cost                & Negligible   & $> \$50$ \\
\bottomrule
\end{tabular}
\end{table}

\section{Discussion}
\subsection{Commonsense and Plausibility Filtering}
\label{filtering_details}
\name enables systematic, controllable, and diverse compositional scene construction through programmatic scene graph enumeration. This allows the synthesized captions to cover not only realistic scenes commonly observed in the real world, but also uncommon, imaginative, and unrealistic scenes.
Many widely-used generative models, including DALL-E, Midjourney, and Sora/Sora2, derive much of their practical value from producing surreal, imaginative, or physically unlikely compositions (e.g., "an astronaut riding a horse on the moon," or "a raccoon astronaut with a glowing space donut"). Such prompts are not outliers; they reflect common user intents in art, game design, advertising, and entertainment. Users frequently employ abstract or fantastical combinations precisely to explore the model’s creativity, and the community often discusses and evaluates models based on performance on these highly “unrealistic” prompts. From a research perspective, a broad and systematically controlled compositional space is essential for improving and benchmarking modern generative models. Limiting sampling to only strictly "realistic" combinations would substantially reduce both the training and the evaluation value. 

Our approach is specifically designed to meet this need for diverse captions and systematic visual representations. At the same time, \name differentiates uncommon or unrealistic scenes from nonsensical scenes. The taxonomy enforces strong type-level constraints, e.g., architectural attributes apply only to buildings, human-specific attributes only to the “person” subtree, and attentional relations only between animate entities, ensuring that generated scenes remain meaningful and structurally valid, even when creatively unrealistic. Beyond these inherent structural constraints, \name additionally provides an optional two-stage commonsense and plausibility filtering mechanism to support use cases that require higher visual realism. (1) Pre-population filtering. We maintain for every object/attribute/relation its LAION-5B~\cite{schuhmann2022laion} frequency and embedding representation, and select candidates by jointly enforcing minimum frequency thresholds and semantic coherence: for each newly added relation or attribute attached to a given object, we compute the top-k semantically compatible candidates based on embedding similarity to that object. Likewise, when expanding a relation triple, we compute candidate object similarity to the anchor object within the triplet, including all attributes and relations already attached to the anchor, and then sample from the top-k most semantically compatible objects (where k is user-configurable). Users may further specify complexity limits to avoid highly complex scenes. (2) Post-population filtering. After population, once the scene graph is translated into a caption, we compute its Vera score~\cite{liu2023verageneralpurposeplausibilityestimation} and caption perplexity, and discard captions falling below plausibility or above perplexity thresholds. These mechanisms ensure that \name preserves meaningfulness while still enabling broad creative coverage.

\subsection{Social Bias}
Assessing social bias is important for understanding whether synthetic data introduces unintended shifts in model behavior. To examine this, we evaluate models on gender-related prompts from the DALL-Eval~\cite{Cho2023DallEval} benchmark, comparing SDv1.5, SDv1.5 fine-tuned on CC3M captions, and SDv1.5 fine-tuned on \name captions. The gender MAD results are shown in Table~\ref{tab:socal}.
The experiment shows that fine-tuning with \name does not amplify gender bias relative to the base model. We attribute this to our design choices. First, \name does not generate data by propagating textual descriptions or cultural associations from these sources; instead, our metadata is used purely as a structural vocabulary of objects, attributes, and relations. \name doesn't sample linguistic definitions or corpus-derived stereotypes from WordNet. Second, the systematic, programmatic nature of our scene-graph enumeration further reduces the pathways through which social bias present in real-world distributions could propagate. Also, any more debiased metadata can be plugged into \name engine seamlessly.

\begin{table}[h]
\centering
\small
\caption{Gender MAD Scores on DALLEval}
\begin{tabular}{@{}l c@{}}
\toprule
Model & MAD $\downarrow$ \\
\midrule
SDv1.5 & 0.3602 \\
FT w/ CC3M & 0.3476 \\
FT w/ GAS & 0.3555 \\
\bottomrule
\end{tabular}
\label{tab:socal}
\end{table}

\section{Limitation}
\paragraph{Programmatically generated prompts can be unrealistic and biased.} Programmatically generated prompts can be unrealistic and biased. Although our system is capable of producing a wide range of rare compositional scenes and corresponding prompts, some of these outputs may violate rules or conventions, going beyond what is even considered imaginable or plausible. We also implement a pipeline to filter the commonsense of the generated prompts using the \textit{Vera score} (a large language model-based commonsense metric) and \textit{Perplexity}, but we make this pipeline \textbf{optional}.

\paragraph{Linguistic diversity of programmatic prompts is limited.}
While \name excels at generating diverse and compositional scene graphs and prompts, its ability to produce varied language expressions is somewhat constrained. The programmatic approach to generating content ensures diversity in terms of the elements of the scene, but it is limited when it comes to linguistic diversity and the richness of expression. To address this, we introduce a pipeline that leverages large language models (LLMs) to paraphrase prompts, enhancing linguistic variety. However, this addition introduces new challenges. LLMs are prone to biases and hallucinations, which can affect the quality and reliability of the output. Furthermore, the use of LLMs risks distorting the integrity of the original scene graph structure, compromising the coherence and accuracy of the generated content. So we make this LLM paraphrase pipeline \textbf{optional} for our paper.

\paragraph{Toward curriculum-aware GRPO training.}
Our proposed \name framework plays a central role in GRPO training by providing structured scene graphs that serve as the foundation for a semantically grounded and controllable reward function. This design enables effective optimization by aligning generation objectives with fine-grained visual semantics. Beyond this, we also observe that \name also offers broader potential: the scene graphs it produces vary in complexity, such as in the number of objects, attributes, relationships and graph degree. These variations naturally correspond to different levels of generation difficulty and reward variance. This property suggests an opportunity for curriculum-based training, where the model could be progressively exposed to increasingly complex scene graphs. Such a strategy may improve training stability and efficiency, especially in the early stages of learning. We identify this as a promising direction for future work, further leveraging the controllability of \name to guide structured policy learning.




\newpage

\end{document}